\documentclass[10pt,twocolumn,letterpaper]{article}

\usepackage[pagenumbers]{cvpr} 

\usepackage{graphicx}
\usepackage{amsmath}
\usepackage{amssymb}
\usepackage{booktabs}
\usepackage{multirow}
\usepackage{tabularx}
\usepackage{cite}
\usepackage{wrapfig}
\usepackage{color}
\usepackage{colortbl}
\usepackage{paralist}
\usepackage{overpic}
\usepackage{enumitem}
\definecolor{myRed}{rgb}{0.8,0.2,0.2}
\definecolor{myBlue}{rgb}{0.2,0.2,0.8}

\newcommand{\myPara}[1]{\vspace{5pt}\noindent\textbf{#1}}
\usepackage[pagebackref,breaklinks,colorlinks]{hyperref}

\newcounter{question}
\newcommand{\question}[1]{\item[Q\refstepcounter{question}\thequestion.] \textit{#1}}
\newcommand{\answer}[1]{\item[A\thequestion.] #1}
\def\sArt{state-of-the-art~}
\usepackage[capitalize]{cleveref}
\usepackage[linesnumbered, ruled]{algorithm2e}
\SetKwRepeat{Do}{do}{while}%

\newtheorem{assumption}{Assumption}

\usepackage[capitalize]{cleveref}
\crefname{section}{Sec.}{Secs.}
\Crefname{section}{Section}{Sections}
\Crefname{table}{Table}{Tables}
\crefname{table}{Tab.}{Tabs.}

\newcommand{\figref}[1]{Fig.~\ref{#1}}
\newcommand{\tabref}[1]{Tab.~\ref{#1}}
\newcommand{\secref}[1]{Sec.~\ref{#1}}
\newcommand{\AlgRef}[1]{Algorithm~\ref{#1}}
\newcommand{\eqnref}[1]{Eqn.~(\ref{#1})}
\newcommand{\Assumref}[1]{\textbf{Assumption}~\ref{#1}}

\def\cvprPaperID{330} 
\def\confName{CVPR}
\def\confYear{2023}
\def\sArt{state-of-the-art~}

\begin{document}

\title{Looking Through the Glass: Neural Surface Reconstruction Against \\ High Specular Reflections}
\author{
  Jiaxiong Qiu$^1$ ~~ Peng-Tao Jiang$^2$ ~~ Yifan Zhu$^1$ ~~ Ze-Xin Yin$^1$ ~~ Ming-Ming Cheng$^1$ ~~ Bo Ren$^1$\thanks{Bo Ren is the corresponding author.} \\
$^1$VCIP, CS, Nankai University \quad
$^2$State Key Lab of CAD\&CG, Zhejiang University \\
}
\maketitle
\begin{abstract}
Neural implicit methods have achieved high-quality 3D object surfaces under slight specular highlights. However, high specular reflections (HSR) often appear in front of target objects when we capture them through glasses. The complex ambiguity in these scenes violates the multi-view consistency, then makes it challenging for recent methods to reconstruct target objects correctly. To remedy this issue, we present a novel surface reconstruction framework, NeuS-HSR, based on implicit neural rendering. In NeuS-HSR, the object surface is parameterized as an implicit signed distance function (SDF). To reduce the interference of HSR, we propose decomposing the rendered image into two appearances: the target object and the auxiliary plane. We design a novel auxiliary plane module by combining physical assumptions and neural networks to generate the auxiliary plane appearance. Extensive experiments on synthetic and real-world datasets demonstrate that NeuS-HSR outperforms state-of-the-art approaches for accurate and robust target surface reconstruction against HSR. 
Code is available at \url{https://github.com/JiaxiongQ/NeuS-HSR}.
\end{abstract}

\section{Introduction}
\begin{figure}[t]
\centering
\includegraphics[width=\linewidth]{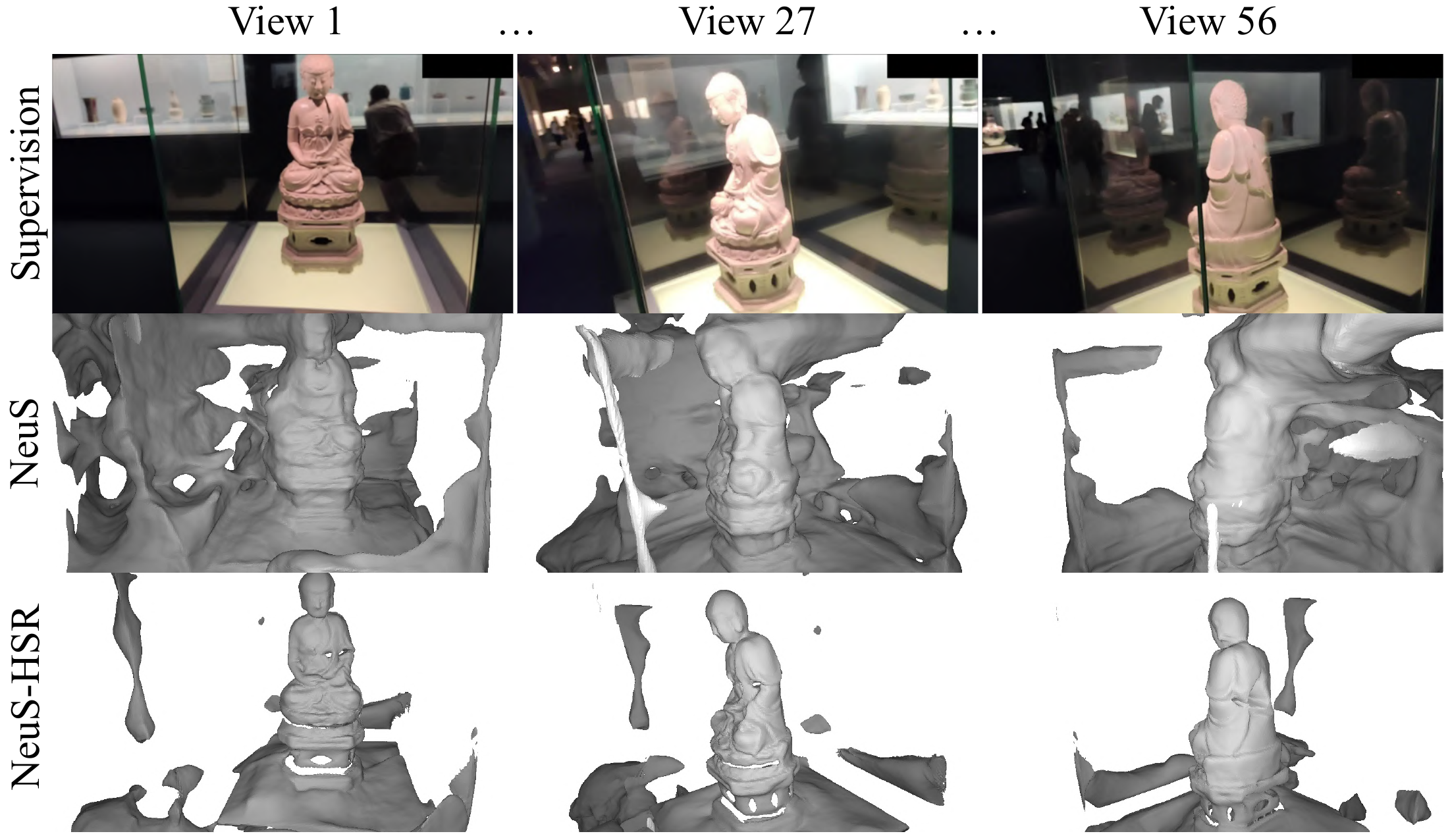}
\vspace*{-5mm}
\caption{3D object surface reconstruction under high specular reflections (HSR). Top: A real-world scene captured by a mobile phone. Middle: The \sArt method NeuS~\cite{wang2021neus} fails to reconstruct the target object (\ie, the Buddha). Bottom: We propose NeuS-HSR, which recovers a more accurate target object surface than NeuS.}
\vspace*{-5mm}
\label{hsr}
\end{figure}
Reconstructing 3D object surfaces from multi-view images is a challenging task in computer vision and graphics. 
%
Recently, NeuS~\cite{wang2021neus} combines the surface rendering~\cite{barnes2009patchmatch,galliani2016gipuma,schonberger2016pixelwise,yariv2020multiview} and volume rendering~\cite{de1999poxels,mildenhall2020nerf}, for reconstructing objects with thin structures and achieves good performance on the input with slight specular reflections.
However, when processing the scenes under high specular reflections (HSR), NeuS fails to recover the target object surfaces, 
as shown in the second row of \figref{hsr}.
High specular reflections are ubiquitous when we use a camera to capture the target object through glasses. 
As shown in the first row of \figref{hsr}, in the captured views with HSR, we can recognize the virtual image in front of the target object. 
The virtual image introduces the photometric variation on the object surface visually, which degrades the multi-view consistency and encodes extreme ambiguities for rendering, then confuses NeuS to reconstruct the reflected objects instead of the target object. 

%
%

\begin{figure}[t]
\centering
\includegraphics[width=\linewidth]{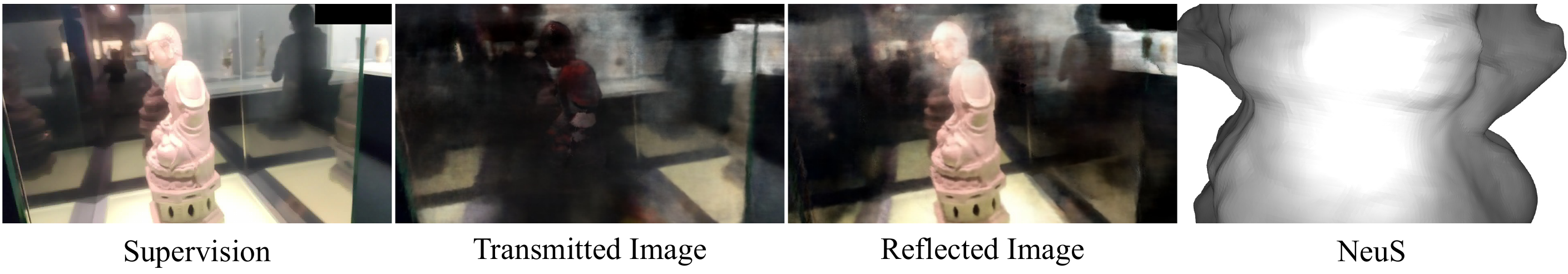}
\caption{Decomposition of NeRFReN~\cite{guo2022nerfren}. NeRFReN fails to separate specular reflections and the target object appearance in this view, then makes NeuS fail to recover the target object surface.}
\label{nerfren}
\vspace*{-5mm}
\end{figure}

To adapt to the HSR scenes, one intuitive solution is firstly applying reflection removal methods to reduce HSR, 
then reconstructing the target object with the enhanced target object appearance as the supervision. 
However, most recent single-image reflection removal works~\cite{lei2021robust, Dong_2021_ICCV,Chang_2021_WACV,liu2022adaptive,song2022multi,lei2022categorized} need 
the ground-truth background or reflection as supervision, which is hard to be acquired. 
Furthermore, for these reflection removal methods, testing scenes should be present in the training sets, which limits their generalization. 
These facts demonstrate that explicitly using the reflection removal methods to enhance the target object appearance is impractical. 
A recent unsupervised reflection removal approach, NeRFReN~\cite{guo2022nerfren} decomposes the rendered image into reflected and transmitted parts by implicit representations. However, it is limited by constrained view directions and simple planar reflectors. When we apply it to scenes for multi-view reconstruction, as \figref{nerfren} presents, it takes the target object as the content in the reflected image and fails to generate the correct transmitted image for target object recovery. 

\begin{figure}[t]
\centering
\includegraphics[width=\linewidth]{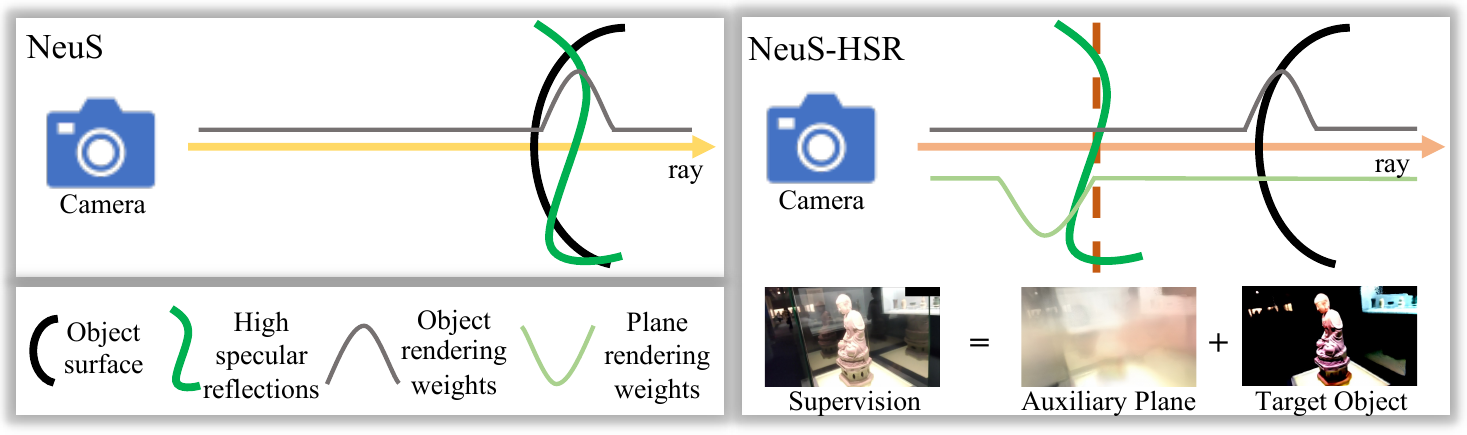}
\caption{NeuS-HSR. High specular reflections (HSR) make NeuS tend to reconstruct the reflected object in HSR. NeuS-HSR physically decomposes the rendered image into the target object and auxiliary plane parts, which encourages NeuS to focus on the target object.}
\vspace*{-5mm}
\label{moti}
\end{figure}

The two-stage intuitive solution struggles in our task as discussed above. To tackle this issue, we consider a more effective decomposition strategy than NeRFReN, to enhance the target object appearance for accurate surface reconstruction in one stage. To achieve our goal, we construct the following assumptions:
\begin{assumption}\label{assum_1}
A scene that suffers from HSR can be decomposed into the target object and planar reflector components. Except for the target object, HSR and most other contents in a view are reflected and transmitted through the planar reflectors (\ie, glasses).
\end{assumption}
\begin{assumption}\label{assum_2}
Planar reflectors intersect with the camera view direction since all view direction vectors generally point to the target object and pass through planar reflectors.
\end{assumption}

Based on the above physical assumptions, we propose NeuS-HSR, a novel object reconstruction framework to recover the target object surface against HSR from a set of RGB images. 
For \Assumref{assum_1}, as \figref{moti} shows, we design an auxiliary plane to represent the planar reflector since we aim to enhance the target object appearance through it. With the aid of the auxiliary plane, we faithfully separate the target object and auxiliary plane parts from the supervision. 
For the target object part, we follow NeuS~\cite{wang2021neus} to generate the target object appearance. 
For the auxiliary plane part, we design an auxiliary plane module with the view direction as the input for \Assumref{assum_2}, by utilizing neural networks to generate attributes (including the normal and position) of the view-dependent auxiliary plane. When the auxiliary plane is determined, we acquire the auxiliary plane appearance based on the reflection transformation~\cite{goldman1990matrices} and neural networks. 
Finally, we add two appearances and then obtain the rendered image, which is 
supervised by the captured image for one-stage training.

We conduct a series of experiments to evaluate NeuS-HSR. The experiments demonstrate that NeuS-HSR is superior to other \sArt methods on the synthetic dataset and recovers high-quality target objects from HSR-effect images in real-world scenes. 

To summarize, our main contributions are as follows:
\begin{itemize}[leftmargin=*] 
\item We propose to recover the target object surface, which suffers from HSR, by separating the target object and auxiliary plane parts of the scene.
\item We design an auxiliary plane module to generate the appearance of the auxiliary plane part physically to enhance the appearance of the target object part.
\item Extensive experiments on synthetic and real-world scenes demonstrate that our method reconstructs more accurate target objects than other \sArt methods quantitatively and qualitatively.
\end{itemize}

\begin{figure*}[t]
\centering
\includegraphics[width=\textwidth]{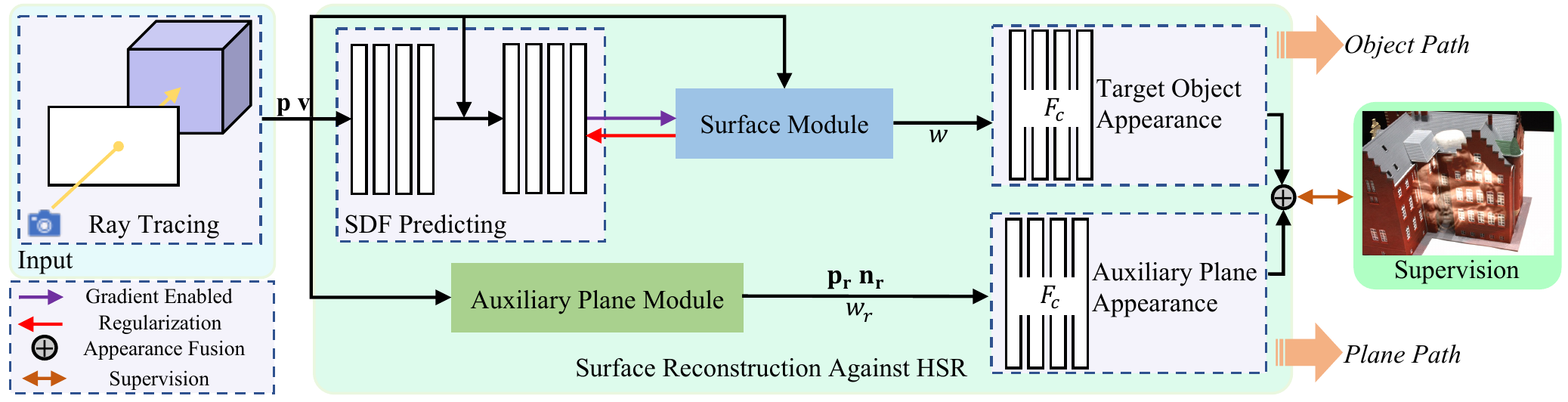}
\caption{Pipeline of NeuS-HSR. The sampled points $\mathbf{p}$ and the view direction $\mathbf{v}$ are fed into the target object path and the plane path respectively. 
In the object path, the implicit SDF $f$ is generated by the head neural networks. The surface module takes $f$, $\mathbf{p}$, and $\mathbf{v}$ as the input, producing the rendering weights $w$. In the plane path, the reflection module generates the plane normal $\mathbf{n}_r$, 3D locations $\mathbf{p}_r$, and the rendering weights $w_r$ of the auxiliary plane from $\mathbf{p}$ and $\mathbf{v}$. Finally, we acquire two appearances by the appearance function $F_c$ and volume rendering.}
\label{pip}
\vspace*{-5mm}
\end{figure*}

\section{Related Works}
\subsection{Traditional Surface Reconstruction}
The classical multi-view surface reconstruction methods mainly consist of two categories: photometric stereo~\cite{hernandez2008multiview,cheng2021multi,cheng2022diffeomorphic,yang2022s} and multi-view stereo~\cite{seitz1999photorealistic, seitz2006comparison, goesele2006multi, furukawa2009accurate, godard2015multi, schoenberger2016mvs,goel2022differentiable} reconstruction. 
The photometric stereo reconstruction is limited by the strict experimental environment. For the multi-view stereo reconstruction, the input images are captured around the target object in common scenes. 
The early multi-view stereo methods~\cite{seitz2006comparison,goesele2006multi,furukawa2009accurate,schoenberger2016mvs} focus on the object surfaces with diffuse materials. They all obey the Lambertian assumption that the same detected region of the object surfaces has little change in all views. However, obvious specular reflections often occur on objects in real-world scenes, \eg, the highlight. The Lambertian assumption no longer holds in real-world scenes with obvious specular reflections. 
The widely-used Structure From Motion (SFM) methods~\cite{ullman1979interpretation, wu2011visualsfm, schonberger2016structure} is designed to calibrate the camera and produce a sparse depth map at each viewpoint firstly. 
Then the object surface can be acquired by Poisson Surface Reconstruction~\cite{kazhdan2013screened} with depth fusion. However, the quality of the output surface is easily affected by the feature point detection, and surface areas without rich textures on the target object always have artifacts or empty holes. In this work, we focus on the neural implicit method to achieve accurate 3D object surfaces in more realistic scenarios (\ie, non-Lambertian surfaces). 

\subsection{Neural Implicit Surface Rendering}
Implicit representations based on neural networks have achieved promising results on novel view synthesis~\cite{lombardi2019neural, sitzmann2019scene, mildenhall2020nerf,wei2021nerfingmvs,lin2021barf} and 3D reconstruction~\cite{niemeyer2020differentiable, yariv2020multiview, oechsle2021unisurf, wang2021neus, yariv2021volume, sun2022neural, worchel2022multi, darmon2022improving,fu2022geo,rasmuson2022addressing,wang2022neuris,wanghf}. 
They have characteristics that classical methods do not have, including flexible resolution and natural global consistency. 
Surface rendering based on the differentiable ray casting is applied for surface reconstruction with different forms of implicit shape representations, 
such as the occupancy function~\cite{peng2020convolutional} and signed distance function (SDF)~\cite{yariv2020multiview}. 
IDR~\cite{yariv2020multiview} extracts surface points on the object surface with the zero-level set of SDF representations, 
and utilizes neural network gradients to solve a differentiable rendering formulation. 
UNISURF~\cite{oechsle2021unisurf}, VolSDF~\cite{yariv2021volume} and NeuS~\cite{wang2021neus} learn the implicit surface function by introducing the volume rendering scheme~\cite{mildenhall2020nerf}, 
to improve the surface reconstruction quality from captured images. 
NeuralWarp~\cite{darmon2022improving} is a two-stage method for refining the basic model (\eg, VolSDF).
NeRS~\cite{zhang2021ners} focuses on learning the appearance of object surfaces by introducing the Phong model~\cite{kajiya1986rendering, immel1986radiosity,verbin2022ref}.
It uses a canonical sphere to represent the object surface and learns the object texture with prerequisite masks from a sparse set of images, 
but it mainly deals with objects with reflective surfaces and produces the object surface without fine details. 
In contrast with these works, we propose to extend the object surface reconstruction to more challenging HSR scenes in one stage. We aim to correctly recover the object surface through glasses instead of the reflective surface. Our method achieves much better reconstruction accuracy and robustness than previous works in HSR scenes. 

\section{Method}
In this work, we focus on reconstructing the object surfaces in high specular reflection (HSR) scenes. 
As mentioned in the introduction section, HSR encodes non-target object information, resulting in a low-quality target object surface.
To tackle HSR scenes, we introduce a novel object surface reconstruction method, NeuS-HSR, which is based on the 
implicit neural rendering.
The pipeline of NeuS-HSR is shown in \figref{pip}.

Specifically, we decompose an HSR scene into two components: the target object and the auxiliary plane. To render the target object appearance, we adopt the scheme of NeuS and pack it as a surface module. To render the auxiliary plane appearance, we design an auxiliary plane module based on the reflection transformation~\cite{goldman1990matrices} and multi-layer perceptrons (MLPs).
Finally, we apply a linear summation to fuse two appearances to obtain the rendered image, which receives supervision from the captured image 
in a view.
In the following, we introduce NeuS-HSR in three parts, 
including the surface module (\secref{section:sm}), the auxiliary plane module (\secref{section:apm}), and the rendering process (\secref{section:render}).

%

\subsection{Surface Module}~\label{section:sm}
We apply NeuS~\cite{wang2021neus} to render the target object appearance.
Specifically, NeuS builds an unbiased and occlusion-aware weight function $w$ based on the implicit SDF $f: \mathbb{R}^{3}\to\mathbb{R}$ on each camera ray $\mathbf{h}_s$. Firstly, $w$ is defined as: 
\begin{equation}\label{eq1}
\setlength{\abovedisplayskip}{3pt}
\setlength{\belowdisplayskip}{3pt}
w(t) = T(t)\rho(t),\\
T(t) = \mathrm{exp}\left ( -\int_{0}^{t}\rho(u)du \right ).
\end{equation}
where $t\in\mathbb{R}$ is the depth value along $\mathbf{h}_s$, then $\rho(t)$ is constructed by:
\begin{equation}\label{eq2}
\setlength{\abovedisplayskip}{3pt}
\setlength{\belowdisplayskip}{3pt}
\rho(t) = \mathrm{max} \left (\frac{-\frac{d\Theta_{s}}{dt}(f(\mathbf{p}(t)))}{\Theta_{s}(f(\mathbf{p}(t)))}, \mathrm{0} \right).
\end{equation}
where the object surface $S$ can be modeled by a zero-level set of the signed distance at the point $\mathbf{p}$: $S = \{\mathbf{p}\in\mathbb{R}^{3}|f(\mathbf{p})=\mathrm{0}\}$. The logistic density distribution $\theta_{S}(\mathbf{p}) = se^{-s\mathbf{p}}/(1+e^{-s\mathbf{p}})^2$, which is the derivative of the Sigmoid function $\Theta_s(\mathbf{p}) = (1+e^{-s\mathbf{p}})^{-1}$. $1/s$ is the standard deviation of $\theta_{S}(\mathbf{p})$.

The construction of $w$ is the key contribution of NeuS. It connects the implicit SDF and the volume rendering properly to handle complex object structures. The camera ray $\mathbf{h}_s$ at point $\mathbf{p}$ can be denoted as: $\mathbf{h}_s(t)=\mathbf{o}+t\mathbf{v}$, where $\mathbf{o}$ and $\mathbf{v}$ represent the camera center and view direction separately. We sample $m$ points along $\mathbf{h}_s$, then the pixel color value $C$ is acquired by follows:
\begin{equation}\label{eq3}
\setlength{\abovedisplayskip}{3pt}
\setlength{\belowdisplayskip}{3pt}
C = \sum_{i=1}^m w_i c_i.
\end{equation}
where $w_i = \alpha_i\prod_{j=1}^{i-1}(1-\alpha_j)$, $c_i$ is the learned color from MLPs, and $\mathrm{\alpha}_i$ is the discretization of \eqnref{eq2}.

In our task, $C$ encodes the content of HSR, which causes the ambiguity of $w$. 
This ambiguity makes MLPs of SDF predicting tend to model the content of HSR, producing excessive noise around the target object surface. 
Thus, we propose an auxiliary plane module to divert the attention of MLPs to the target object to handle the interference of HSR.

\subsection{Auxiliary Plane Module}~\label{section:apm}
\begin{figure}[t]
\centering
\includegraphics[width=\linewidth]{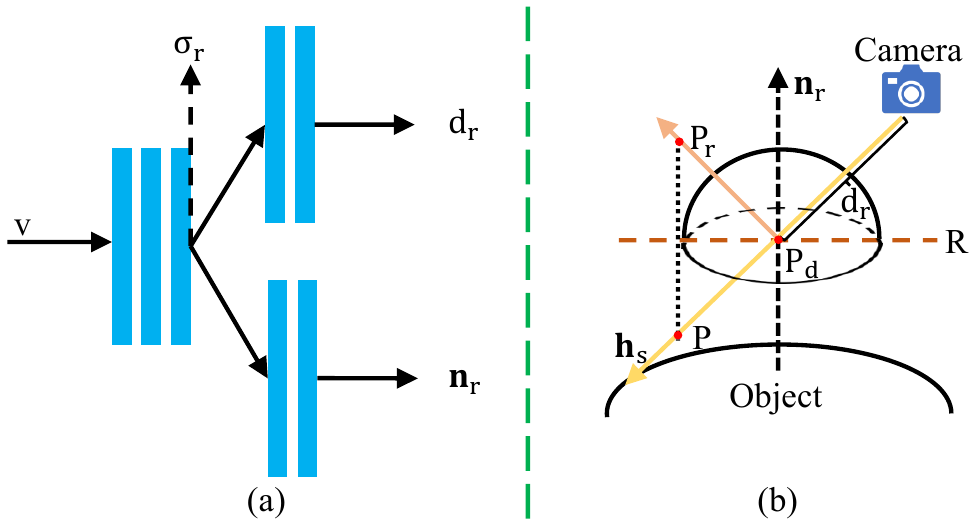}
\vspace*{-7mm}
\caption{Auxiliary plane module. (a): We design a novel neural network $F_r$ to predict the volume density $\mathrm{\sigma_r}$, position $\mathrm{d_r}$ and surface normal $\mathbf{n}_r$ of the auxiliary plane $\mathrm{R}$ on the camera ray $\mathbf{h}_s$ from the input view direction $\mathbf{v}$. (b): When the auxiliary plane is determined, we project the sampled point $\mathrm{P}$ behind $\mathrm{R}$ to its reflected point $\mathrm{P_r}$ by the reflection transformation~\cite{goldman1990matrices}.}
\label{ref_m} 
\vspace*{-5mm}
\end{figure}
In HSR scenes, the virtual images which appear on the planar reflectors and in front of the target object encode extremely ambiguous information for the target object reconstruction. Decomposing the rendered image for enhancing the target object appearance without prior information is an ill-posed problem. 
NeRFReN~\cite{guo2022nerfren} applies a depth smoothness prior and a bidirectional depth consistency constraint, to split the rendered image into two components: the transmitted image and the reflected image by implicit representations. This scheme works well in scenes with limited view directions and simple planar reflectors. However, it fails to preserve the target object in transmitted images in HSR scenes.
Motivated by NeRFReN, we propose an auxiliary plane module to enhance the target object appearance in the rendered image. 

Formally, we use an auxiliary plane $\mathrm{R}$ to represent the actual planar reflector for each camera ray. To determine $\mathrm{R}$ physically, we design a novel neural network $F_r: \mathbb{S}^2 \rightarrow \mathbb{R} \times \mathbb{R} \times \mathbb{R}^3$ as shown in \figref{ref_m}~(a).
$F_r$ maps the view direction $\mathbf{v}$ to the volume density $\mathrm{\sigma_r}$ for generating the rendering weights, and attributes of $\mathrm{R}$ (including the position $\mathrm{d_r}$ and the plane normal $\mathbf{n}_r$), that is:
\begin{equation}\label{eq4_1}
\setlength{\abovedisplayskip}{3pt}
\setlength{\belowdisplayskip}{3pt}
\{\mathrm{\sigma_r, d_r}, \mathbf{n}_r\} = F_r(\mathbf{v}). 
\end{equation}
We assume $\mathrm{R}$ is built in the camera coordinate system, the 3D point $\mathrm{p_d=d_r}\mathbf{v}$ is the interaction of $\mathrm{R}$ and $\mathbf{h}_s$. Then $\mathrm{p_d}$ is on $\mathrm{R}$ obviously. Given $\mathbf{n}_r=[A,B,C]$, $\mathrm{R}$ can be defined as:
\begin{equation}\label{eq4}
\setlength{\abovedisplayskip}{3pt}
\setlength{\belowdisplayskip}{3pt}
\mathrm{Ax+By+Cz+D=0}. 
\end{equation}
where $\mathrm{A^2+B^2+C^2=1}$. We substitute $\mathrm{P_d}$ into \eqnref{eq4} then have $\mathrm{D=-d_r}\mathbf{n}_r\cdot\mathbf{v}$. 
\begin{figure}[t]
\centering
\includegraphics[width=\linewidth]{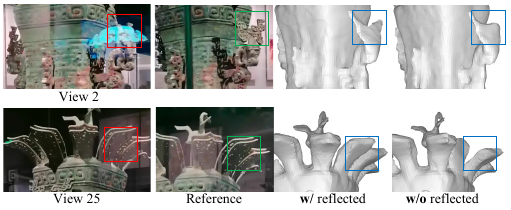}
\vspace*{-7mm}
\caption{Effect of projecting sampled points to reflected points. \textcolor[rgb]{1,0,0}{Red} boxes show the regions of HSR. \textcolor[rgb]{0,1,0}{Green} boxes show the reference regions related to red boxes. \textcolor[rgb]{0,0,1}{Blue} boxes show effects of HSR on object surface reconstruction.}
\label{reflect}
\vspace*{-5mm}
\end{figure}

Moreover, the sampled points along camera rays are part of the inputs for acquiring color values by MLPs. To further model HSR physically, 
as shown in \figref{ref_m}~(b), for a point $\mathrm{p}$ sampled along $\mathbf{h}_s$ and behind $\mathrm{R}$, we project it to the reflected point $\mathrm{p_r}$ along the incident light path based on the reflection transformation~\cite{goldman1990matrices}. 
Then MLPs can implicitly trace the incident light to render HSR. \figref{reflect} demonstrates the effectiveness of this operation in reducing the interference of HSR. Reflected points help MLPs physically model HSR, then reduce the ambiguity of the scene and recover more accurate target object surfaces. The details of our projection algorithm are explained in supplementary materials.

\subsection{Rendering}~\label{section:render}
\begin{figure}[t]
\centering
\includegraphics[width=\linewidth]{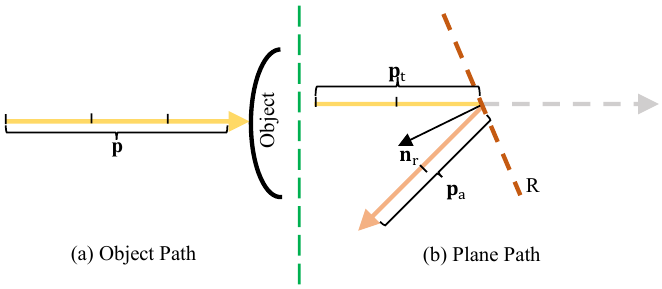}
\vspace*{-7mm}
\caption{Spatial sampled points along camera rays are used for acquiring color values of two paths. For the object path, we use the original sampled points $\mathbf{p}$. $\mathbf{p'}$ are $\mathbf{p}$ in the camera coordinate system. For the plane path, we use the part points $\mathbf{p}_t$ of $\mathbf{p'}$ which are in front of $\mathrm{R}$ along the camera ray and reflected points $\mathbf{p}_a$.}
\label{fusion}
\vspace*{-5mm}
\end{figure}

We adopt the neural network $F_c$ for predicting color values of the object path $c_t$ and the plane path $c_a$ separately. 
The input of each path is different. For the object path, we follow NeuS and utilize the sampled points $\mathbf{p}$ along the camera ray, surface normal $\mathbf{n}$ of the target object, the view direction $\mathbf{v}$ and features $\mathrm{f_p}$ of the implicit SDF as input. 
Then we have $c_t = F_c(\mathbf{p},\mathbf{n},\mathbf{v},\mathrm{f_p})$. 
For the plane path, the sampled points in the camera coordinate system are formed as: $\mathbf{p'}=\mathbf{p}-\mathbf{o}$. 
As \figref{fusion} shows, we utilize both the part points $\mathbf{p}_t$ of $\mathbf{p'}$ in front of $\mathrm{R}$ and the reflected points $\mathbf{p}_a$ as the input points $\mathbf{p}_r = {\mathbf{p}_t \cup \mathbf{p}_a}$. 
We utilize the plane normal $\mathbf{n}_r$ as input normal.
Then for the plane path, we have $c_r = F_c(\mathbf{p}_r,\mathbf{n}_r,\mathbf{v},\mathrm{f_p})$.
 
To generate the rendered appearance of each path, we also need to construct two weights of rendering. For the object path, We follow the scheme of NeuS to produce weights $w$ as defined in~\ref{section:sm}. 
For the auxiliary plane path, given the volume density $\mathrm{\sigma_r}$ learned from 
the plane network $F_r$, we adopt the scheme of NeRFReN to generate weights $w_r$ by:
\begin{equation}\label{eq5}
\setlength{\abovedisplayskip}{3pt}
\setlength{\belowdisplayskip}{3pt}
w_r^i = \mathrm{exp}(-\sum_{j=1}^{i-1}\sigma_r^j\delta_j)(1-\mathrm{exp}(-\sigma_r^i\delta_i)).
\end{equation}
where $\delta_i=t_{i+1}-t_i$.
Finally, the target object appearance $C_t(w,c_t)$ and auxiliary plane appearance $C_r(w_r,c_r)$ can be generated by \eqnref{eq3}. The final rendered image $C$ is obtained by a linear combination of $C_t$ and $C_r$, which is formulated by:
\begin{equation}\label{eq6}
\setlength{\abovedisplayskip}{3pt}
\setlength{\belowdisplayskip}{3pt}
C = \mathrm{\varphi}_1 C_t + \mathrm{\varphi}_2 C_r.
\end{equation}
where $\mathrm{\varphi_1 + \varphi_2 = 1}$. In practice, we set $\mathrm{\varphi_1=0.3}$ and $\mathrm{\varphi_2=0.7}$ by default. The details of this setting are illustrated in the supplementary material.

\subsection{Loss Function}~\label{section:loss}
During the training procedure of NeuS-HSR, we optimize the difference between the rendered image $C$ and the captured image $\tilde{C}$. We follow the loss function defined in NeuS, which consists of three terms: the color loss $\mathcal{L}_c$~\cite{yariv2020multiview,wang2021neus}, the regularized loss $\mathcal{L}_r$~\cite{gropp2020implicit} of the implicit SDF 
and $\mathcal{L}_n$ of the plane normal. 
The loss functions are formulated as follows:
\begin{equation}\label{eq7}
\setlength{\abovedisplayskip}{3pt}
\setlength{\belowdisplayskip}{3pt}
\left\{  
\begin{aligned}
\mathcal{L}_c &= \frac{1}{b} \sum_i \mathcal{L}_1 (C_i,\tilde{C}_i), \\
\mathcal{L}_r &= \frac{1}{bm} \sum_{k,i} (|\nabla f(\mathbf{p}_k^i)|-1)^2, \\
\mathcal{L}_n &= \frac{1}{b} \sum_{i} (|\mathbf{n}_r^i|-1)^2,
\end{aligned}
\right.
\end{equation}
where $b$ denotes the batch size and $m$ denotes the number of sampled points along a camera ray.
Then the final loss function can be defined as:
\begin{equation}\label{eq8}
\setlength{\abovedisplayskip}{3pt}
\setlength{\belowdisplayskip}{3pt}
\mathrm{\mathcal{L} = \mathcal{L}_c + \lambda_1 (\mathcal{L}_r} + \mathrm{\mathcal{L}_n}).
\end{equation}
where $\mathrm{\lambda_1}$ is a constant. Practically, we set $\mathrm{\lambda_1=0.1}$ by default. 
\begin{figure}[t]
\centering
\includegraphics[width=\linewidth]{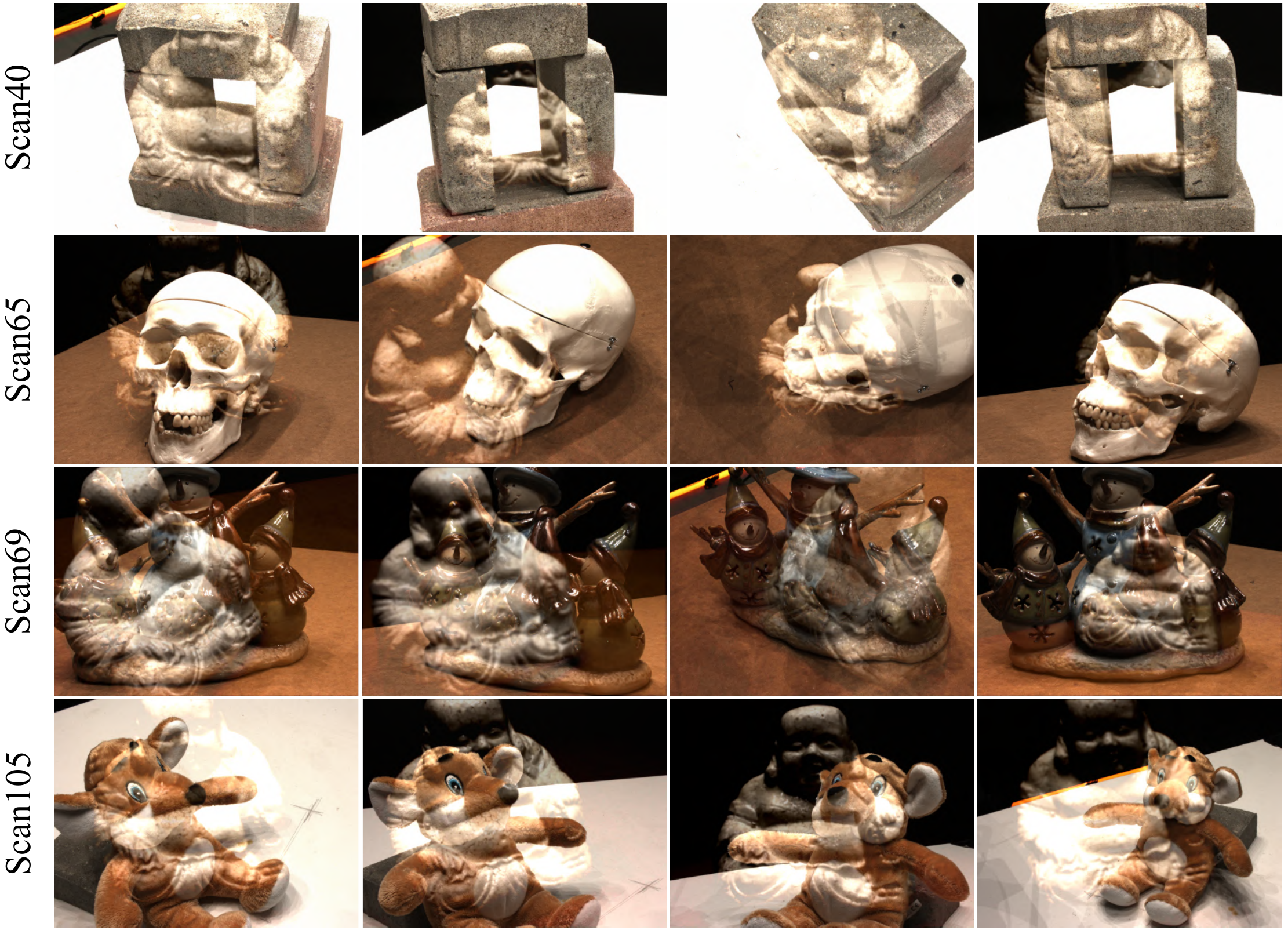}
\vspace*{-7mm}
\caption{Examples of the synthetic dataset. We apply the widely-used method~\cite{zhang2018single} to synthesize HSR scenes on the DTU dataset~\cite{aanaes2016large}.}
\label{syn}
\end{figure}

\begin{figure*}[t]
\centering
\includegraphics[width=\textwidth]{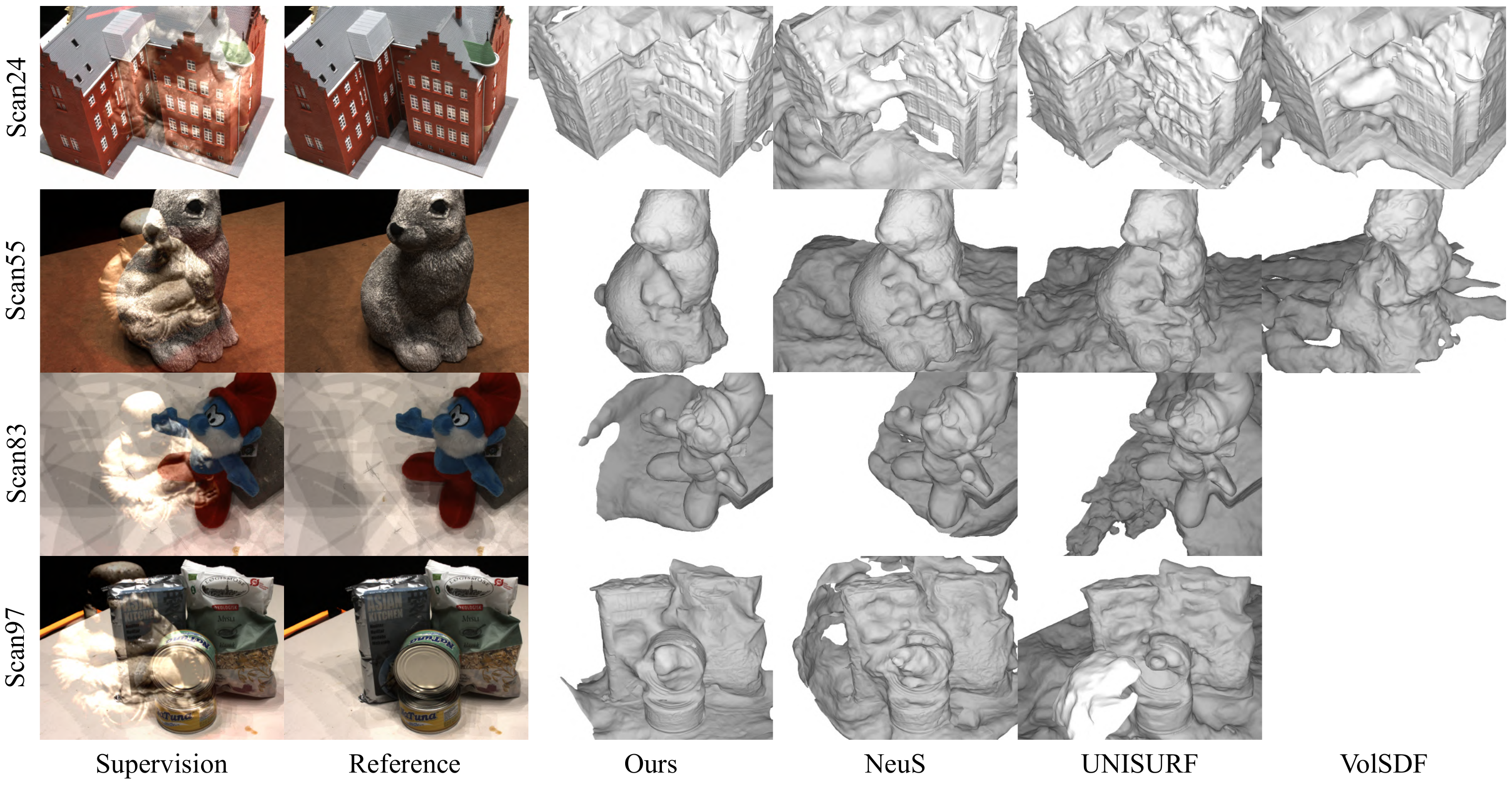}
\vspace*{-7mm}
\caption{Qualitative comparisons on the synthetic dataset. `Reference' contains the original appearance of the target object. VolSDF~\cite{yariv2021volume} fails to recover the meaningful object surfaces on `Scan83' and `Scan97'. Our method achieves better object surface quality when compared to other methods.}
\vspace*{-5mm}
\label{comp1}
\end{figure*}

\section{Experiments}
We conduct extensive experiments which show our method outperforms other approaches quantitatively (\tabref{compt}) and qualitatively (\figref{comp1}, \figref{comp3}). 
We also provide several ablation experiments to reveal the necessity of our design choices (\figref{ab1}).
\subsection{Datasets}
\myPara{Synthetic dataset.}
To evaluate the performance of NeuS-HSR and other methods quantitatively, we synthesize $\mathrm{10}$ scenes from the DTU dataset~\cite{aanaes2016large}. We follow the common single-image reflection synthesis method~\cite{zhang2018single} to generate the synthetic dataset. Given the transmission image $\mathrm{T}$ (\ie, the image which contains the target object) 
and the reflection image $\mathrm{R'}$, the image $\mathrm{I}$ with reflections can be defined as: 
\begin{equation}\label{eq9}
\setlength{\abovedisplayskip}{3pt}
\setlength{\belowdisplayskip}{3pt}
\mathrm{I = T + \mathcal{K} \otimes R'}.
\end{equation}
$\mathcal{K}$ is a Gaussian kernel, and $\mathrm{\otimes}$ means the convolutional operation. 
We randomly select a scene as the reflection part, and other scenes are set as the transmission part. 
Then we adopt \eqnref{eq9} to acquire HSR scenes.
Examples of the synthetic dataset are shown in \figref{syn}.

\myPara{Real-world dataset.}
To validate the effectiveness of our method in real-world scenes, we collect $\mathrm{6}$ HSR scenes from the Internet. We utilize the widely-used tool, COLMAP~\cite{schoenberger2016mvs}, 
to estimate camera parameters.

\subsection{Settings}
\myPara{Implementation details.} 
The signed distance function (SDF) $f$ is parameterized by MLPs, 
which consists of $\mathrm{8}$ linear layers. Then the target object surface is produced from the implicit SDF by marching cubes~\cite{lorensen1987marching}. The auxiliary plane function $F_r$ consists of $\mathrm{3}$-layer MLPs for predicting the volume density and $\mathrm{2}$-layer MLPs for predicting the plane attributes. The rendering appearance function $F_c$ is modeled by 
$\mathrm{4}$-layer MLPs. 
All spatial points are sampled inside a unit sphere, where the scene outside is produced by NeRF++~\cite{zhang2020nerf++}. 
Positional encoding~\cite{mildenhall2020nerf} is adopted to sampled points $\mathbf{p}$ along camera rays and view directions $\mathbf{v}$. The approximate SDF is pre-processed by the geometric initialization~\cite{atzmon2020sal}. 
The batch size of rays is set to $\mathrm{512}$. 
We train NeuS-HSR for $\mathrm{200k}$ iterations, consuming about $\mathrm{12}$ hours on a single $\mathrm{NVIDIA~Tesla~V100~GPU}$.

\begin{table}[t]
\footnotesize
\centering
\caption{Quantitative results on the synthetic dataset by measuring the Chamfer distance. VolSDF~\cite{yariv2021volume} fails to recover the meaningful object surfaces in the last $\mathrm{7}$ scenes. Our method outperforms other methods on average. The best metrics are \textbf{highlighted}.}
\vspace*{-3mm}
\renewcommand{\arraystretch}{1.10}
\setlength{\tabcolsep}{2.0mm}{
\begin{tabular}{l||c|c|c|c} \toprule[1pt]
ScanID & UNISURF~\cite{oechsle2021unisurf} & VolSDF~\cite{yariv2021volume} & NeuS~\cite{wang2021neus} & Ours \\
\midrule[0.8pt]
scan24 & 2.92 & 3.89 & 5.30 & \textbf{2.07} \\ 
scan37 & 4.26 & 2.91 & 2.29 & \textbf{1.89} \\
scan40 & 3.36 & 2.44 & \textbf{2.02} & 2.17 \\
scan55 & 2.11 & 3.95 & 1.73 & \textbf{1.25} \\
scan63 & 2.73 & - & 2.75 & \textbf{1.94} \\
scan65 & 1.57 & - & \textbf{0.93} & 1.15 \\
scan69 & 5.00 & - & 4.15 & \textbf{3.54} \\
scan83 & 1.81 & - & 2.55 & \textbf{1.42} \\
scan97 & 3.85 & - & 4.62 & \textbf{2.82}\\
scan105 & 2.01 & - & 1.53 & \textbf{1.31} \\
\midrule[0.8pt]
mean & 2.96 & / & 2.79 & \textbf{1.96}  \\
\bottomrule[1pt]
\end{tabular}}
\label{compt}
\vspace*{-5mm}
\end{table}

\begin{figure*}[t]
\centering
\includegraphics[width=\linewidth]{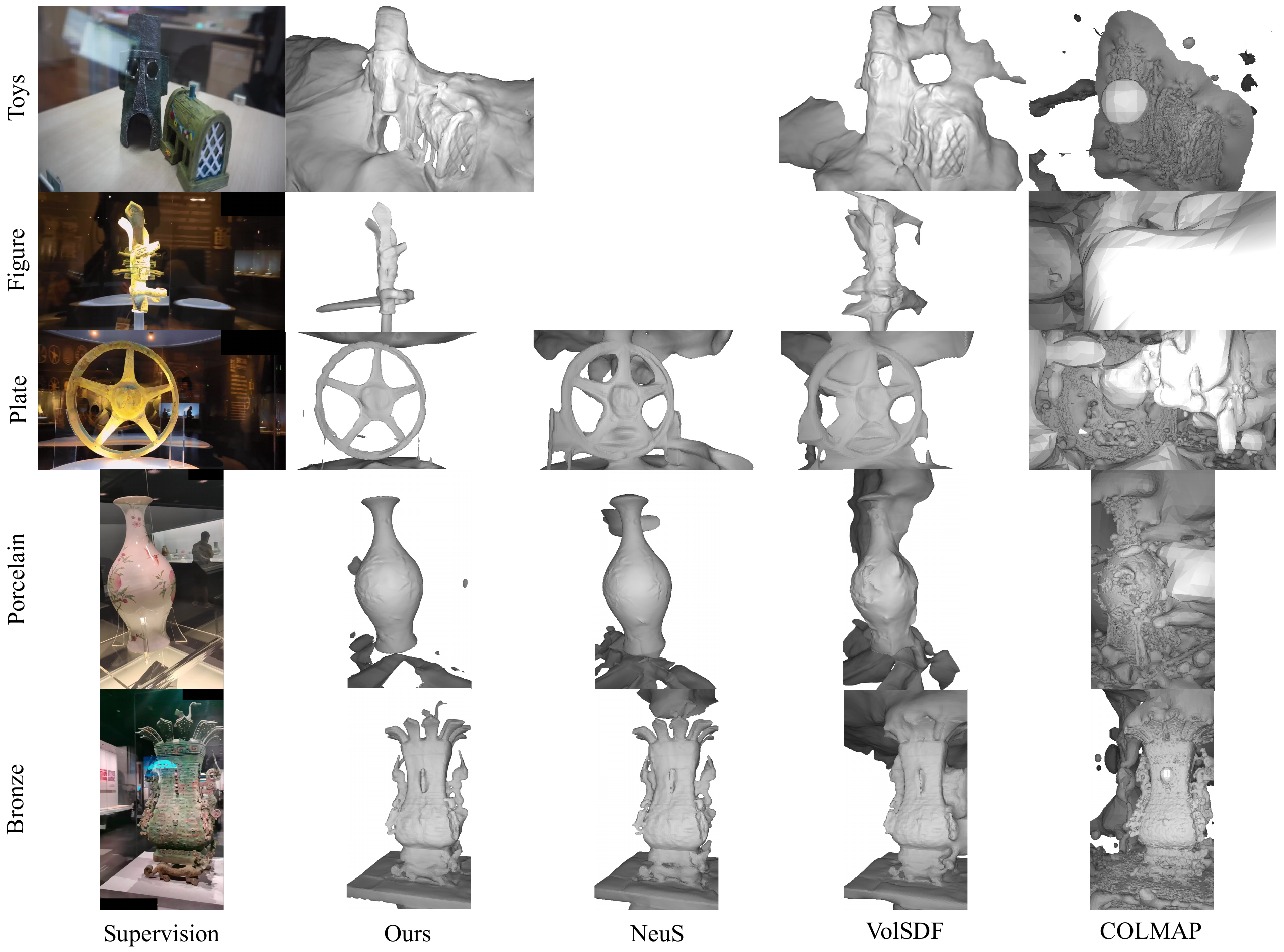}
\vspace*{-7mm}
\caption{Qualitative comparisons on the real-world dataset. NeuS~\cite{wang2021neus} fails to recover the meaningful object surfaces on `Toys' and `Figure'. Our method recovers target objects against HSR with physically correct surfaces, while other methods generate noisy results.}
\label{comp3}
\vspace*{-1mm}
\end{figure*}

\begin{figure*}[t]
\centering
\includegraphics[width=\textwidth]{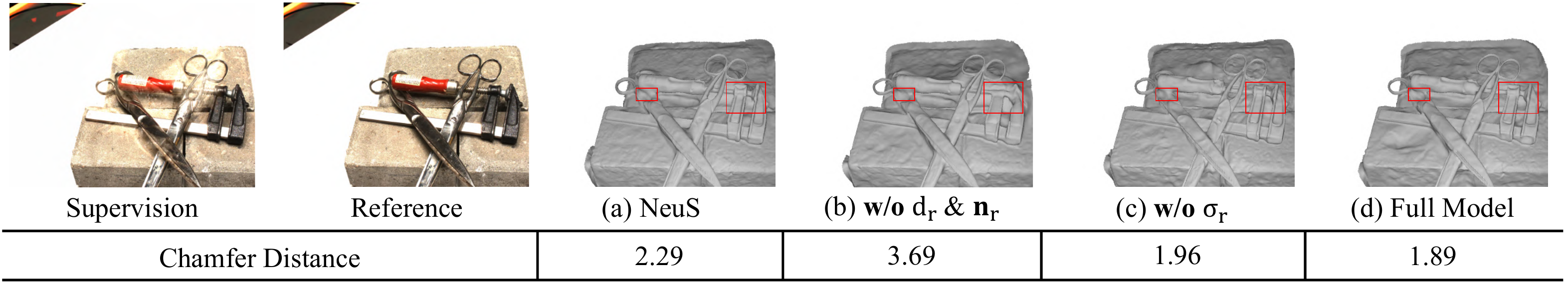}
\vspace*{-7mm}
\caption{Ablation study on a synthetic scene. When decomposing the scene without attributes ($\mathrm{d_r}$ and $\mathbf{n}_r$) of auxiliary planes, the performance degrades remarkably. The recovered target object surface loses fine details while disabling the volume density $\mathrm{\sigma_r}$.}
\label{ab1}
\vspace*{-5mm}
\end{figure*}
\myPara{Compared Methods.}
We compare our method against other related approaches with fair settings. 
The related approaches includes (i) \sArt neural implicit surface reconstruction approaches: NeuS~\cite{wang2021neus}, VolSDF~\cite{yariv2021volume} and UNISURF~\cite{oechsle2021unisurf}, (ii) the classical multi-view stereo method: COLMAP~\cite{schoenberger2016mvs}.
For COLMAP, we apply Screened Poisson~\cite{kazhdan2013screened} to 
reconstruct its dense mesh from the estimated point cloud. 
All learning-based models in this paper are trained without ground-truth masks.

\subsection{Quantitative Comparisons}
For the quantitative evaluation, we conduct comparisons on the synthetic dataset.
Following \cite{wang2021neus,yariv2021volume,oechsle2021unisurf}, we utilize the Chamfer distance as the evaluation metric, which represents the reconstruction quality of the target object. 
We report the metric scores in \tabref{compt}. 
It can be seen that the proposed method is better than other methods in most scenes. 
As a result, NeuS-HSR surpasses other methods in terms of the average Chamfer distance. 
\begin{figure*}[t]
\centering
\includegraphics[width=\textwidth]{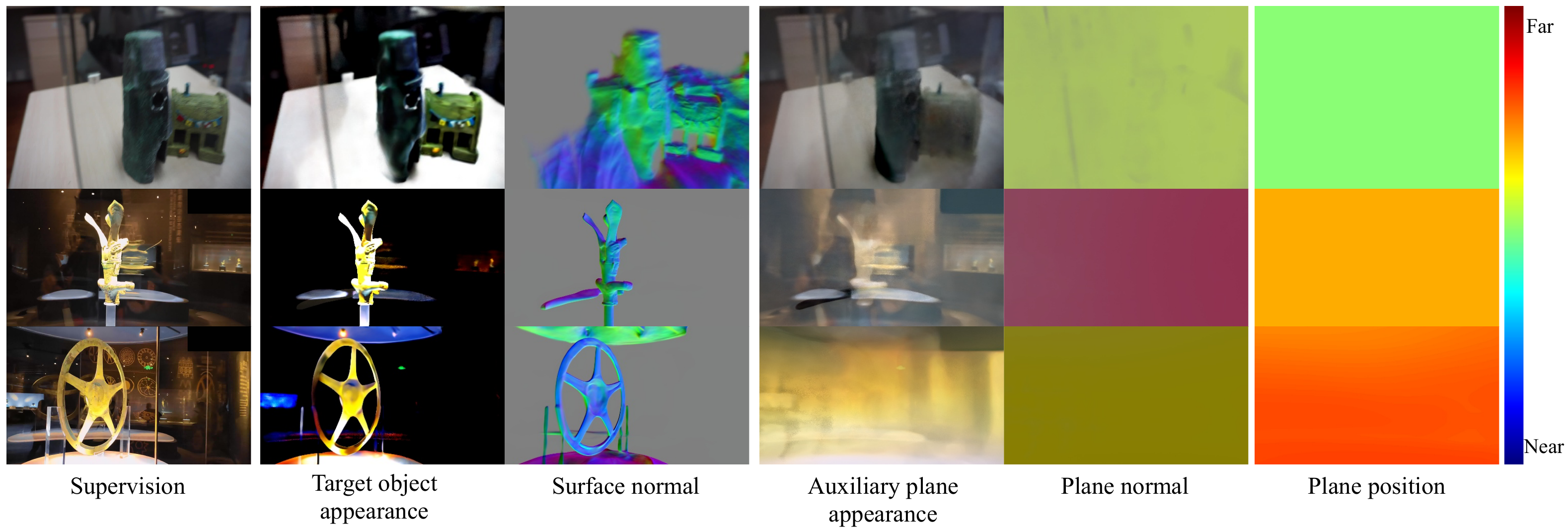}
\vspace*{-5mm}
\caption{Components of NeuS-HSR. The rendered image of NeuS-HSR is decomposed into two appearances: the target object and the auxiliary plane. Target object appearances encode complete target objects with high sharpness, and auxiliary plane appearances enhance the content of HSR. The color bar represents the color map of plane position (`Near' means the auxiliary plane is close to the camera).}
\label{ab2}
\vspace*{-3mm}
\end{figure*}

\begin{figure}[t]
\centering
\includegraphics[width=\linewidth]{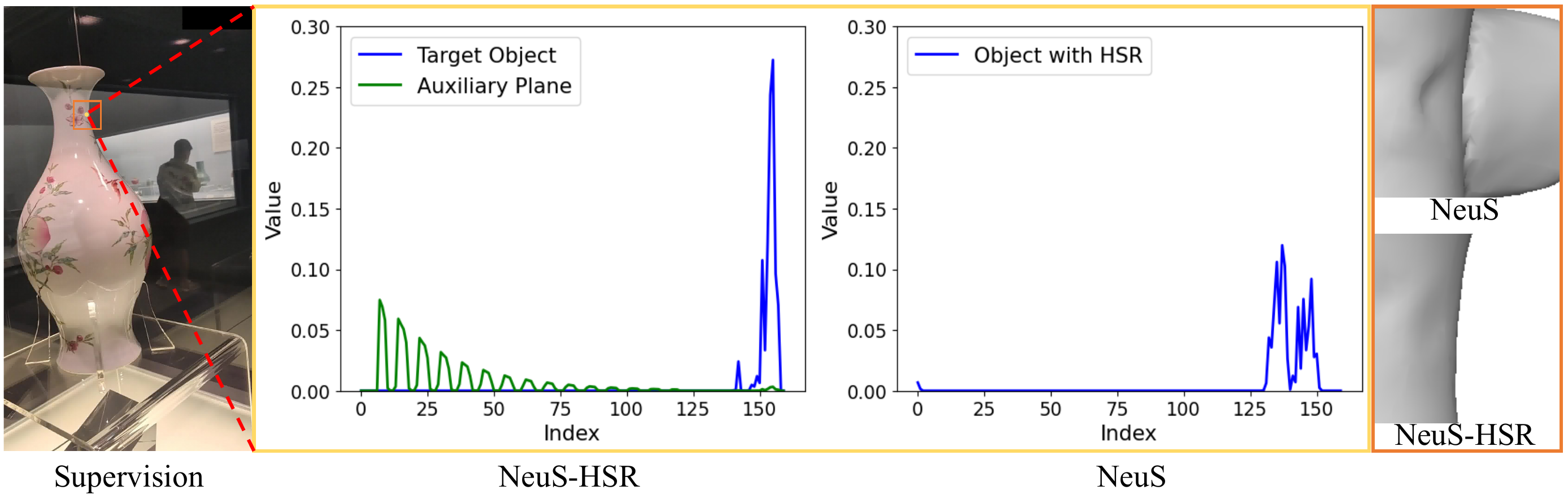}
\vspace*{-5mm}
\caption{Rendering weights in a real-world HSR scene. NeuS-HSR enables NeuS to pay more attention to the target object against HSR, as the cropped meshes on the right show.}
\label{weights}
\vspace*{-3mm}
\end{figure}
\subsection{Qualitative Comparisons}
As shown in \figref{comp1}, we present some reconstruction results generated 
from different methods.
It can be observed that other neural implicit methods generate 
incomplete object surfaces with noise and tends to model the 
fake specular reflection attached to the target object. 
The HSR is harmful to these methods to recover the target geometry. 
%
On the contrary, our approach achieves clearer results and 
reconstructs object surfaces with correct geometric details. 
This fact demonstrates the proposed auxiliary plane module can 
reduce the interface of the HSR and reconstruct the correct target 
object surface.
Besides, we further evaluate the robustness of each method 
in more challenging real-world HSR scenes. 
The real-world HSR encodes more diverse ambiguous information 
than the synthetic HSR for recovering the target object surface. 
The results in \figref{comp3} illustrate that our method generates 
better results containing thin structures of target objects 
when compared against \sArt neural implicit methods.

\subsection{Ablation Study}
We conduct several ablation experiments to study the impact of different 
settings on the auxiliary plane module, including 
the volume density $\mathrm{\sigma_r}$ and 
the plane attributes (including the position $\mathrm{d_r}$ and the plane normal $\mathbf{n}_r$). 
\figref{ab1} presents the results of each setting.

\myPara{Effect of the plane attributes.}
For each camera ray in a view, we use MLPs to generate the volume density $\mathrm{\sigma_r}$, and the attributes ($d_r$ and $\mathbf{n}_r$) of an auxiliary plane. 
When we remove the attributes of the auxiliary plane and only utilize $\mathrm{\sigma_r}$ to generate the weight, the performance degrades drastically compared to the full model.
Without the plane attributes, MLPs fail to implicitly trace the incident light to separate the target object appearance and the other part physically.  

\myPara{Effect of the volume density.}
To determine whether the volume density $\mathrm{\sigma_r}$ is necessary for recovering the object surface, we disable MLPs to output $\mathrm{\sigma_r}$ and adopt the same weights $w$ as the object path to render two appearances. This operation introduces the ambiguity from two paths to the MLPs of predicting SDF, then produces a worse result than the full model. However, our model with this setting still achieves better performance than the baseline NeuS because of the robust auxiliary planes. 

\section{Discussion}
\myPara{Components.}
Our model consists of two parts: the target object and the auxiliary plane. \figref{ab2} shows the components of each part. The target object appearances are faithfully enhanced and the HSR is captured by the auxiliary plane module. The surface normal and position of the auxiliary plane are adaptively learned by MLPs. The plane normals and positions on all camera rays of a view tend to be the same, which physically models a planar reflector.

\myPara{Attention Analysis.}
In HSR scenes, to recover accurate target objects, our model should pay more attention to the object path rather than the plane path. As \figref{weights} shows, the rendering weights of the target object have a higher peak value and a more concentrated distribution than the weights of the auxiliary plane and NeuS. This demonstrates that the auxiliary plane module makes MLPs focus on the target object and then reduces the interference of HSR to achieve more accurate results.

\myPara{Limitation.}
The proposed method inherits the ill-posed limitation from neural implicit methods of multi-view reconstruction. Due to the lack of priors, our model generates inaccurate geometry of target objects in unseen areas. 
A possible solution is introducing the symmetry of objects.

\section{Conclusion}
In this work, we have proposed a task of multi-view object reconstruction under the interference of HSR. To tackle this task, we present NeuS-HSR, a novel framework that recovers accurate 3D object surfaces against HSR. We propose decomposing scenes captured through glasses into the target object part and the auxiliary plane part for enhancing the target object by the auxiliary plane. We design an auxiliary plane module to physically generate the auxiliary plane appearance by using MLPs and the reflection transformation. Comprehensive experiments on both synthetic and real-world scenes illustrate that NeuS-HSR outperforms previous methods in quantitative reconstruction quality and visual inspection. Besides, the discussion explores the effectiveness of our decomposition in our task.

\section*{Acknowledgments}
This work is supported by the National Key Research and Development Program of China Grant (No.2018AAA0100400), NSFC (No.61922046) and NSFC (No.62132012).

{\small
\bibliographystyle{ieee_fullname}
\bibliography{egbib}

\begin{thebibliography}{10}\itemsep=-1pt

\bibitem{aanaes2016large}
Henrik Aan{\ae}s, Rasmus~Ramsb{\o}l Jensen, George Vogiatzis, Engin Tola, and
  Anders~Bjorholm Dahl.
\newblock Large-scale data for multiple-view stereopsis.
\newblock {\em International Journal of Computer Vision}, 120(2):153--168,
  2016.

\bibitem{atzmon2020sal}
Matan Atzmon and Yaron Lipman.
\newblock Sal: Sign agnostic learning of shapes from raw data.
\newblock In {\em Proceedings of the IEEE/CVF Conference on Computer Vision and
  Pattern Recognition}, pages 2565--2574, 2020.

\bibitem{barnes2009patchmatch}
Connelly Barnes, Eli Shechtman, Adam Finkelstein, and Dan~B Goldman.
\newblock Patchmatch: A randomized correspondence algorithm for structural
  image editing.
\newblock {\em ACM Trans. Graph.}, 28(3):24, 2009.

\bibitem{Chang_2021_WACV}
Ya-Chu Chang, Chia-Ni Lu, Chia-Chi Cheng, and Wei-Chen Chiu.
\newblock Single image reflection removal with edge guidance, reflection
  classifier, and recurrent decomposition.
\newblock In {\em Proceedings of the IEEE/CVF Winter Conference on Applications
  of Computer Vision (WACV)}, pages 2033--2042, January 2021.

\bibitem{cheng2021multi}
Ziang Cheng, Hongdong Li, Yuta Asano, Yinqiang Zheng, and Imari Sato.
\newblock Multi-view 3d reconstruction of a texture-less smooth surface of
  unknown generic reflectance.
\newblock In {\em Proceedings of the IEEE/CVF Conference on Computer Vision and
  Pattern Recognition}, pages 16226--16235, 2021.

\bibitem{cheng2022diffeomorphic}
Ziang Cheng, Hongdong Li, Richard Hartley, Yinqiang Zheng, and Imari Sato.
\newblock Diffeomorphic neural surface parameterization for 3d and reflectance
  acquisition.
\newblock In {\em ACM SIGGRAPH 2022 Conference Proceedings}, pages 1--10, 2022.

\bibitem{darmon2022improving}
Fran{\c{c}}ois Darmon, B{\'e}n{\'e}dicte Bascle, Jean-Cl{\'e}ment Devaux,
  Pascal Monasse, and Mathieu Aubry.
\newblock Improving neural implicit surfaces geometry with patch warping.
\newblock In {\em Proceedings of the IEEE/CVF Conference on Computer Vision and
  Pattern Recognition}, pages 6260--6269, 2022.

\bibitem{de1999poxels}
Jeremy~S De~Bonet and Paul Viola.
\newblock Poxels: Probabilistic voxelized volume reconstruction.
\newblock In {\em Proceedings of International Conference on Computer Vision
  (ICCV)}, volume~2, 1999.

\bibitem{Dong_2021_ICCV}
Zheng Dong, Ke Xu, Yin Yang, Hujun Bao, Weiwei Xu, and Rynson~W.H. Lau.
\newblock Location-aware single image reflection removal.
\newblock In {\em Proceedings of the IEEE/CVF International Conference on
  Computer Vision (ICCV)}, pages 5017--5026, October 2021.

\bibitem{fu2022geo}
Qiancheng Fu, Qingshan Xu, Yew-Soon Ong, and Wenbing Tao.
\newblock Geo-neus: Geometry-consistent neural implicit surfaces learning for
  multi-view reconstruction.
\newblock {\em Advances in Neural Information Processing Systems (NeurIPS)},
  2022.

\bibitem{furukawa2009accurate}
Yasutaka Furukawa and Jean Ponce.
\newblock Accurate, dense, and robust multiview stereopsis.
\newblock {\em IEEE transactions on pattern analysis and machine intelligence},
  32(8):1362--1376, 2009.

\bibitem{galliani2016gipuma}
Silvano Galliani, Katrin Lasinger, and Konrad Schindler.
\newblock Gipuma: Massively parallel multi-view stereo reconstruction.
\newblock {\em Publikationen der Deutschen Gesellschaft f{\"u}r
  Photogrammetrie, Fernerkundung und Geoinformation e. V}, 25(361-369):2, 2016.

\bibitem{godard2015multi}
Clement Godard, Peter Hedman, Wenbin Li, and Gabriel~J Brostow.
\newblock Multi-view reconstruction of highly specular surfaces in uncontrolled
  environments.
\newblock In {\em 2015 International Conference on 3D Vision}, pages 19--27.
  IEEE, 2015.

\bibitem{goel2022differentiable}
Shubham Goel, Georgia Gkioxari, and Jitendra Malik.
\newblock Differentiable stereopsis: Meshes from multiple views using
  differentiable rendering.
\newblock In {\em Proceedings of the IEEE/CVF Conference on Computer Vision and
  Pattern Recognition}, pages 8635--8644, 2022.

\bibitem{goesele2006multi}
Michael Goesele, Brian Curless, and Steven~M Seitz.
\newblock Multi-view stereo revisited.
\newblock In {\em 2006 IEEE Computer Society Conference on Computer Vision and
  Pattern Recognition (CVPR'06)}, volume~2, pages 2402--2409. IEEE, 2006.

\bibitem{goldman1990matrices}
Ronald Goldman.
\newblock {\em Matrices and transformations}.
\newblock Graphics Gems, 1990.

\bibitem{gropp2020implicit}
Amos Gropp, Lior Yariv, Niv Haim, Matan Atzmon, and Yaron Lipman.
\newblock Implicit geometric regularization for learning shapes.
\newblock In {\em International Conference on Machine Learning}, pages
  3789--3799. PMLR, 2020.

\bibitem{guo2022nerfren}
Yuan-Chen Guo, Di Kang, Linchao Bao, Yu He, and Song-Hai Zhang.
\newblock Nerfren: Neural radiance fields with reflections.
\newblock In {\em Proceedings of the IEEE/CVF Conference on Computer Vision and
  Pattern Recognition}, pages 18409--18418, 2022.

\bibitem{hernandez2008multiview}
Carlos Hernandez, George Vogiatzis, and Roberto Cipolla.
\newblock Multiview photometric stereo.
\newblock {\em IEEE Transactions on Pattern Analysis and Machine Intelligence},
  30(3):548--554, 2008.

\bibitem{immel1986radiosity}
David~S Immel, Michael~F Cohen, and Donald~P Greenberg.
\newblock A radiosity method for non-diffuse environments.
\newblock {\em Acm Siggraph Computer Graphics}, 20(4):133--142, 1986.

\bibitem{kajiya1986rendering}
James~T Kajiya.
\newblock The rendering equation.
\newblock In {\em Proceedings of the 13th annual conference on Computer
  graphics and interactive techniques}, pages 143--150, 1986.

\bibitem{kazhdan2013screened}
Michael Kazhdan and Hugues Hoppe.
\newblock Screened poisson surface reconstruction.
\newblock {\em ACM Transactions on Graphics (ToG)}, 32(3):1--13, 2013.

\bibitem{lei2021robust}
Chenyang Lei and Qifeng Chen.
\newblock Robust reflection removal with reflection-free flash-only cues.
\newblock In {\em Proceedings of the IEEE/CVF Conference on Computer Vision and
  Pattern Recognition}, pages 14811--14820, 2021.

\bibitem{lei2022categorized}
Chenyang Lei, Xuhua Huang, Chenyang Qi, Yankun Zhao, Wenxiu Sun, Qiong Yan, and
  Qifeng Chen.
\newblock A categorized reflection removal dataset with diverse real-world
  scenes.
\newblock In {\em Proceedings of the IEEE/CVF Conference on Computer Vision and
  Pattern Recognition}, pages 3040--3048, 2022.

\bibitem{li2020through}
Zhengqin Li, Yu-Ying Yeh, and Manmohan Chandraker.
\newblock Through the looking glass: Neural 3d reconstruction of transparent
  shapes.
\newblock In {\em Proceedings of the IEEE/CVF Conference on Computer Vision and
  Pattern Recognition}, pages 1262--1271, 2020.

\bibitem{lin2021barf}
Chen-Hsuan Lin, Wei-Chiu Ma, Antonio Torralba, and Simon Lucey.
\newblock Barf: Bundle-adjusting neural radiance fields.
\newblock In {\em Proceedings of the IEEE/CVF International Conference on
  Computer Vision}, pages 5741--5751, 2021.

\bibitem{liu2022adaptive}
Ming Liu, Jianan Pan, Zifei Yan, Wangmeng Zuo, and Lei Zhang.
\newblock Adaptive network combination for single-image reflection removal: A
  domain generalization perspective.
\newblock {\em arXiv preprint arXiv:2204.01505}, 2022.

\bibitem{lombardi2019neural}
Stephen Lombardi, Tomas Simon, Jason Saragih, Gabriel Schwartz, Andreas
  Lehrmann, and Yaser Sheikh.
\newblock Neural volumes: learning dynamic renderable volumes from images.
\newblock {\em ACM Transactions on Graphics (TOG)}, 38(4):1--14, 2019.

\bibitem{lorensen1987marching}
William~E Lorensen and Harvey~E Cline.
\newblock Marching cubes: A high resolution 3d surface construction algorithm.
\newblock {\em ACM siggraph computer graphics}, 21(4):163--169, 1987.

\bibitem{mildenhall2020nerf}
Ben Mildenhall, Pratul~P Srinivasan, Matthew Tancik, Jonathan~T Barron, Ravi
  Ramamoorthi, and Ren Ng.
\newblock Nerf: Representing scenes as neural radiance fields for view
  synthesis.
\newblock In {\em European conference on computer vision}, pages 405--421.
  Springer, 2020.

\bibitem{niemeyer2020differentiable}
Michael Niemeyer, Lars Mescheder, Michael Oechsle, and Andreas Geiger.
\newblock Differentiable volumetric rendering: Learning implicit 3d
  representations without 3d supervision.
\newblock In {\em Proceedings of the IEEE/CVF Conference on Computer Vision and
  Pattern Recognition}, pages 3504--3515, 2020.

\bibitem{oechsle2021unisurf}
Michael Oechsle, Songyou Peng, and Andreas Geiger.
\newblock Unisurf: Unifying neural implicit surfaces and radiance fields for
  multi-view reconstruction.
\newblock In {\em Proceedings of the IEEE/CVF International Conference on
  Computer Vision}, pages 5589--5599, 2021.

\bibitem{peng2020convolutional}
Songyou Peng, Michael Niemeyer, Lars Mescheder, Marc Pollefeys, and Andreas
  Geiger.
\newblock Convolutional occupancy networks.
\newblock In {\em Computer Vision--ECCV 2020: 16th European Conference,
  Glasgow, UK, August 23--28, 2020, Proceedings, Part III 16}, pages 523--540.
  Springer, 2020.

\bibitem{rasmuson2022addressing}
Sverker Rasmuson, Erik Sintorn, and Ulf Assarsson.
\newblock Addressing the shape-radiance ambiguity in view-dependent radiance
  fields.
\newblock {\em arXiv preprint arXiv:2203.01553}, 2022.

\bibitem{schonberger2016structure}
Johannes~L Schonberger and Jan-Michael Frahm.
\newblock Structure-from-motion revisited.
\newblock In {\em Proceedings of the IEEE conference on computer vision and
  pattern recognition}, pages 4104--4113, 2016.

\bibitem{schonberger2016pixelwise}
Johannes~L Sch{\"o}nberger, Enliang Zheng, Jan-Michael Frahm, and Marc
  Pollefeys.
\newblock Pixelwise view selection for unstructured multi-view stereo.
\newblock In {\em European conference on computer vision}, pages 501--518.
  Springer, 2016.

\bibitem{schoenberger2016mvs}
Johannes~Lutz Sch\"{o}nberger, Enliang Zheng, Marc Pollefeys, and Jan-Michael
  Frahm.
\newblock Pixelwise view selection for unstructured multi-view stereo.
\newblock In {\em European Conference on Computer Vision (ECCV)}, 2016.

\bibitem{seitz2006comparison}
Steven~M Seitz, Brian Curless, James Diebel, Daniel Scharstein, and Richard
  Szeliski.
\newblock A comparison and evaluation of multi-view stereo reconstruction
  algorithms.
\newblock In {\em 2006 IEEE computer society conference on computer vision and
  pattern recognition (CVPR'06)}, volume~1, pages 519--528. IEEE, 2006.

\bibitem{seitz1999photorealistic}
Steven~M Seitz and Charles~R Dyer.
\newblock Photorealistic scene reconstruction by voxel coloring.
\newblock {\em International Journal of Computer Vision}, 35(2):151--173, 1999.

\bibitem{sitzmann2019scene}
Vincent Sitzmann, Michael Zollh{\"o}fer, and Gordon Wetzstein.
\newblock Scene representation networks: Continuous 3d-structure-aware neural
  scene representations.
\newblock {\em Advances in Neural Information Processing Systems}, 32, 2019.

\bibitem{song2022multi}
Binbin Song, Jiantao Zhou, and Haiwei Wu.
\newblock Multi-stage curvature-guided network for progressive single image
  reflection removal.
\newblock {\em IEEE Transactions on Circuits and Systems for Video Technology},
  2022.

\bibitem{sun2022neural}
Jiaming Sun, Xi Chen, Qianqian Wang, Zhengqi Li, Hadar Averbuch-Elor, Xiaowei
  Zhou, and Noah Snavely.
\newblock Neural 3d reconstruction in the wild.
\newblock In {\em ACM SIGGRAPH 2022 Conference Proceedings}, pages 1--9, 2022.

\bibitem{ullman1979interpretation}
Shimon Ullman.
\newblock The interpretation of structure from motion.
\newblock {\em Proceedings of the Royal Society of London. Series B. Biological
  Sciences}, 203(1153):405--426, 1979.

\bibitem{verbin2022ref}
Dor Verbin, Peter Hedman, Ben Mildenhall, Todd Zickler, Jonathan~T Barron, and
  Pratul~P Srinivasan.
\newblock Ref-nerf: Structured view-dependent appearance for neural radiance
  fields.
\newblock In {\em 2022 IEEE/CVF Conference on Computer Vision and Pattern
  Recognition (CVPR)}, pages 5481--5490. IEEE, 2022.

\bibitem{wang2022neuris}
Jiepeng Wang, Peng Wang, Xiaoxiao Long, Christian Theobalt, Taku Komura,
  Lingjie Liu, and Wenping Wang.
\newblock Neuris: Neural reconstruction of indoor scenes using normal priors.
\newblock In {\em Computer Vision--ECCV 2022: 17th European Conference, Tel
  Aviv, Israel, October 23--27, 2022, Proceedings, Part XXXII}, pages 139--155.
  Springer, 2022.

\bibitem{wang2021neus}
Peng Wang, Lingjie Liu, Yuan Liu, Christian Theobalt, Taku Komura, and Wenping
  Wang.
\newblock Neus: Learning neural implicit surfaces by volume rendering for
  multi-view reconstruction.
\newblock {\em Advances in Neural Information Processing Systems},
  34:27171--27183, 2021.

\bibitem{wanghf}
Yiqun Wang, Ivan Skorokhodov, and Peter Wonka.
\newblock Hf-neus: Improved surface reconstruction using high-frequency
  details.
\newblock In {\em Advances in Neural Information Processing Systems}, 2022.

\bibitem{wei2021nerfingmvs}
Yi Wei, Shaohui Liu, Yongming Rao, Wang Zhao, Jiwen Lu, and Jie Zhou.
\newblock Nerfingmvs: Guided optimization of neural radiance fields for indoor
  multi-view stereo.
\newblock In {\em Proceedings of the IEEE/CVF International Conference on
  Computer Vision}, pages 5610--5619, 2021.

\bibitem{worchel2022multi}
Markus Worchel, Rodrigo Diaz, Weiwen Hu, Oliver Schreer, Ingo Feldmann, and
  Peter Eisert.
\newblock Multi-view mesh reconstruction with neural deferred shading.
\newblock In {\em Proceedings of the IEEE/CVF Conference on Computer Vision and
  Pattern Recognition}, pages 6187--6197, 2022.

\bibitem{wu2011visualsfm}
Changchang Wu.
\newblock Visualsfm: A visual structure from motion system.
\newblock {\em http://www. cs. washington. edu/homes/ccwu/vsfm}, 2011.

\bibitem{xu2022hybrid}
Jiamin Xu, Zihan Zhu, Hujun Bao, and Wewei Xu.
\newblock A hybrid mesh-neural representation for 3d transparent object
  reconstruction.
\newblock {\em arXiv preprint arXiv:2203.12613}, 2022.

\bibitem{yang2022s}
Wenqi Yang, Guanying Chen, Chaofeng Chen, Zhenfang Chen, and Kwan-Yee~K Wong.
\newblock S3-nerf: Neural reflectance field from shading and shadow under a
  single viewpoint.
\newblock {\em Advances in Neural Information Processing Systems (NeurIPS)},
  2022.

\bibitem{yariv2021volume}
Lior Yariv, Jiatao Gu, Yoni Kasten, and Yaron Lipman.
\newblock Volume rendering of neural implicit surfaces.
\newblock {\em Advances in Neural Information Processing Systems},
  34:4805--4815, 2021.

\bibitem{yariv2020multiview}
Lior Yariv, Yoni Kasten, Dror Moran, Meirav Galun, Matan Atzmon, Basri Ronen,
  and Yaron Lipman.
\newblock Multiview neural surface reconstruction by disentangling geometry and
  appearance.
\newblock {\em Advances in Neural Information Processing Systems},
  33:2492--2502, 2020.

\bibitem{zhang2021ners}
Jason Zhang, Gengshan Yang, Shubham Tulsiani, and Deva Ramanan.
\newblock Ners: Neural reflectance surfaces for sparse-view 3d reconstruction
  in the wild.
\newblock {\em Advances in Neural Information Processing Systems}, 34, 2021.

\bibitem{zhang2020nerf++}
Kai Zhang, Gernot Riegler, Noah Snavely, and Vladlen Koltun.
\newblock Nerf++: Analyzing and improving neural radiance fields.
\newblock {\em arXiv preprint arXiv:2010.07492}, 2020.

\bibitem{zhang2018single}
Xuaner Zhang, Ren Ng, and Qifeng Chen.
\newblock Single image reflection separation with perceptual losses.
\newblock In {\em Proceedings of the IEEE conference on computer vision and
  pattern recognition}, pages 4786--4794, 2018.

\end{thebibliography}
}
\clearpage

\section*{Appendix}  
\appendix
Unlike recent methods of transparent object reconstruction~\cite{li2020through,xu2022hybrid} which aim to reconstruct 3D transparent objects, our goal is to recover the object surface behind transparent objects (\ie, glasses). Here, we provide more details of our method and experiments. Specifically, we provide the details of the reflection transformation~\cite{goldman1990matrices} and projection algorithm we adopt in our manuscript (\secref{Sec:rt}), the choice of the ratio in the linear summation (\secref{Sec:ls}), additional details of experiments (\secref{Sec:aid}), additional results (\secref{Sec:mrsd}), additional analysis (\secref{Sec:mrrw}) and the future work (\secref{Sec:fw}). The source code will be publicly released.

\section{Projection Algorithm} \label{Sec:rt}
\begin{figure}[h]
	\centering
	\includegraphics[width=\linewidth]{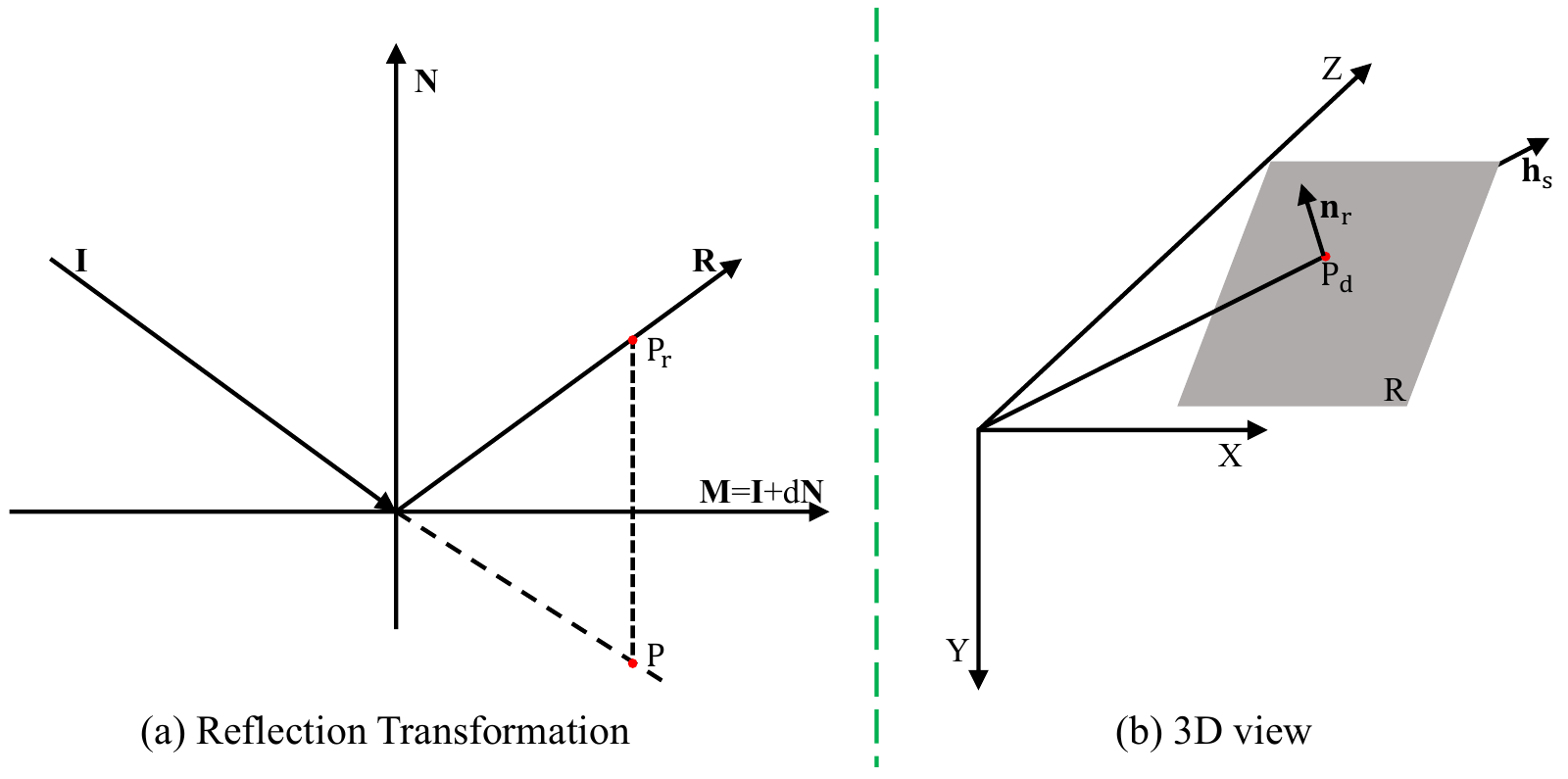}
	\vspace*{-2mm}
	\caption{(a): Reflection transformation~\cite{goldman1990matrices}. (b) 3D view of an auxiliary plane in the camera coordinate system. $\mathrm{R}$ is an auxiliary plane of a camera ray $\mathbf{h}_s$. $\mathbf{n}_r$ is the unit normal vector of $\mathrm{R}$.}
	\label{supp_rt}
\end{figure}
\figref{supp_rt} (a) shows an illustration of the reflection transformation~\cite{goldman1990matrices}. Suppose a ray is incident on a glass $\mathbf{M}$, the incident direction is $\mathbf{I}$, the reflected direction is $\mathbf{R}$ and the plane equation of $\mathbf{M}$ is defined as:
\begin{equation}\label{eq1_a}
\mathrm{L\cdot P = Ax+By+Cz+D = 0}
\end{equation}
where $\mathrm{L=(A,B,C,D)}$ and $\mathrm{P=(x,y,z,1)}$. The unit normal vector $\mathbf{N}$ of $\mathbf{M}$ is $\mathrm{(A,B,C,0)}$.

Given $\mathrm{P_r}$ is a point on the reflected light, and $\mathrm{P}$ is its virtual image, then $\mathrm{L\cdot P_r}$ is the vertical distance of $\mathrm{P_r}$ from $\mathbf{M}$, then we have:
\begin{equation}\label{eq2_a}
\mathrm{P = P_r - 2(L\cdot P_r)}\mathbf{N}\mathrm{ = M_rP_r}
\end{equation}
where $\mathrm{M_r}$ is denoted by:
\begin{equation}\label{eq3_a}
\mathrm{M_r}= \left[
\begin{aligned}
&\mathrm{1-2A^2}& \mathrm{-2AB}& \quad\mathrm{-2AC}& \mathrm{-2AD}  \\  
&\mathrm{-2AB}& \quad\mathrm{1-2B^2}& \quad\mathrm{-2BC}& \mathrm{-2BD} \\
&\mathrm{-2AC}& \mathrm{-2BC}& \quad\mathrm{1-2C^2}& \mathrm{-2CD} \\
&\quad\quad\mathrm{0}&\mathrm{0}&\quad\quad\quad\mathrm{0}&\mathrm{1}
\end{aligned}
\right]
\end{equation}

Obviously, \eqnref{eq2_a} is a differentiable function. In our work, the camera ray is along the negative $\mathbf{R}$ and we use the reflection transformation to trace the incident ray. \figref{supp_rt} (b) shows the 3D view of a auxiliary plane.
The auxiliary plane is built in the camera coordinate system. Given the Cartesian viewing direction unit vector $\mathbf{v}=\mathrm{(x_v,y_v,z_v)}$ and plane position $\mathrm{d_r}$, then we have:
\begin{equation}\label{eq4_a}
\mathrm{P_d} = \mathrm{d_r}\mathbf{v} = \mathrm{(d_rx_v, d_ry_v, d_rz_v)}
\end{equation}
$\mathrm{P_d}$ is on the auxiliary plane (\ie, \eqnref{eq1}). Given $\mathbf{n}_r=\mathrm{(A,B,C)}$, then we have:

\begin{equation}\label{eq5_a}
\begin{aligned}
\mathrm{D} &= \mathrm{-(Ad_rx_v+Bd_ry_v+Cd_rz_v)} \\
           &= \mathrm{-d_r(Ax_v+By_v+Cz_v)} \\
           &= \mathrm{-d_r}\mathbf{n}_r\cdot\mathbf{v}
\end{aligned}
\end{equation}
Based on the above description, we present our strategy of acquiring the input points $\mathbf{p}_r$ of the plane path in \AlgRef{a_rt}. 

\begin{algorithm}[t]
\caption{Transforming sampled points along a camera ray $\mathbf{h}_s$.}
\label{a_rt}
\KwIn{The plane normal $\mathbf{n}_r = \mathrm{(A, B, C)}$, the plane position $\mathrm{d_r}$, the camera center $\mathbf{o}$, the depth $\mathrm{t}$, the view direction $\mathbf{v}$ and the sampled points $\mathbf{p}$ along $\mathbf{h}_s$.}
\KwOut{Spatial points $\mathbf{p}_r$ of the plane path in the camera coordinate system.}
$\mathbf{p'}$ = $\mathbf{p}_t \cup \mathbf{p'}_t$ = $\mathbf{p}-\mathbf{o}$;

$\mathrm{P_d}$ = $\mathrm{d_r\textbf{v}}$;  

$\mathrm{D}$ = $\mathrm{-d_r}\mathbf{n}_r\cdot\mathbf{v}$; 

$\mathbf{p}_t$ = \{$\mathrm{t}\mathbf{v}| \mathrm{t \in [0,d_r]}$\};

$\mathbf{p'}_t$ = \{$\mathrm{t}\mathbf{v}|\mathrm{t \in [d_r, 1]}$\};

\{$\mathrm{D}$, $\mathbf{n}_r$\} $\rightarrow$ $\mathrm{M_r}$;

$\mathbf{p}_a = \mathrm{M_r^{-1}}\mathbf{p'}_{t}$;

$\mathbf{p}_r$ = $\mathbf{p}_t \cup \mathbf{p}_a$.
\end{algorithm}

\section{Linear Summation} \label{Sec:ls}
\begin{table}[t]
\centering
\caption{Effects of different ratios of the target object appearance on `scan24'. The standard deviation is $\mathrm{0.68}$.}
\vspace*{-2mm}
\begin{tabular}{c|c|c|c|c|c}
\cline{1-6}
Ratio   & 0.1 & 0.3 & 0.5 & 0.7 & 0.9 \\ \cline{1-6}
Chamfer distance $\downarrow$  & 3.29 & 2.07 & 2.43 & 3.99 & 3.17  \\ \cline{1-6}
\end{tabular}
\label{supp_ratio}
\end{table}
We fuse the appearances of two paths by a linear summation to be the rendered image, which can be supervised by the captured RGB image. 
As \tabref{supp_ratio} shows, different ratios of the target object appearance have different effects on the reconstruction quality of the target object. We select $\mathrm{0.3}$ as the default ratio in our model according to these results. 

\begin{table*}[t]
\centering
\footnotesize
\caption{Metrics of datasets used in our experiments.}
\vspace*{-2mm}
\begin{tabular}{c|c|c|c|c|c|c|c}
\cline{1-8}
Scene   & Synthetic & Buddha & Toys & Figure & Plate & Porcelain & Bronze \\ \cline{1-8}
Views  & 49/64 & 56 & 23 & 60 & 56 & 60 & 43  \\ \cline{1-8}
Resolution  & 1600 $\times$ 1200  & 1920 $\times$ 1080 & 1372 $\times$ 1029 & 1920 $\times$ 1080 & 1920 $\times$ 1080 & 1080 $\times$ 1920 & 1080 $\times$ 1920  \\ \cline{1-8}
\end{tabular}
\label{data}
\end{table*}
\begin{table}[t]
\centering
\footnotesize
\caption{Model parameters of NeuS-HSR and baselines.}
\vspace*{-2mm}
\begin{tabular}{c|c|c|c|c}
\cline{1-5}
Method   & UNISURF~\cite{oechsle2021unisurf} & VolSDF~\cite{yariv2021volume} & NeuS~\cite{wang2021neus} & NeuS-HSR \\ \cline{1-5}
\#Params  & 0.8M & 1.4M & 1.4M & 1.5M  \\ \cline{1-5}
\end{tabular}
\label{params}
\end{table}

\section{Additional Experimental Details} \label{Sec:aid}
\subsection{Hierarchical Sampling}
We follow the hierarchical sampling of NeuS~\cite{wang2021neus} to generate the input spatial points. Specifically, we sample $\mathrm{64}$ points along a ray uniformly at first, then we perform the importance sampling for $\mathrm{4}$ times. The total number of sampled points is $\mathrm{128}$. We sample extra $\mathrm{32}$ points outside the sphere according to NeRF++~\cite{zhang2020nerf++}.

\subsection{Neural Network Architecture}
The whole network architecture consists of three parts: SDF predicting, auxiliary plane predicting, and color predicting. For SDF predicting, we follow the neural network architecture of NeuS, which is activated by Softplus where $\beta=100$. Weight normalization is adopted for stable training. The input is concatenated with the features from the fourth layer by a skip connection. For auxiliary plane predicting, the volume density part consists of three linear layers with ReLU, the position and normal branches both consist of two linear layers. The hidden layers of two branches are activated by ReLU, while the last layers of two branches are activated by Sigmoid and Tanh respectively. For color predicting, the hidden layers are activated by ReLU, and the last layer is activated by Sigmoid.  

\subsection{Datasets}
\begin{figure*}[!htbp]
	\centering
	\begin{tabular}{*{6}{c@{\hspace{1px}}}}
	    \raisebox{0.3\height}{\rotatebox{90}{Buddha}}
		\includegraphics[width=0.19\textwidth]{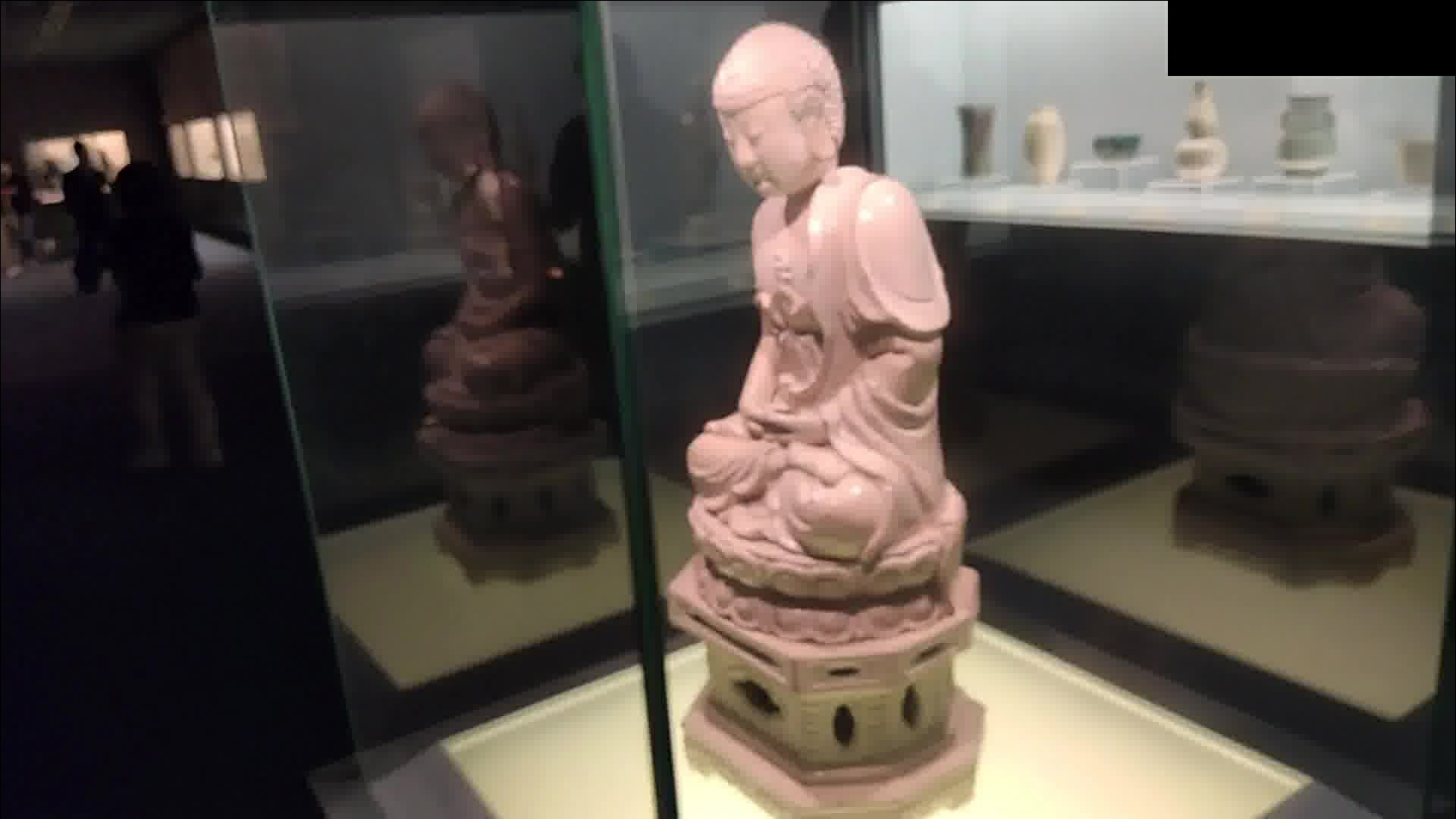}&
		\includegraphics[width=0.19\textwidth]{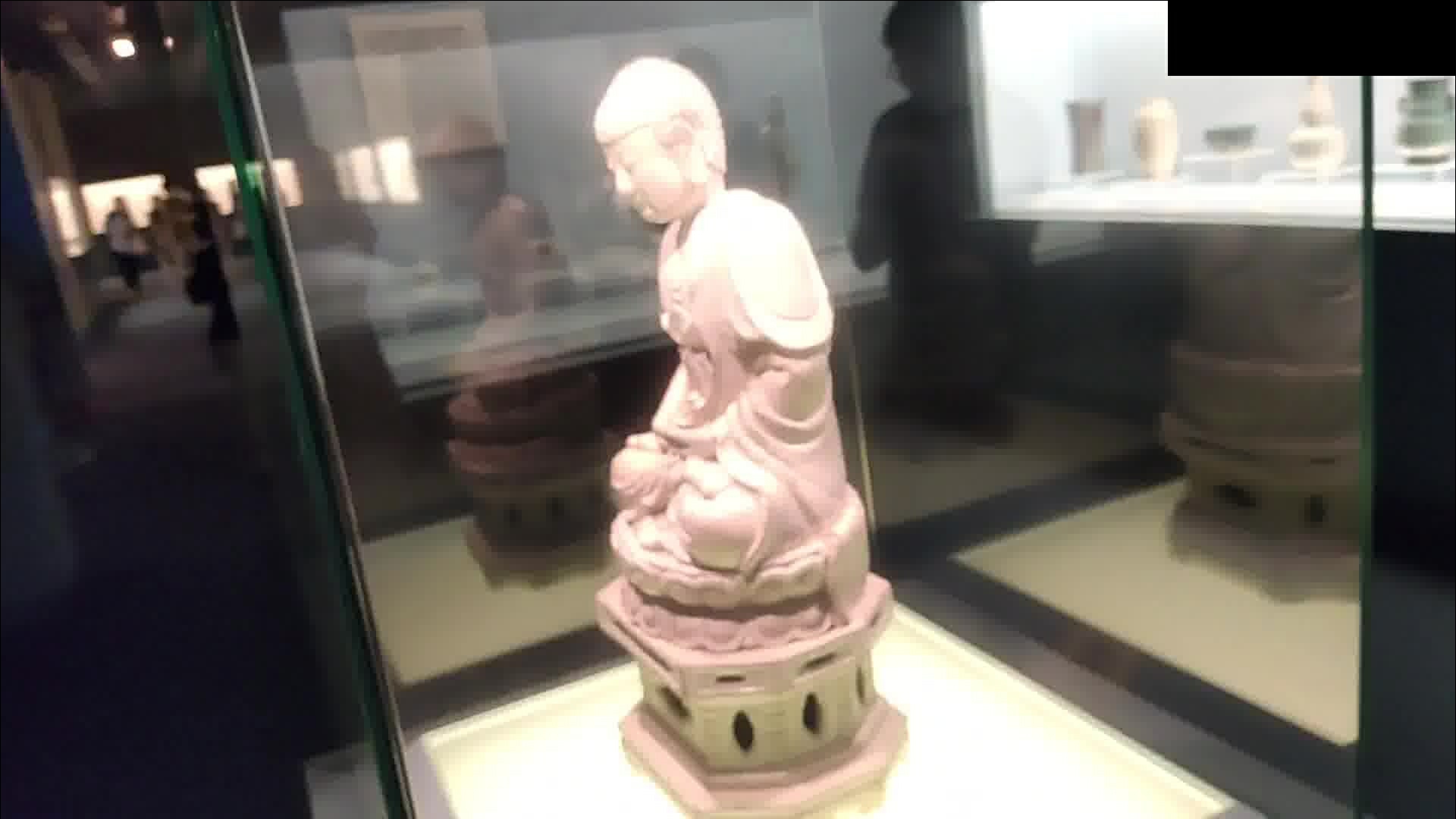}&
		\includegraphics[width=0.19\textwidth]{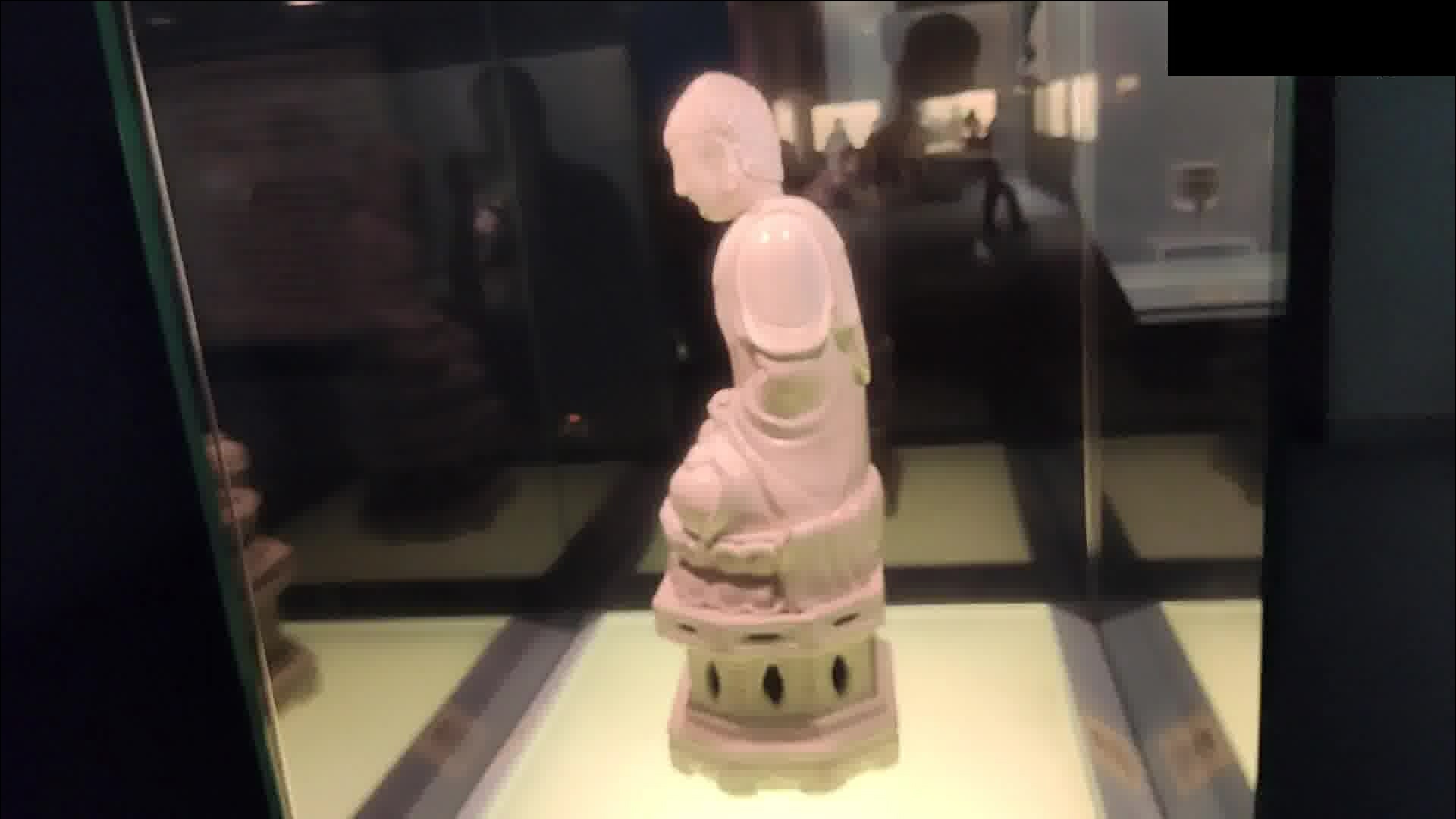}&
		\includegraphics[width=0.19\textwidth]{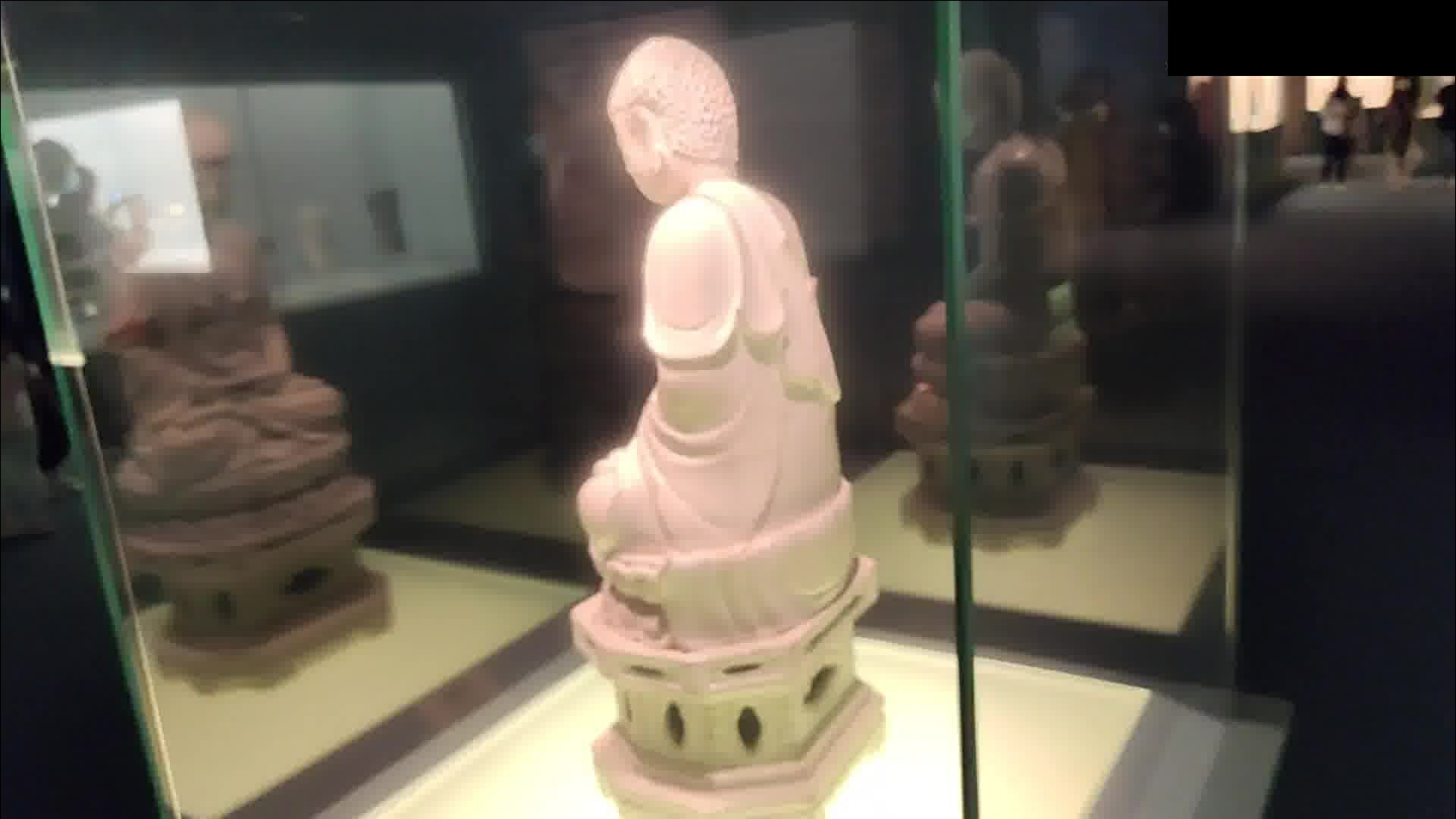}&
		\includegraphics[width=0.19\textwidth]{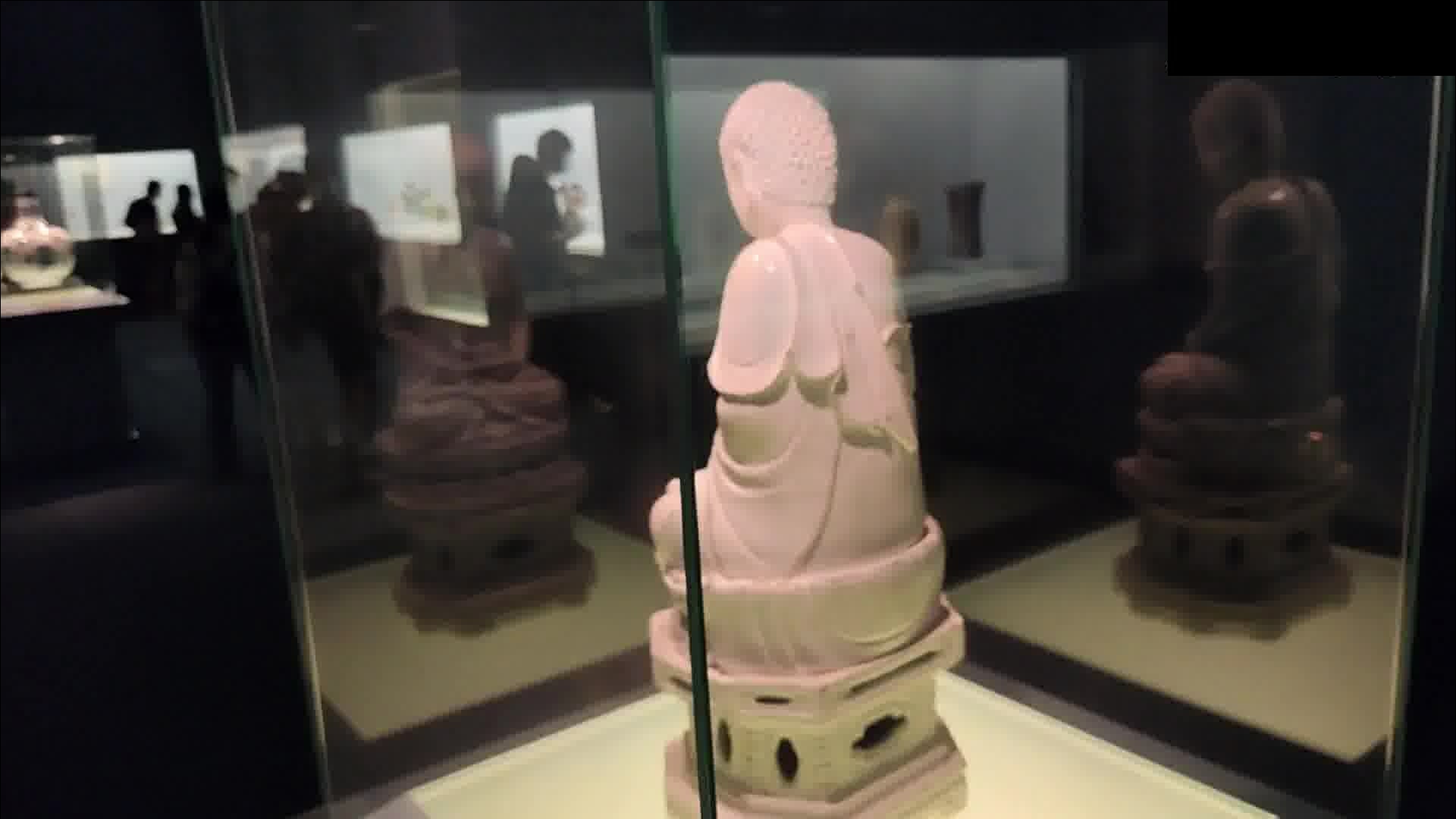}
        \\
        \raisebox{\height}{\rotatebox{90}{Toys}}
		\includegraphics[width=0.19\textwidth]{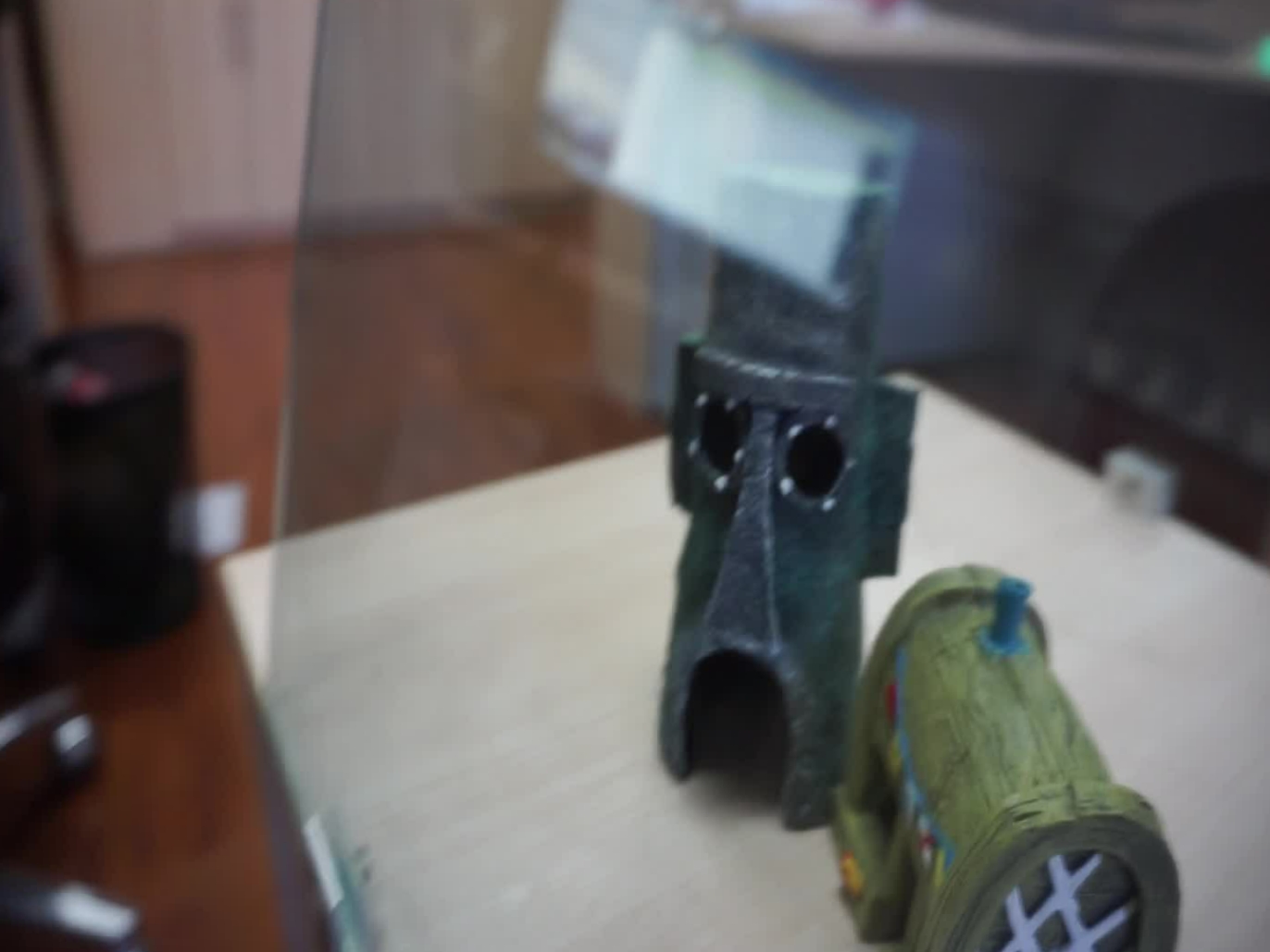}&
		\includegraphics[width=0.19\textwidth]{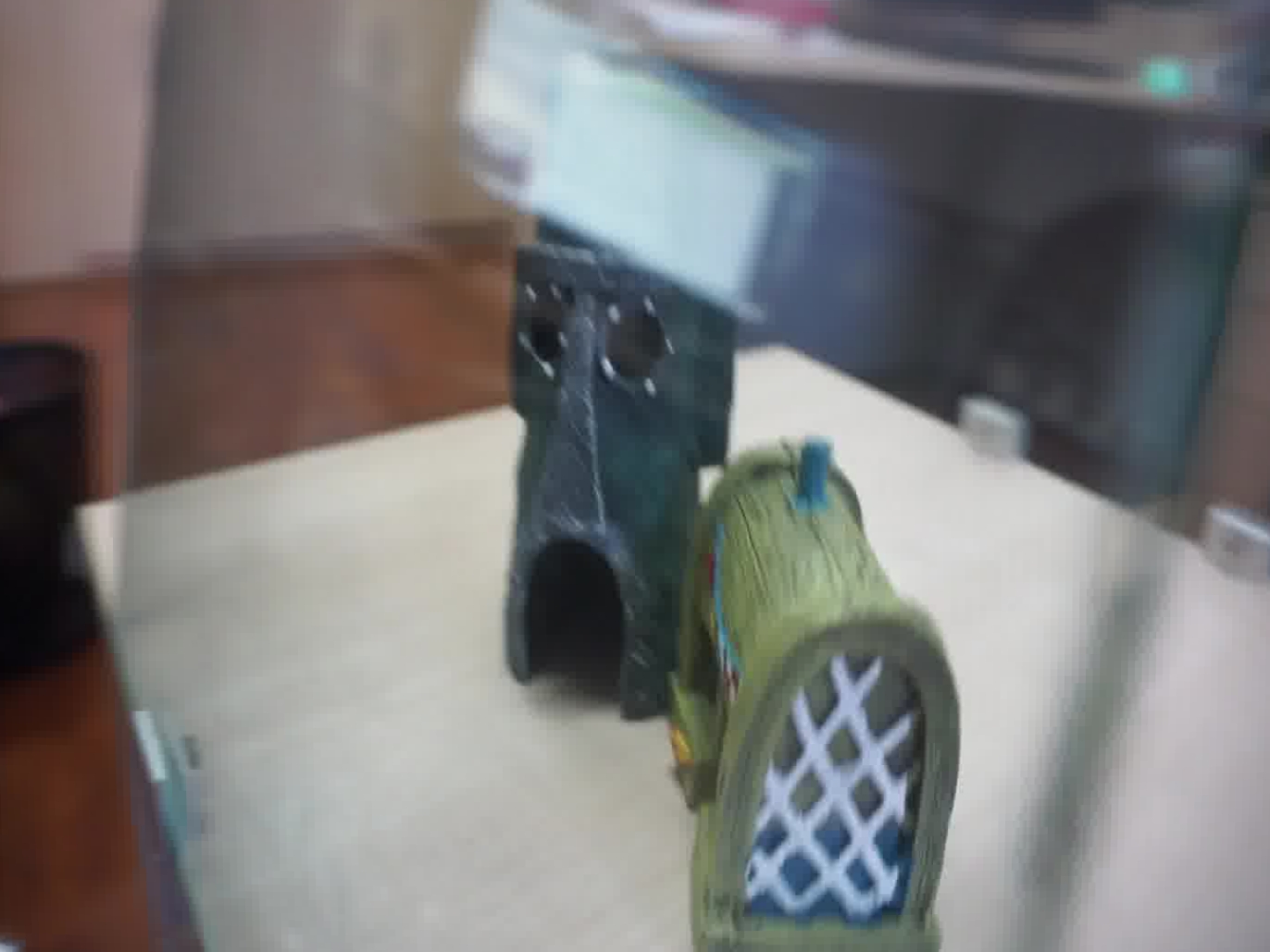}&
		\includegraphics[width=0.19\textwidth]{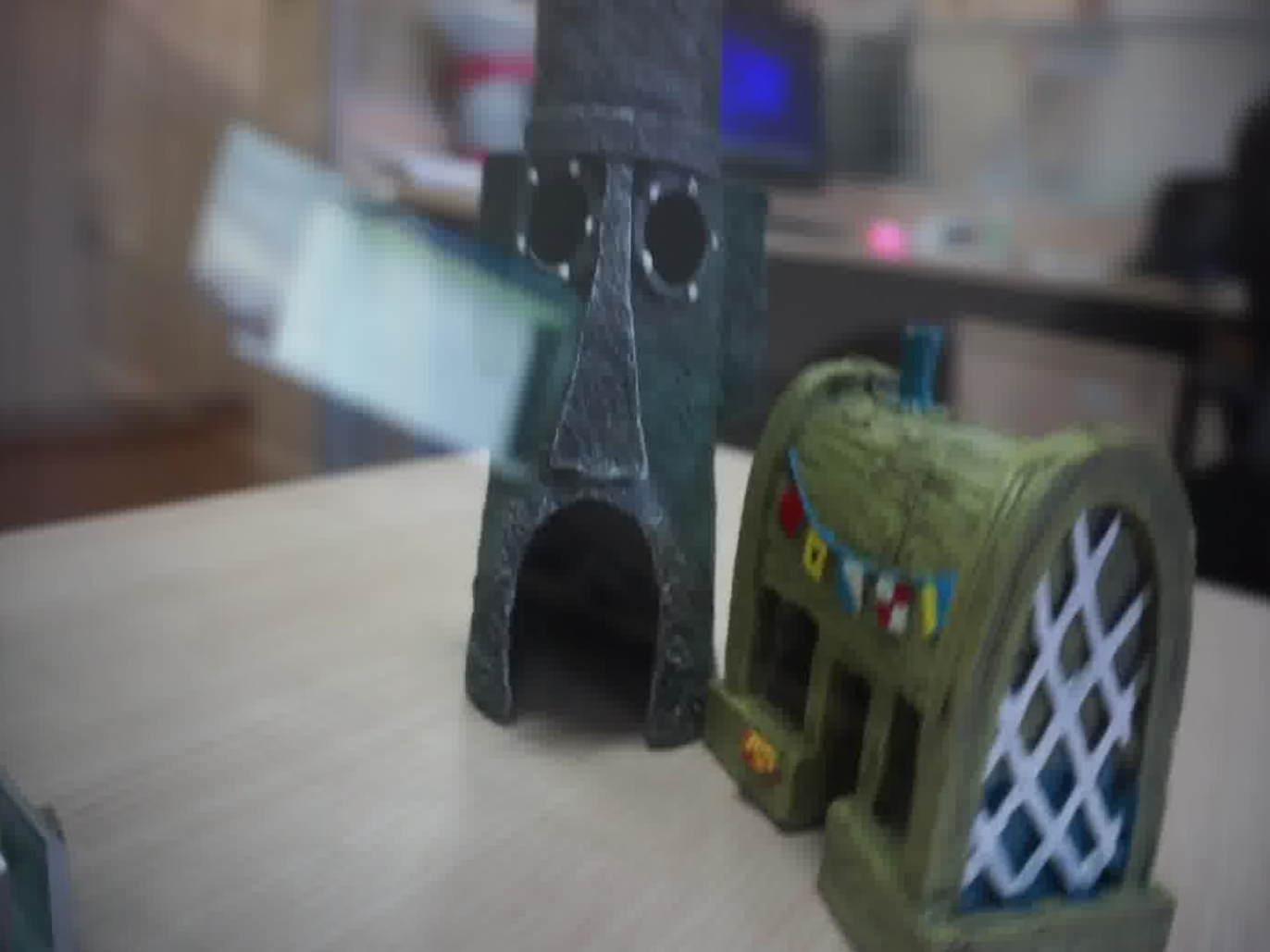}&
		\includegraphics[width=0.19\textwidth]{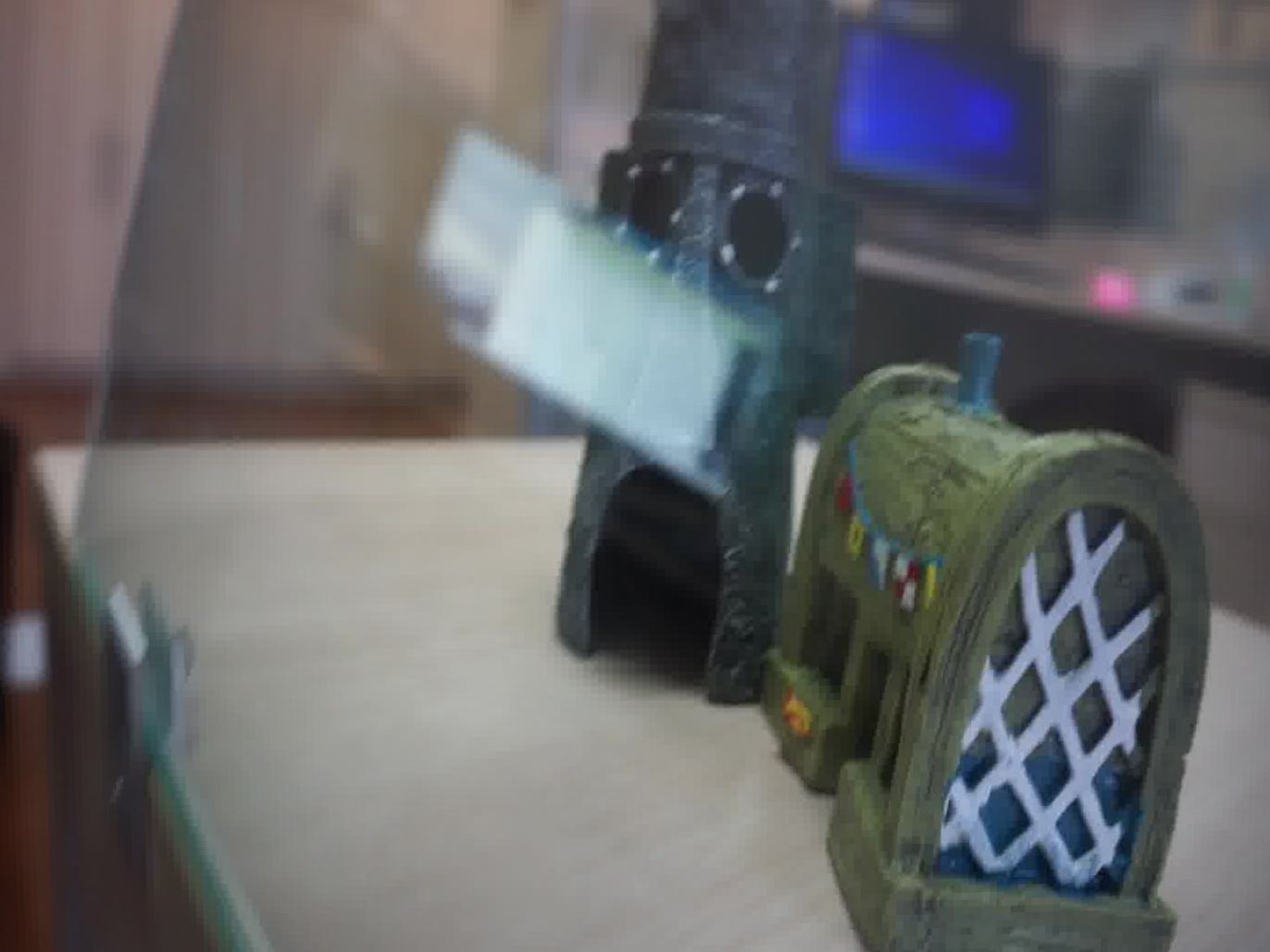}&
		\includegraphics[width=0.19\textwidth]{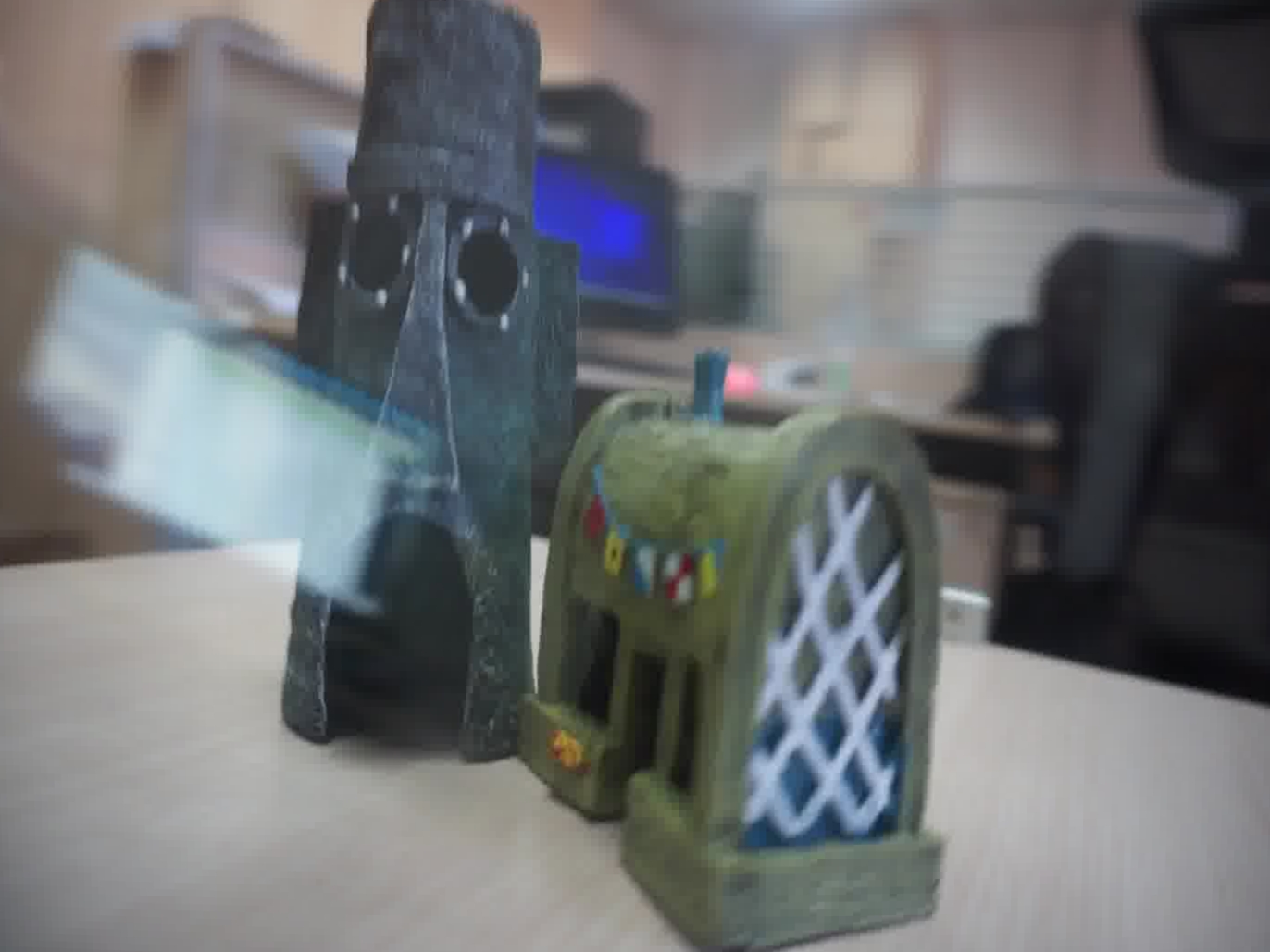}
        \\
        \raisebox{0.5\height}{\rotatebox{90}{Figure}}
		\includegraphics[width=0.19\textwidth]{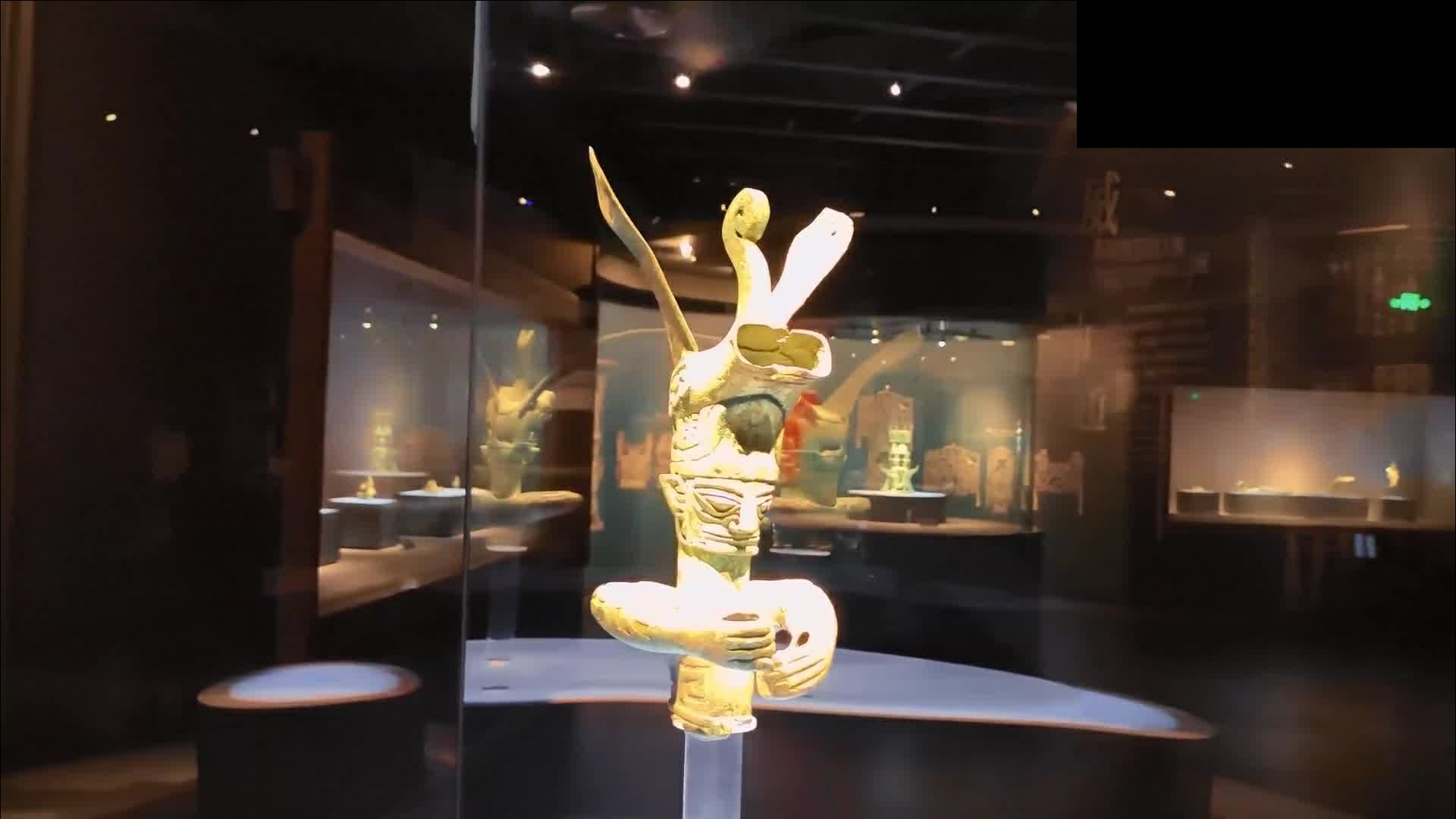}&
		\includegraphics[width=0.19\textwidth]{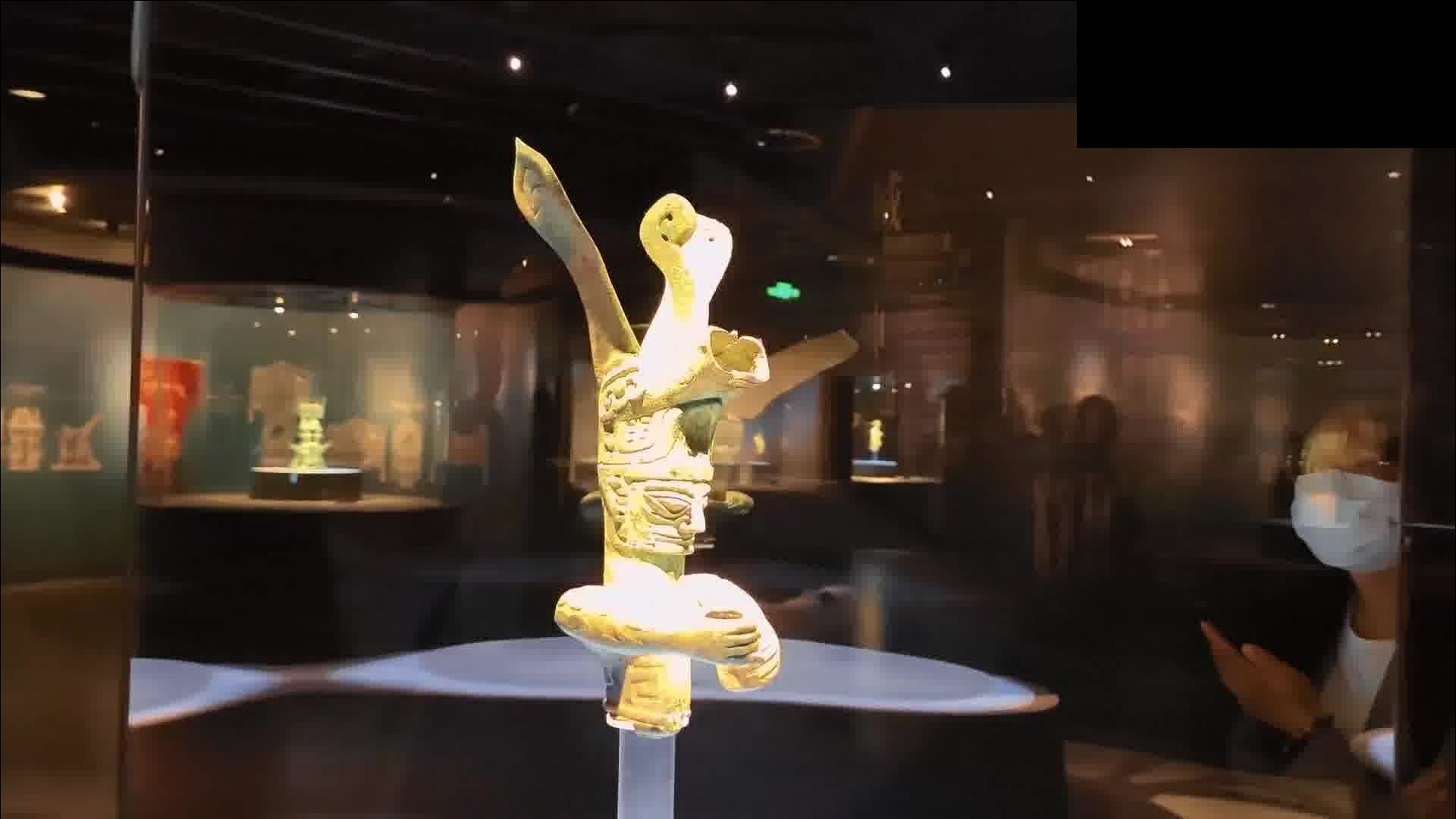}&
		\includegraphics[width=0.19\textwidth]{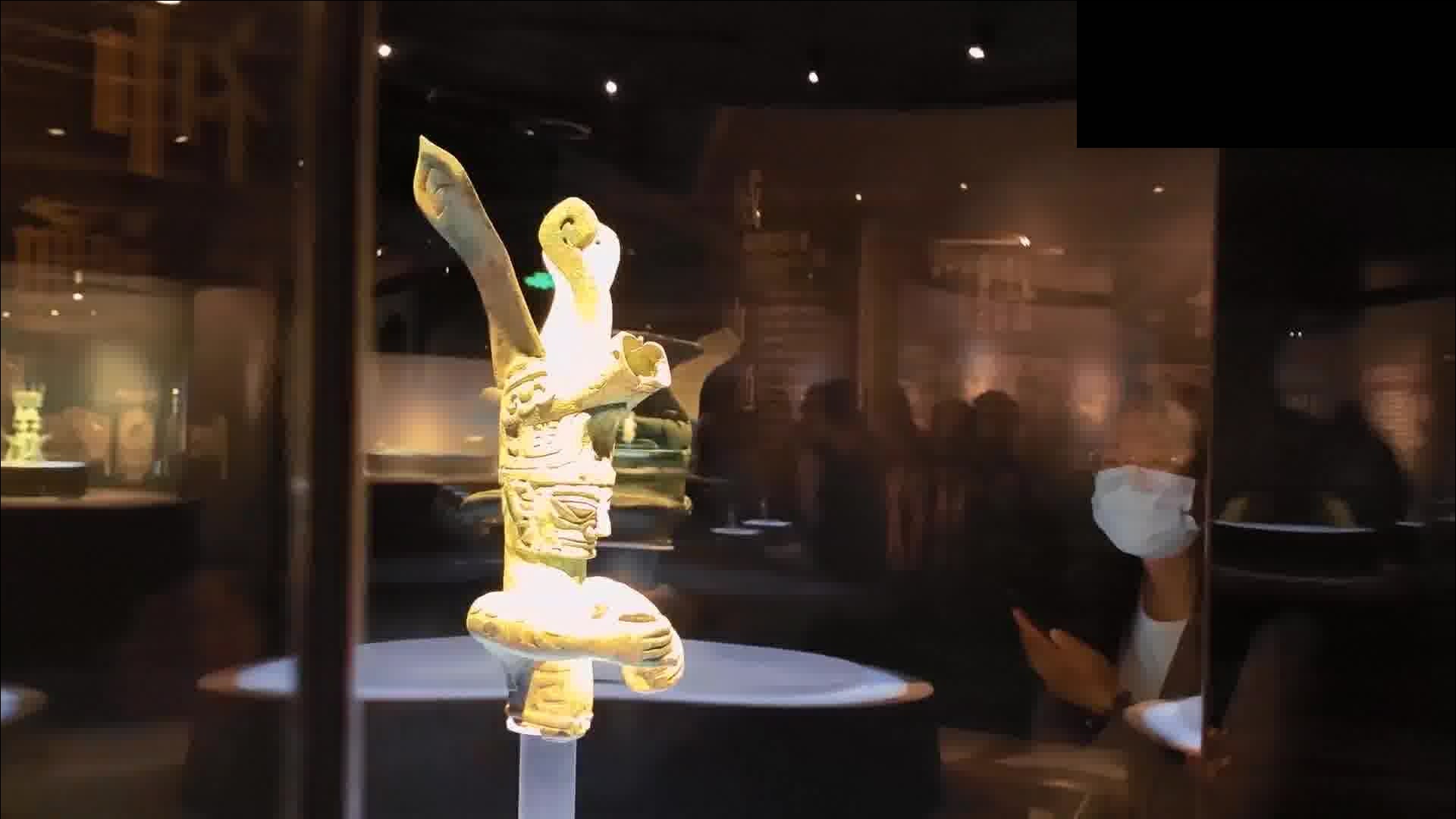}&
		\includegraphics[width=0.19\textwidth]{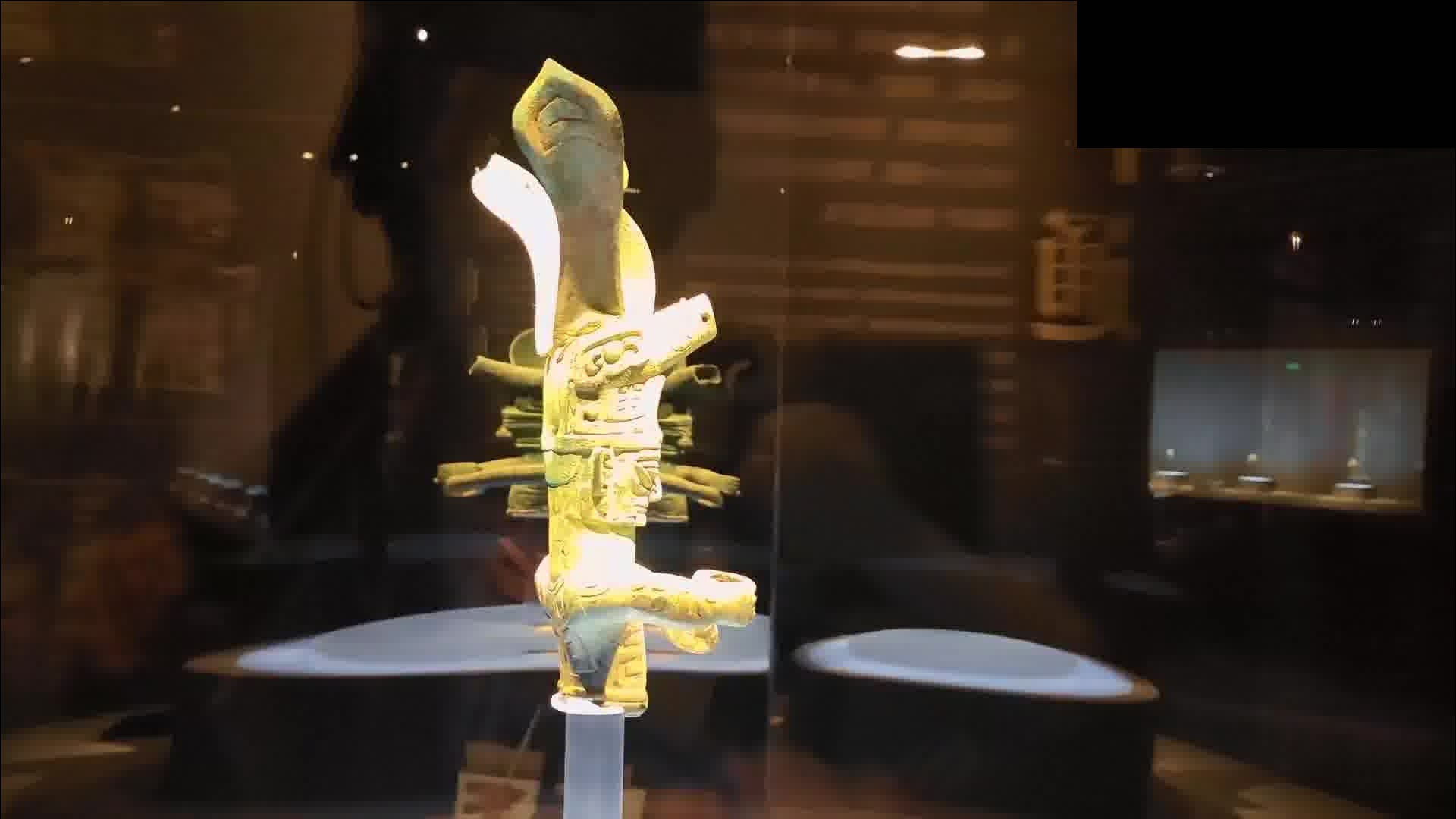}&
		\includegraphics[width=0.19\textwidth]{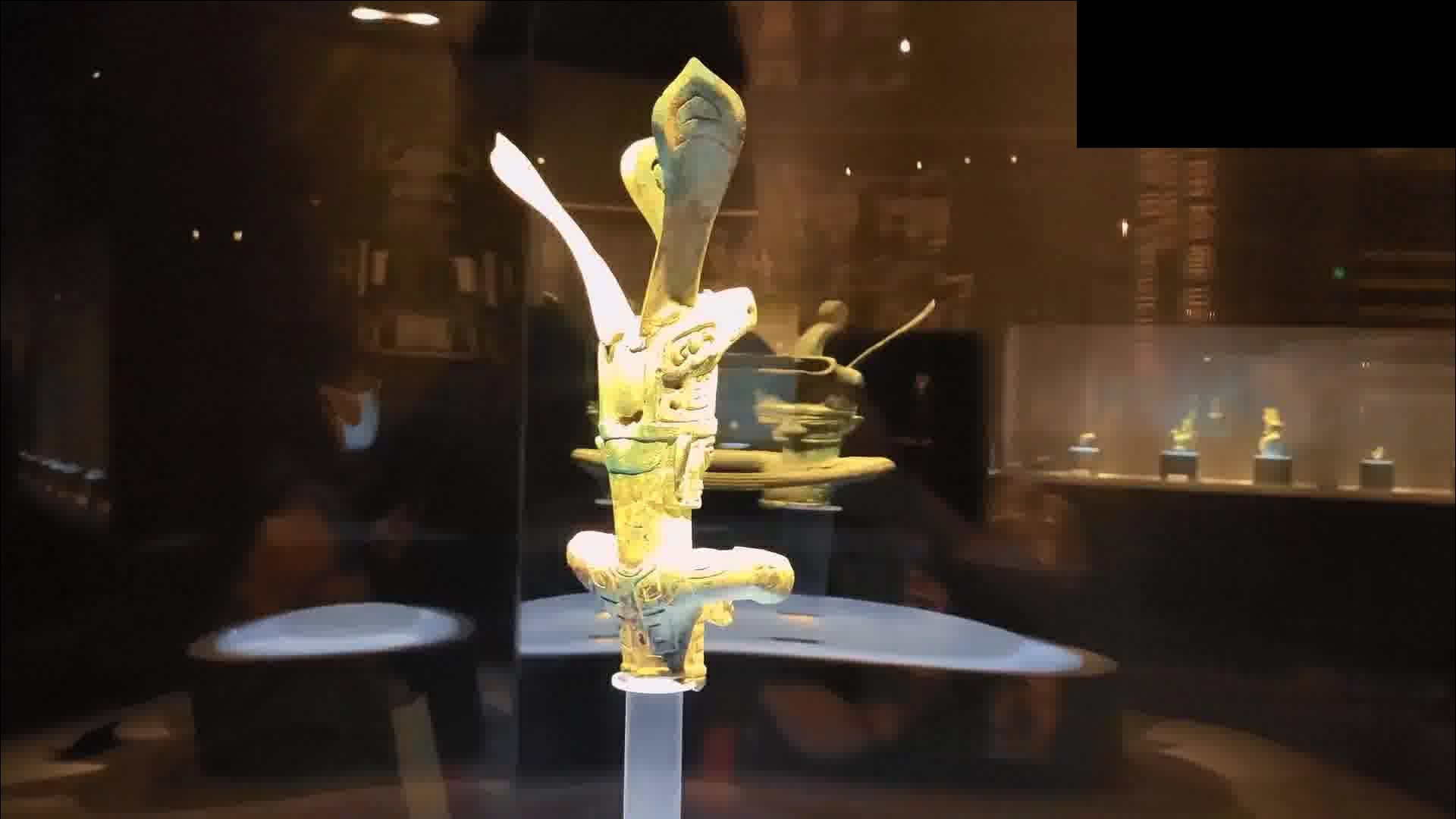}
        \\
        \raisebox{0.6\height}{\rotatebox{90}{Plate}}
		\includegraphics[width=0.19\textwidth]{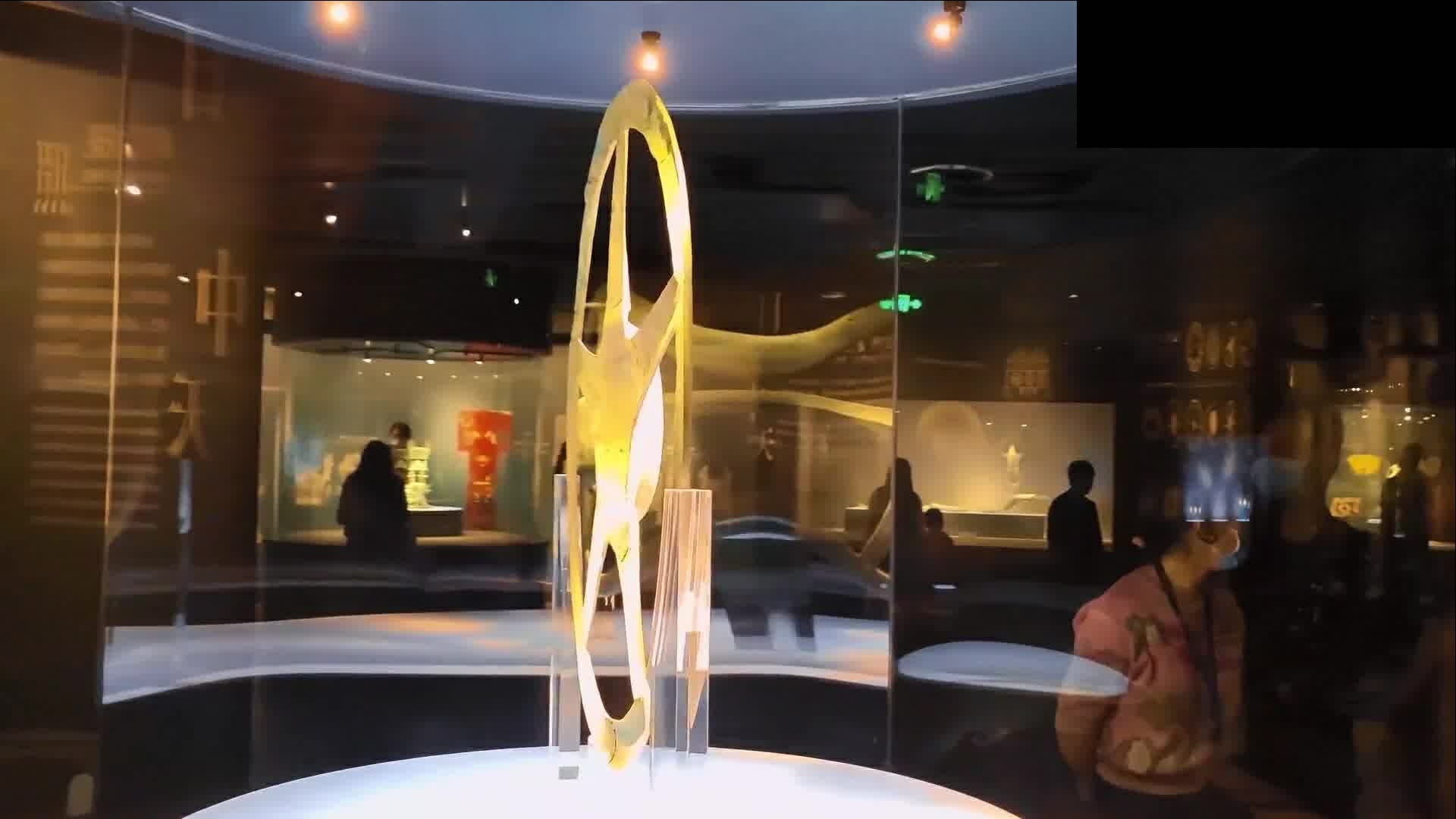}&
		\includegraphics[width=0.19\textwidth]{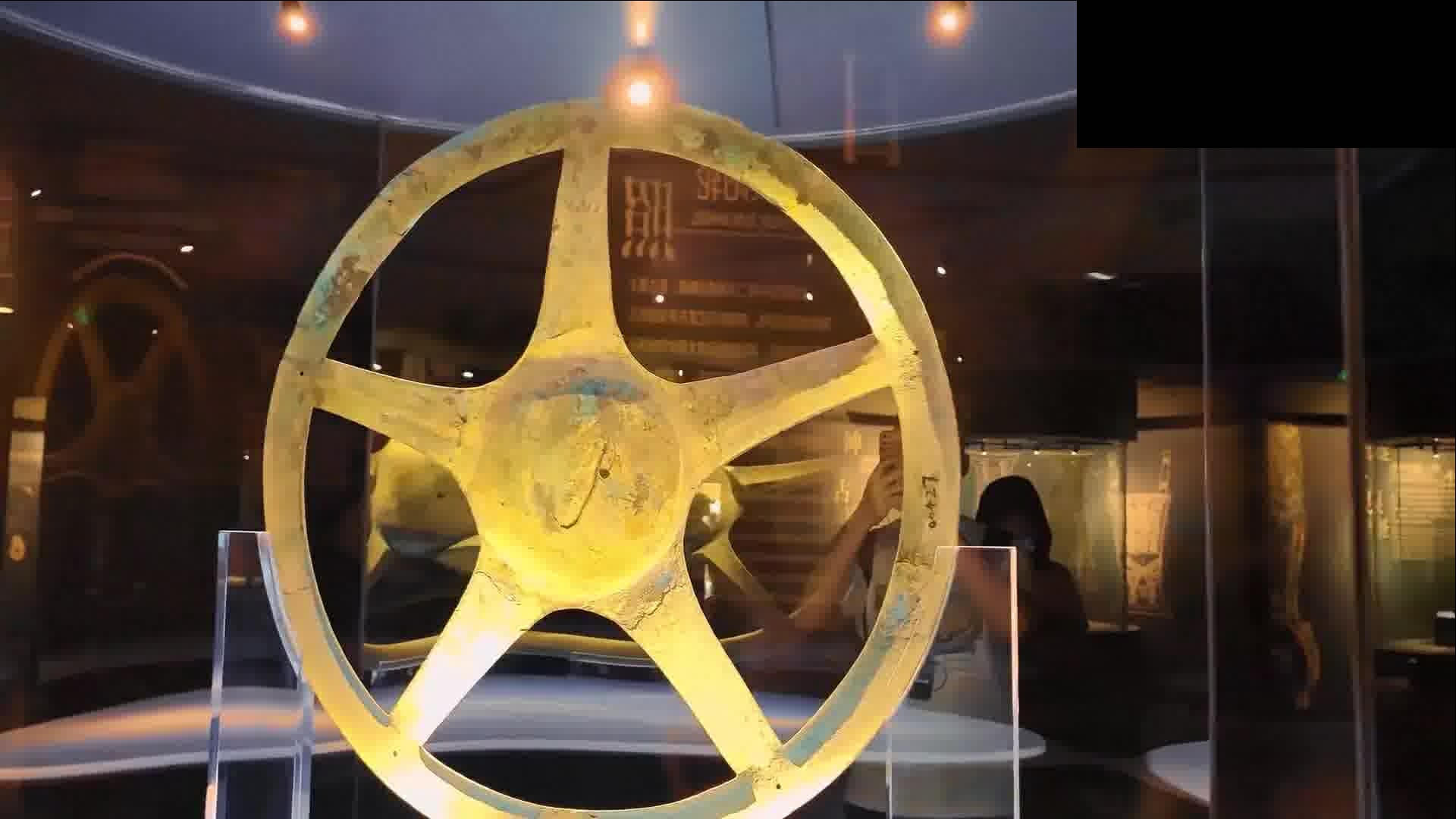}&
		\includegraphics[width=0.19\textwidth]{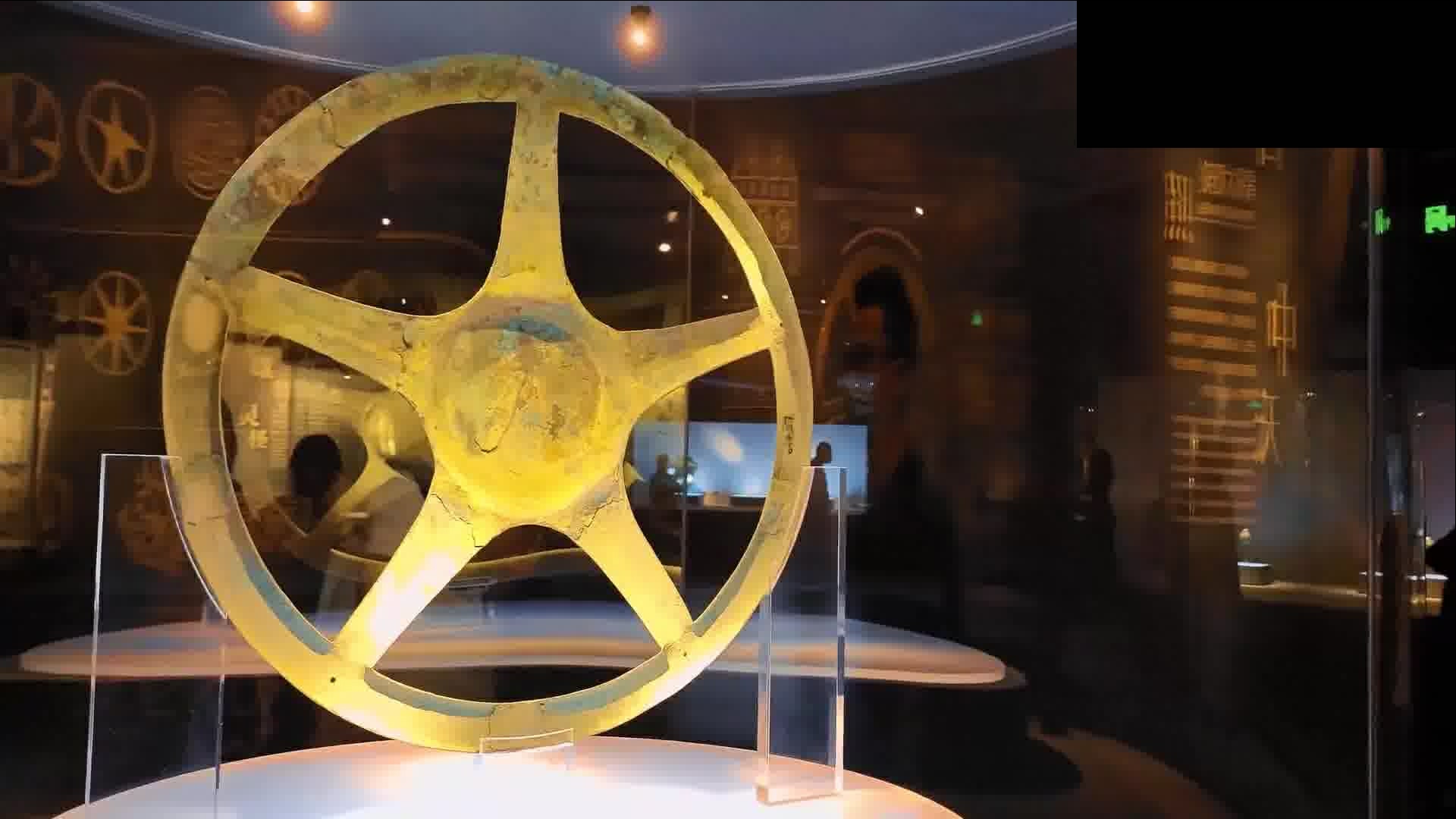}&
		\includegraphics[width=0.19\textwidth]{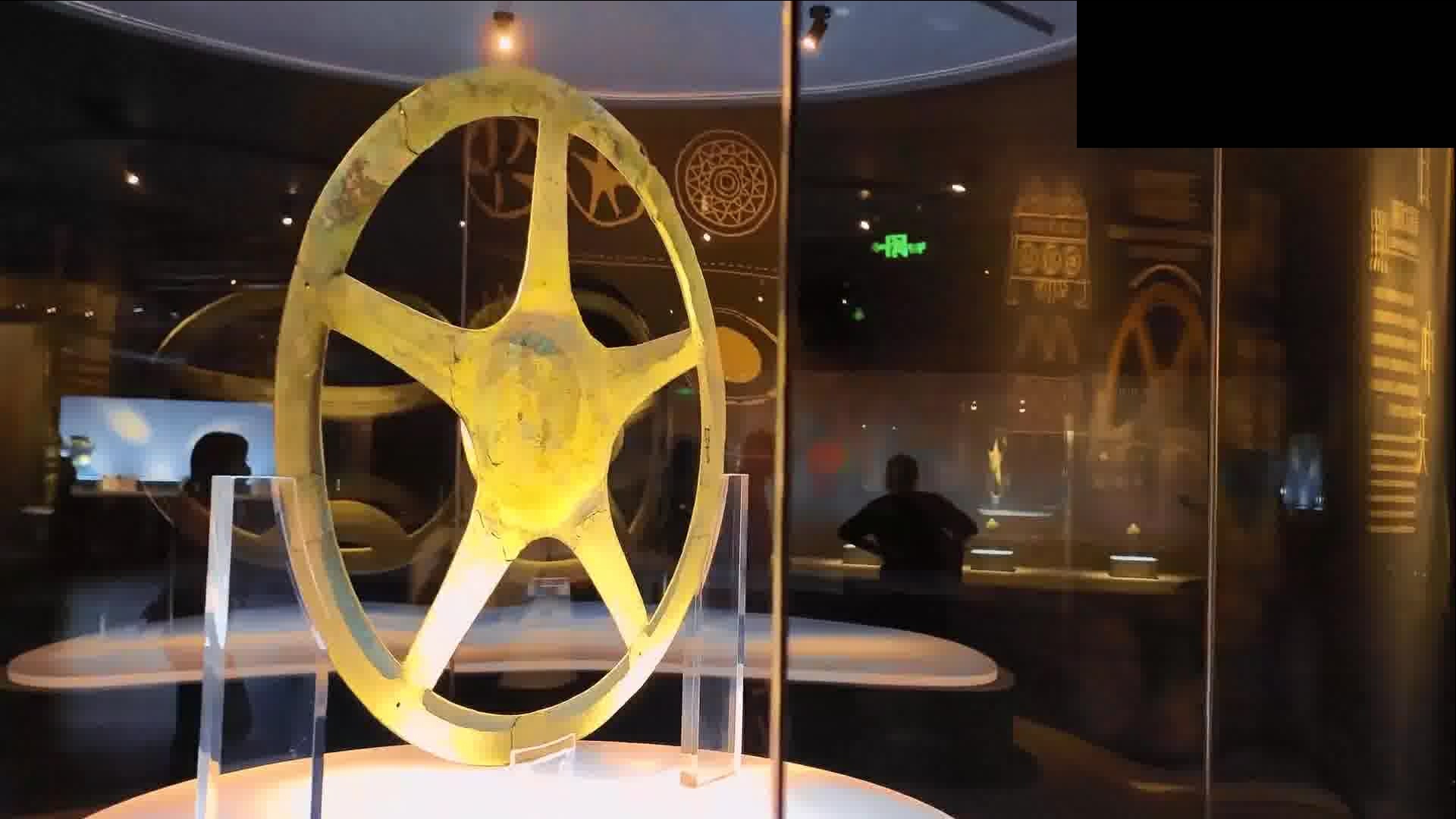}&
		\includegraphics[width=0.19\textwidth]{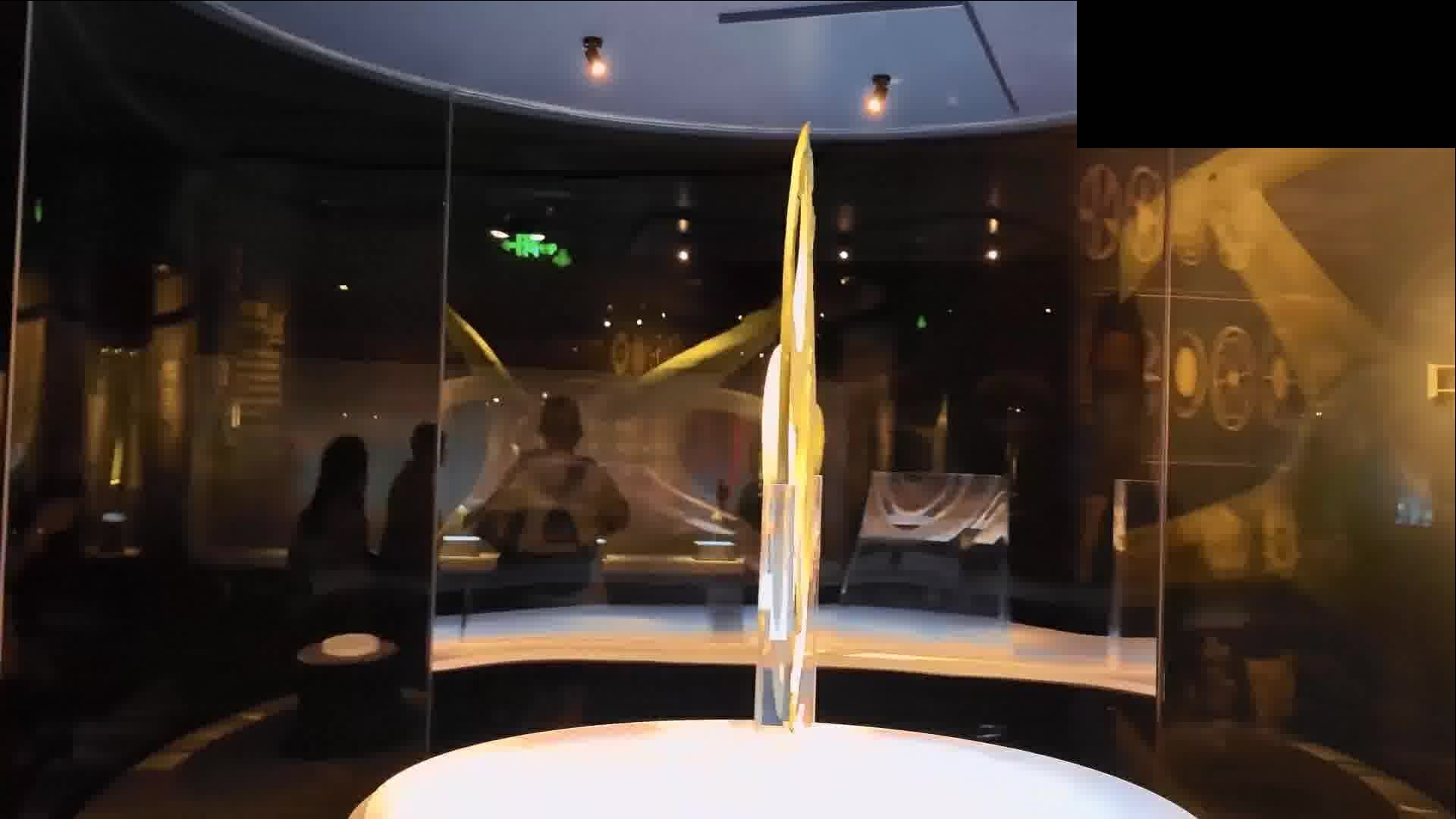}
		\\
        \raisebox{0.6\height}{\rotatebox{90}{Porcelain}}
		\includegraphics[width=0.1\textwidth]{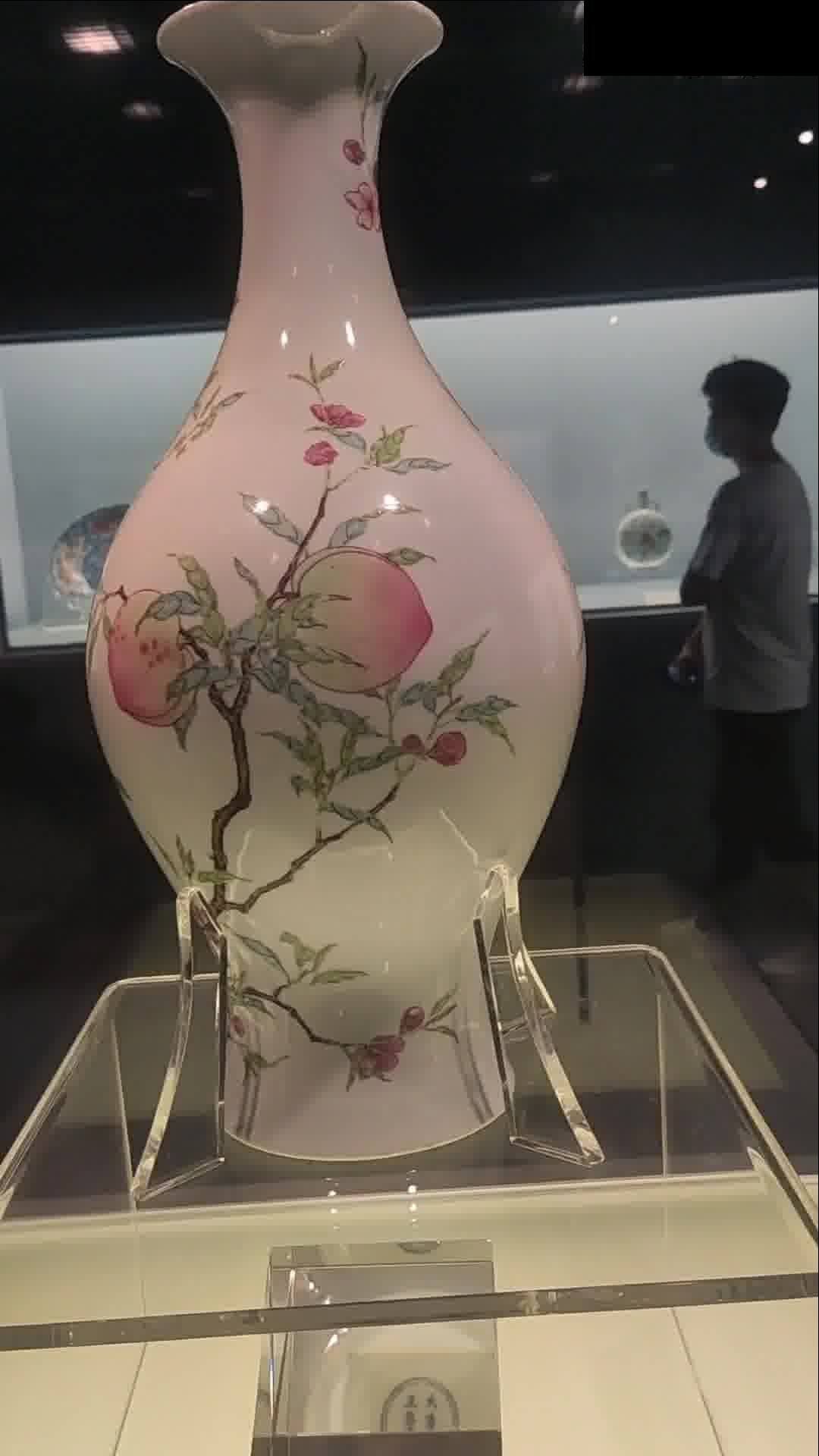}&
		\includegraphics[width=0.1\textwidth]{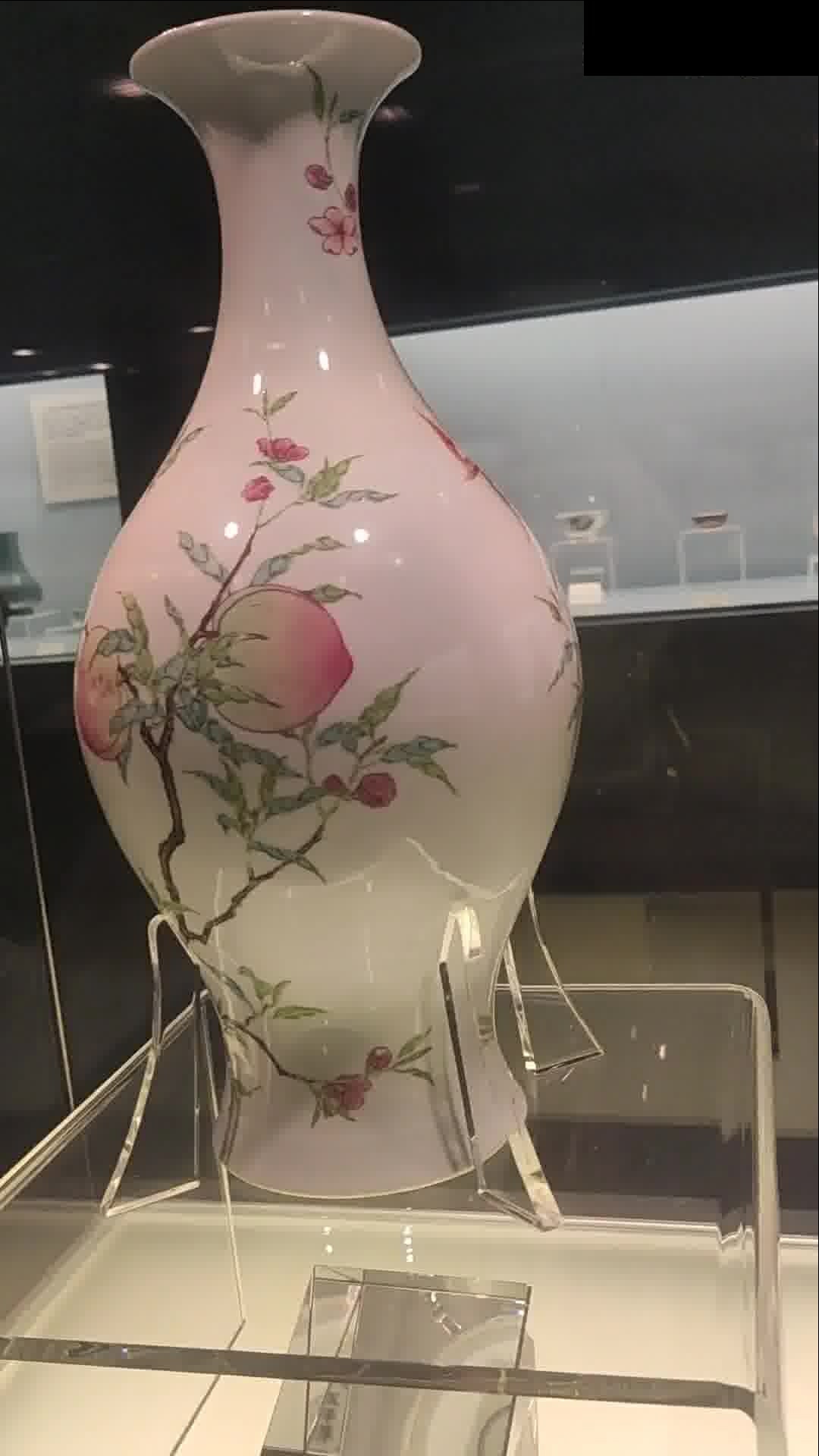}&
		\includegraphics[width=0.1\textwidth]{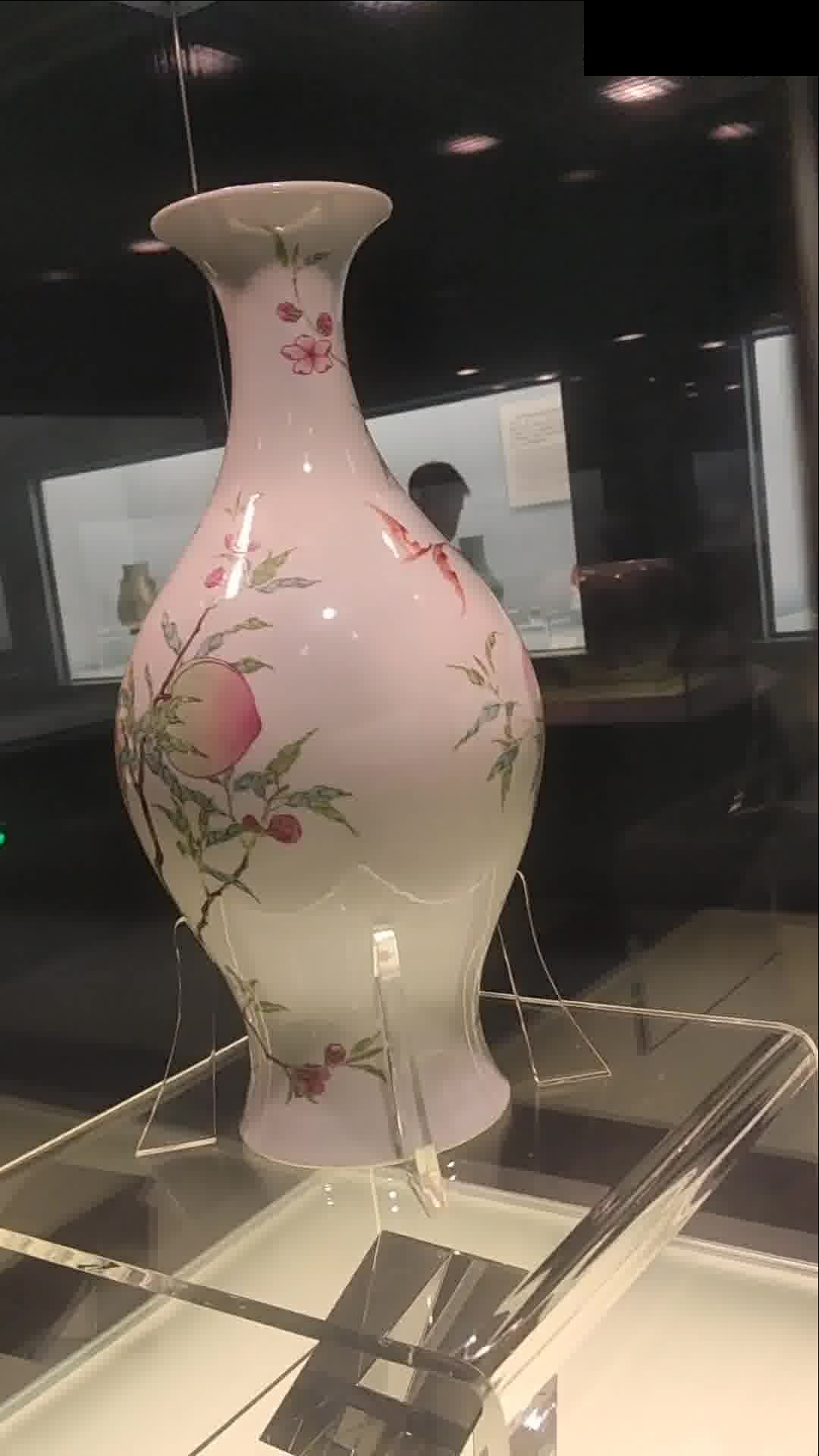}&
		\includegraphics[width=0.1\textwidth]{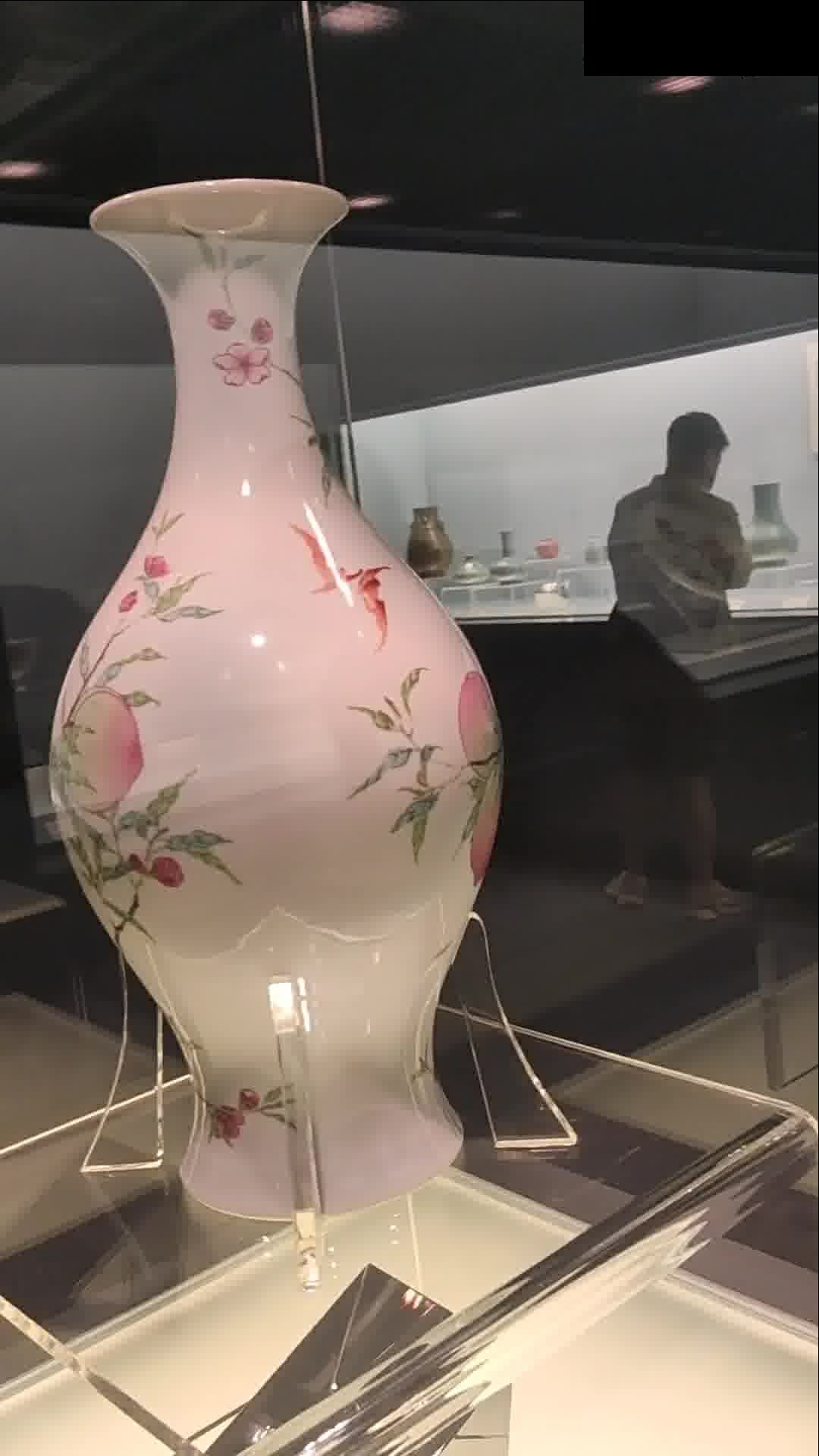}&
		\includegraphics[width=0.1\textwidth]{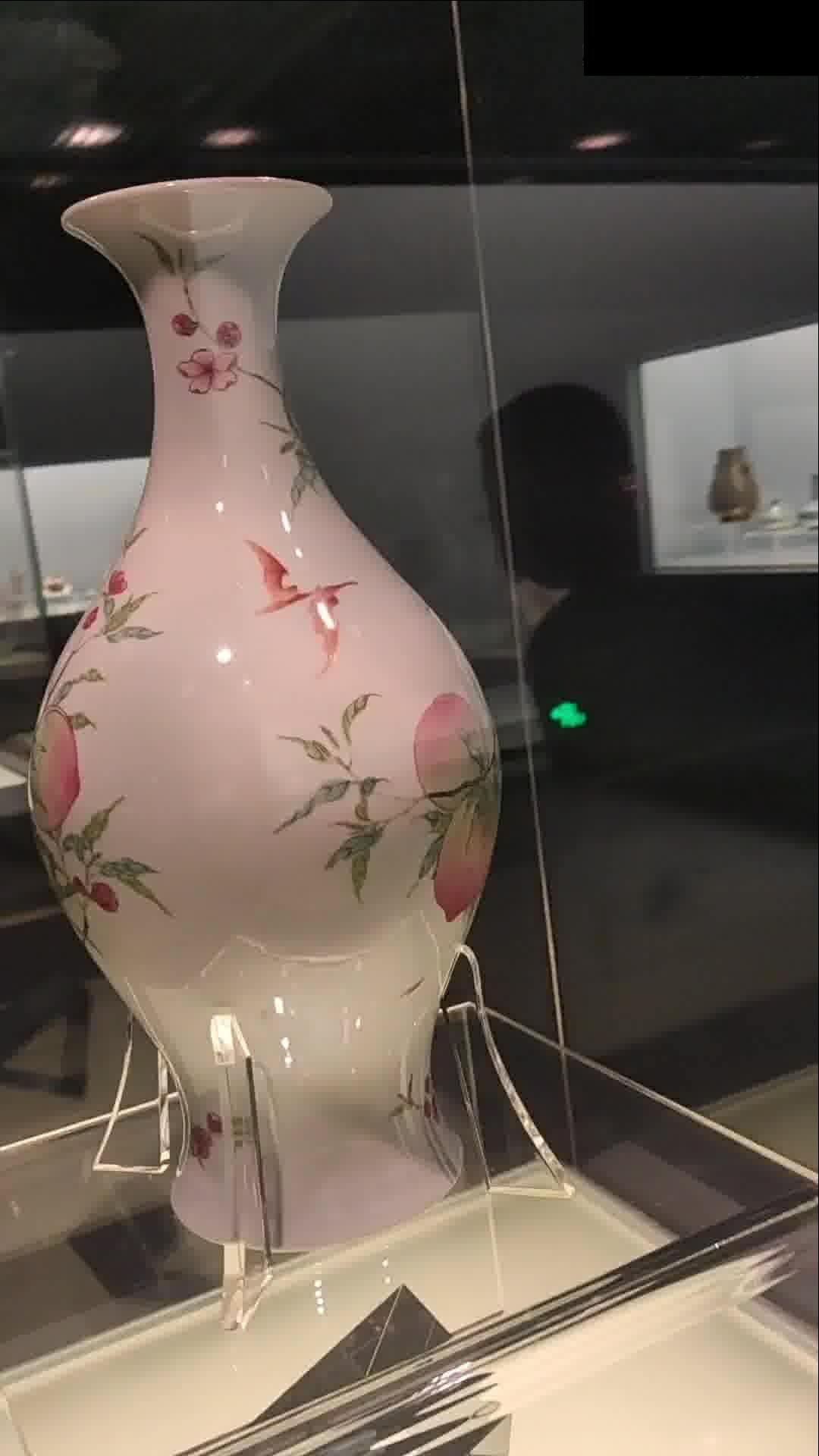}
		\\
        \raisebox{0.8\height}{\rotatebox{90}{Bronze}}
		\includegraphics[width=0.1\textwidth]{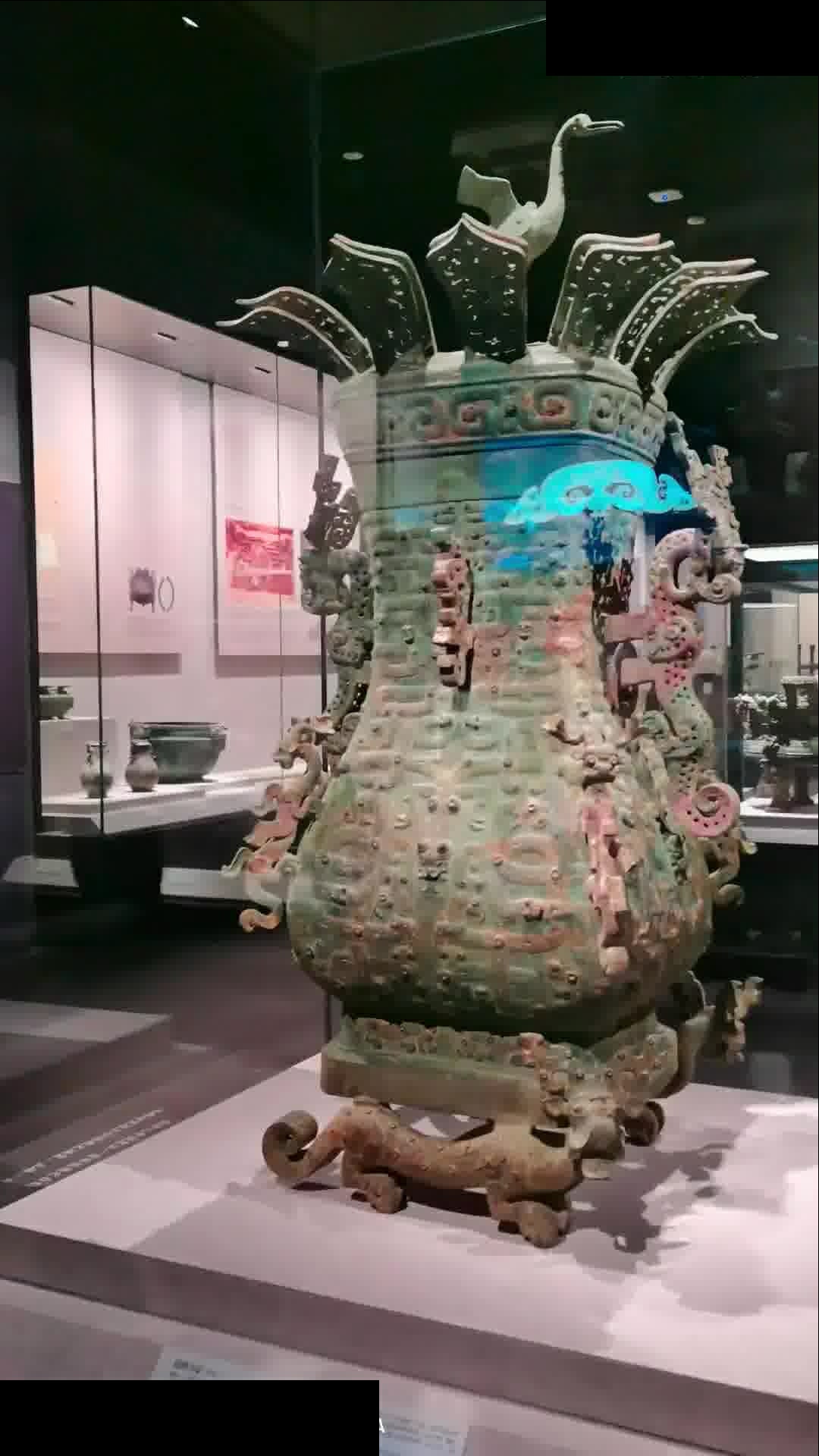}&
		\includegraphics[width=0.1\textwidth]{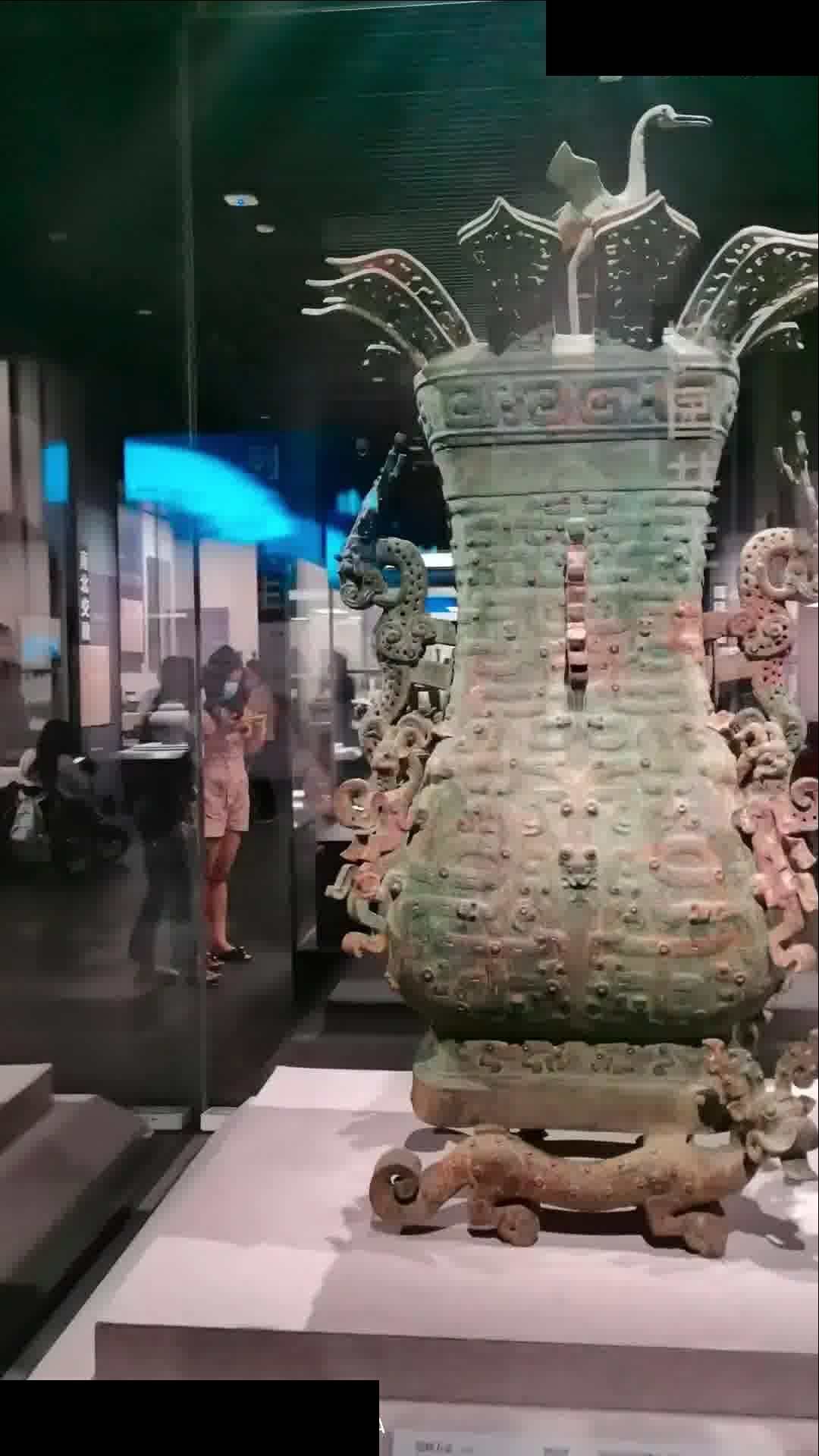}&
		\includegraphics[width=0.1\textwidth]{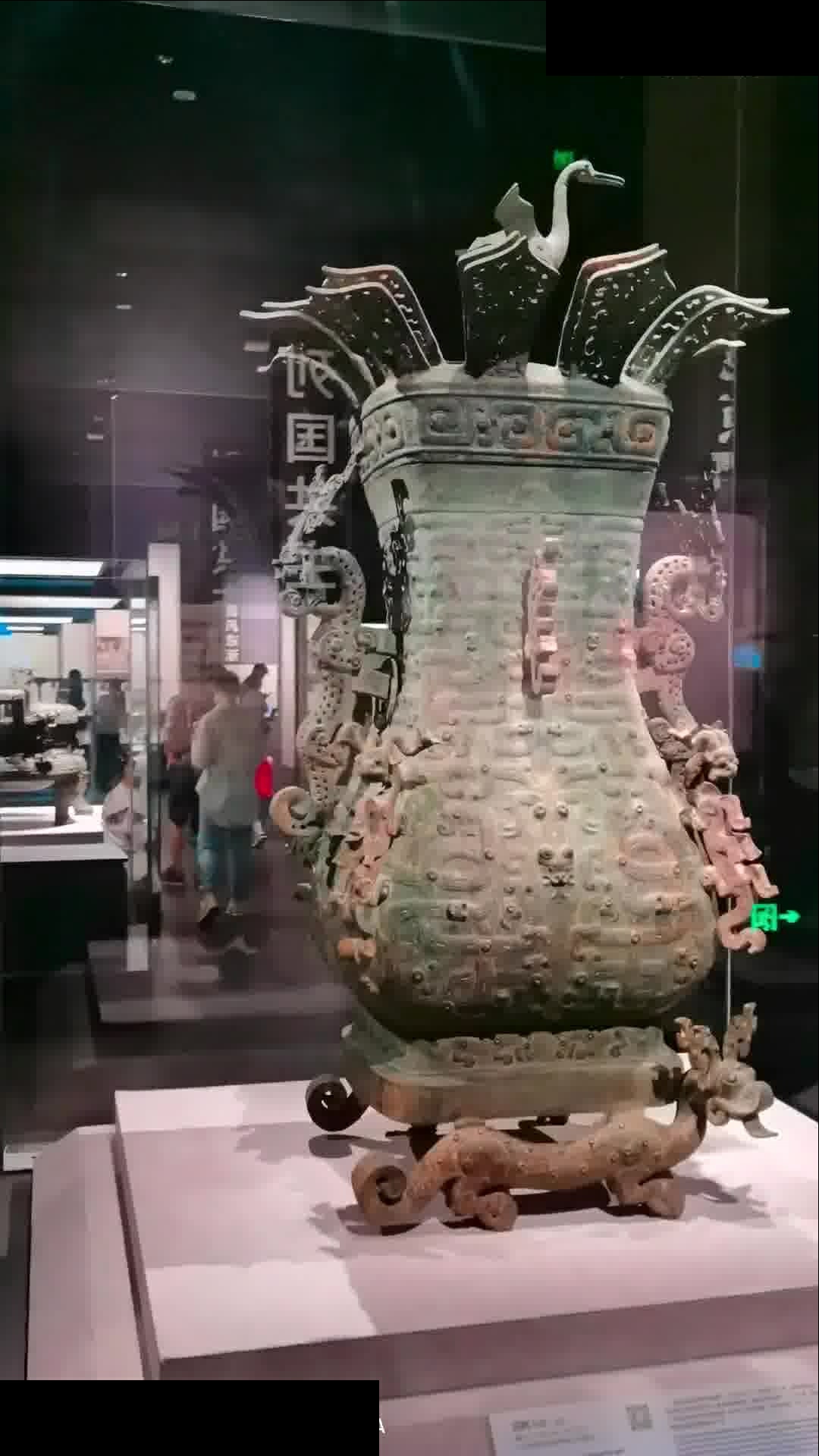}&
		\includegraphics[width=0.1\textwidth]{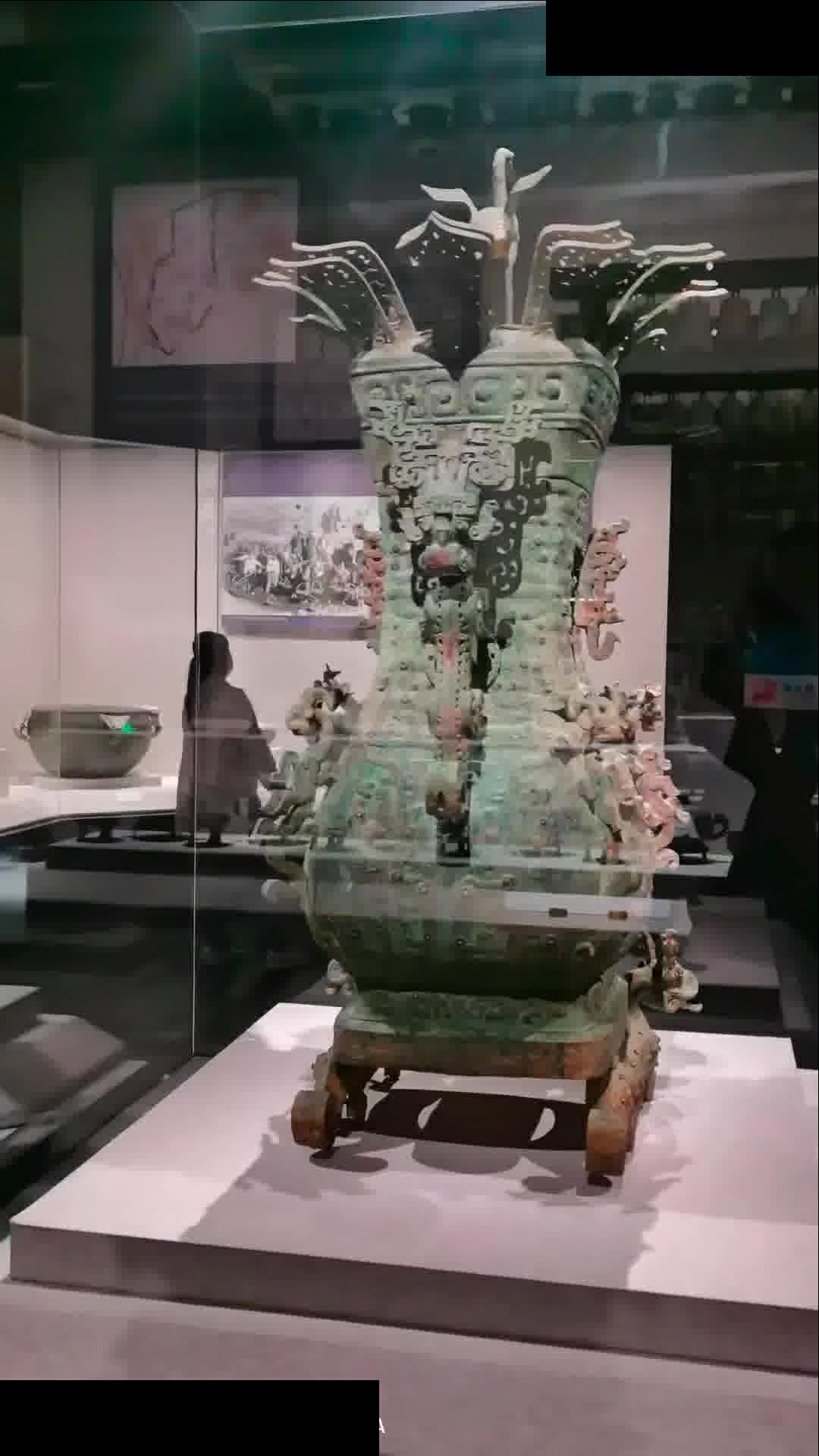}&
		\includegraphics[width=0.1\textwidth]{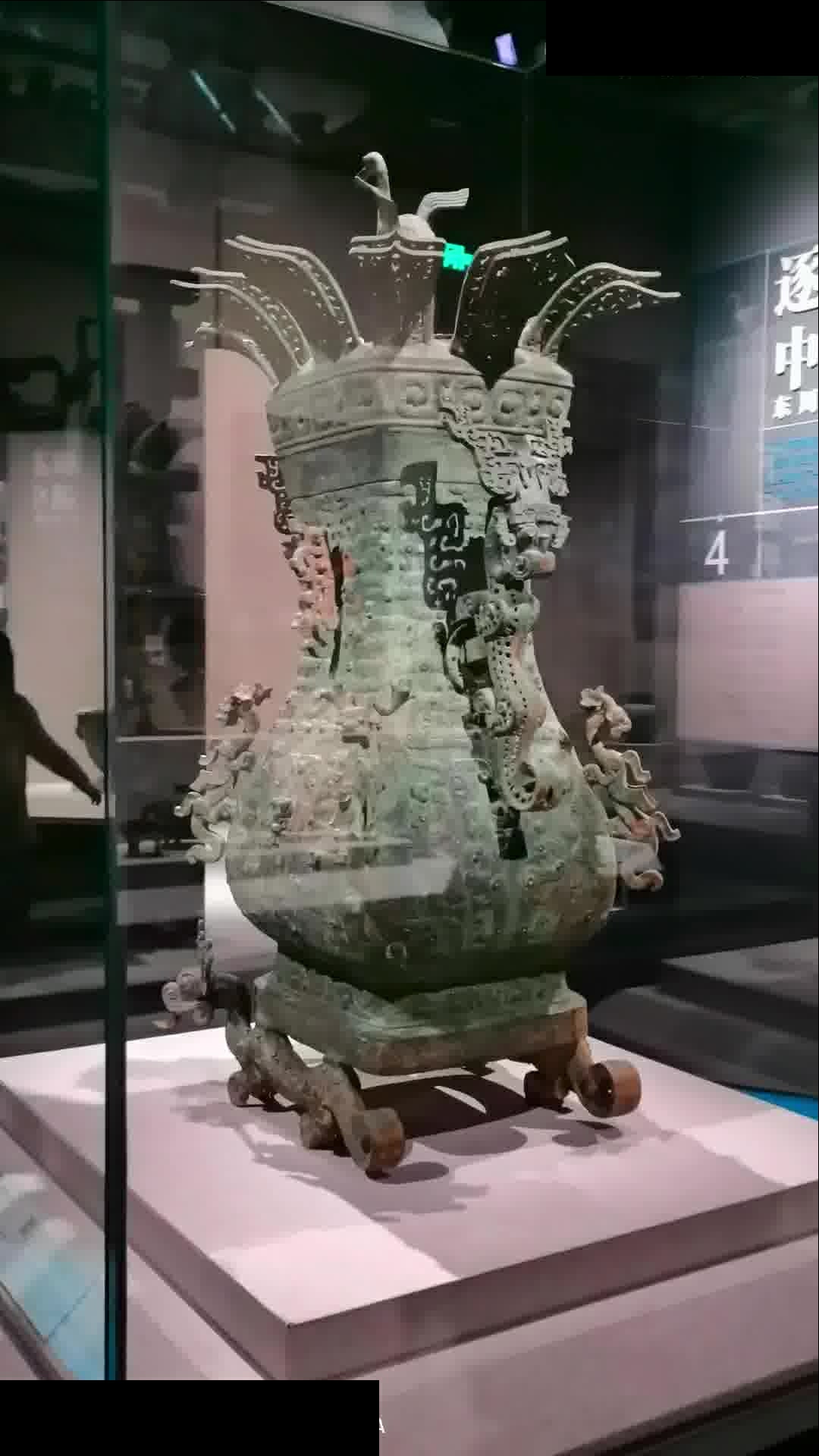}
	\end{tabular}
	\caption{Examples of the real-world dataset.}
	\label{exam_real}
\end{figure*}

\tabref{data} reports the metrics of our synthetic and real-world datasets. For the synthetic dataset, we set the kernel size of the Gaussian filter to $\mathrm{11}$ for generating the reflection effect.  We randomly pick a scene (`Scan114') of the DTU dataset~\cite{aanaes2016large} as the source of high specular reflections. We select $\mathrm{10}$ scenes from a total of 15 scenes on the DTU dataset~\cite{aanaes2016large} based on visual reality. They are: `Scan24', `Scan37', `Scan40', `Scan55', `Scan63', `Scan65', `Scan69', `Scan83', `Scan97', and `Scan105'. 
For the real-world dataset, we capture one scene (`Toys') and collect $\mathrm{5}$ scenes from the Internet: `Buddha'~$\footnote{https://www.bilibili.com/video/BV1M44y1z7XX}$, `Figure'~$\footnote{https://www.bilibili.com/video/BV1BP4y1Y7bV}$, `Plate'~$\footnote{https://www.bilibili.com/video/BV1BP4y1Y7bV}$, `Porcelain'~$\footnote{https://www.bilibili.com/video/BV1UP4y1h7tW}$ and `Bronze'~$\footnote{https://www.bilibili.com/video/BV1SU4y1E7QR}$. Examples of real-world scenes in our experiments are shown in \figref{exam_real}.

\subsection{Inference Time}
For object surface reconstruction, the inference time of NeuS-HSR is $\mathrm{36}$ seconds under resolution = $\mathrm{64}$ and threshold = $\mathrm{0.0}$. For rendering a novel view at the resolution of $\mathrm{800\times 600}$, NeuS-HSR takes around $\mathrm{96}$ seconds without the ground-truth mask on a single $\mathrm{NVIDIA~Tesla~V100~GPU}$.

\subsection{Baselines}
Because original neural implicit baselines are trained and tested on the datasets without HSR, we retrain all these models on each scene of our synthetic and real-world datasets. 

\myPara{NeuS~\cite{wang2021neus}.} 
To obtain the results of NeuS, we use their released official codes~$\footnote{https://github.com/Totoro97/NeuS}$ with the default setting in all scenes.

\myPara{UNISURF~\cite{oechsle2021unisurf}.}
To compare with UNISURF, we adopt their officially released codes~$\footnote{https://github.com/autonomousvision/unisurf}$ with the default setting in the synthetic scenes. 

\myPara{VolSDF~\cite{yariv2021volume}.}
To compare with VolSDF, we use their officially released codes~$\footnote{https://github.com/lioryariv/volsdf}$ with the default setting in all scenes.

\myPara{COLMAP~\cite{schoenberger2016mvs}.}
To obtain the results of COLMAP, we use the official command version of COLMAP~$\footnote{https://github.com/colmap/colmap}$ and run sequential commands provided in their documents~$\footnote{https://colmap.github.io/}$ in all scenes. 

\subsection{Q\&A}
\begin{figure}[t]
	\centering
	\begin{tabular}{*{4}{c@{\hspace{0px}}}}
		\includegraphics[width=0.24\linewidth]{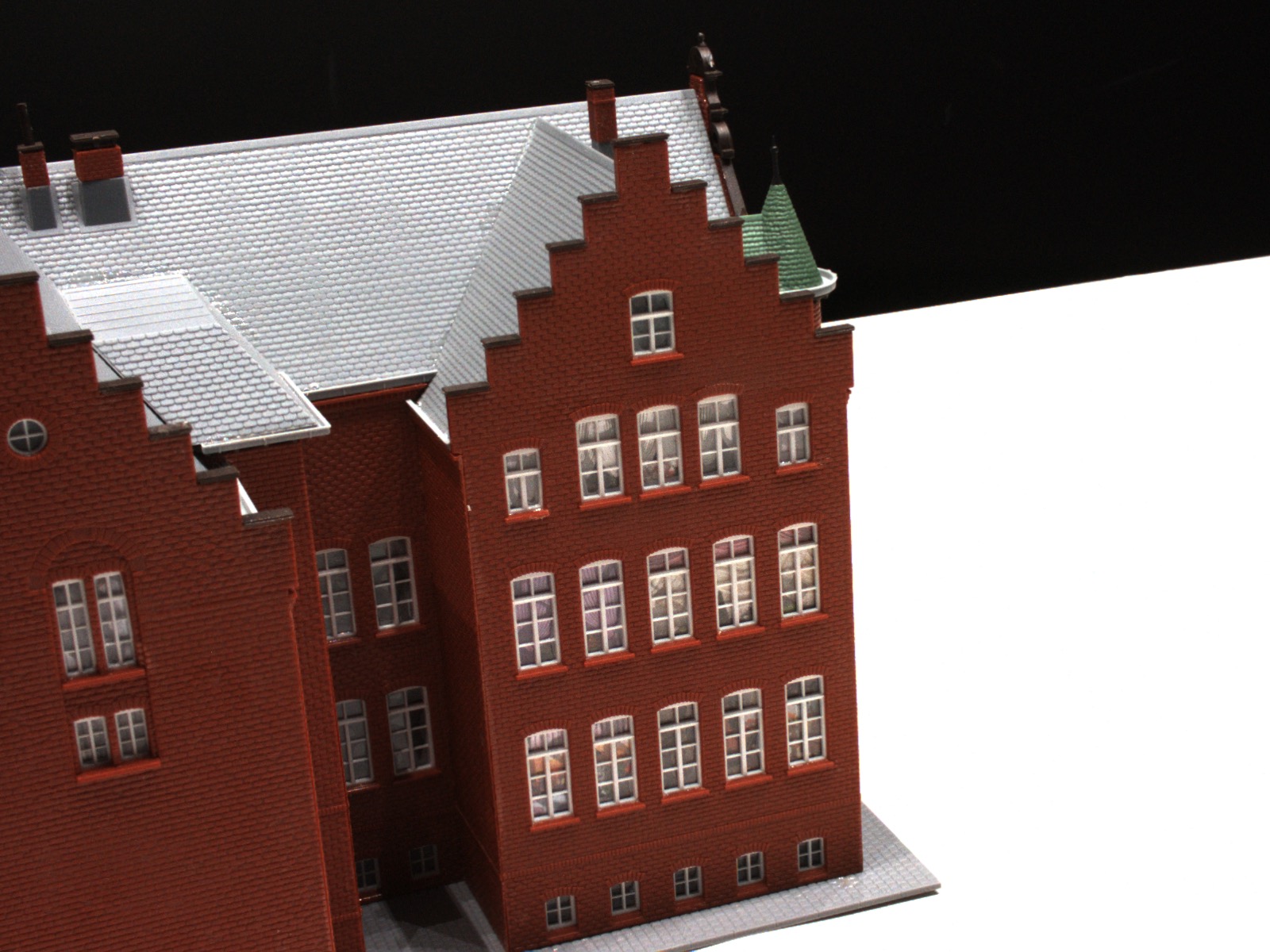} &  
		\includegraphics[width=0.24\linewidth]{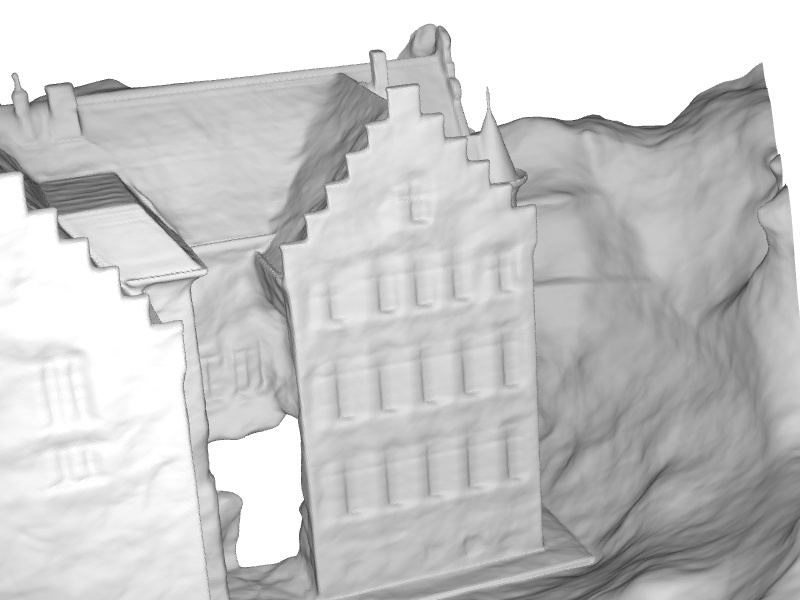}&
		\includegraphics[width=0.24\linewidth]{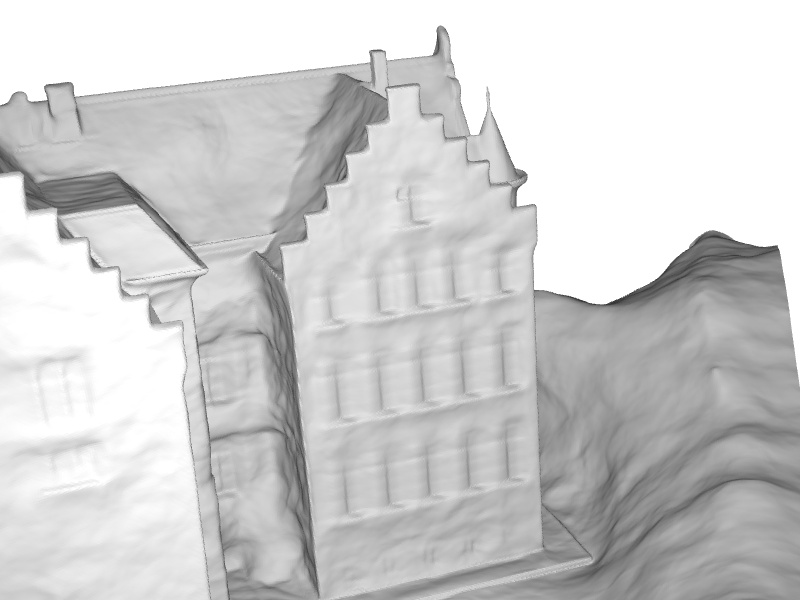}&
		\includegraphics[width=0.24\linewidth]{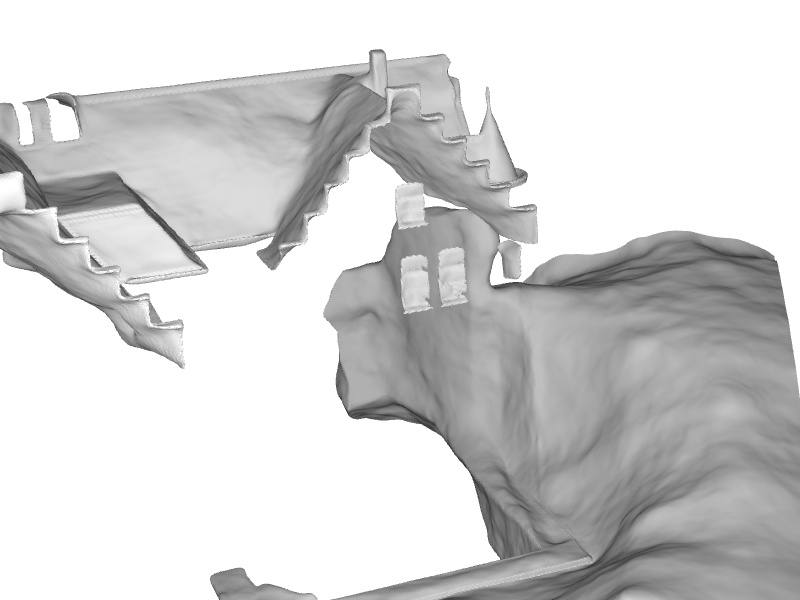}\\
		\small{(a) Supervision} & \small{(b) NeuS ($1.0$)} & \small{(c) Ours ($0.7$)} & \small{(d) Ours ($0.3$)}\\
	\end{tabular}
    \caption{Performance in a non-HSR scene.}
	\label{non_hsr}
\end{figure}

\begin{table}[t]
\centering
\footnotesize
\caption{Comparison of novel view synthesis on `Bronze'.}
\vspace*{-2mm}
\begin{tabular}{c|c|c|c}
\cline{1-4}
Method   & NeRF++~\cite{zhang2020nerf++} & NeuS~\cite{wang2021neus} & NeuS-HSR   \\ \cline{1-4}
PSNR$\uparrow$  & 15.92 & 15.51 & 15.93  \\ \cline{1-4}
SSIM$\uparrow$  & 0.480 & 0.489 & 0.502  \\ \cline{1-4}
\end{tabular}
\label{novel}
\end{table}
\begin{itemize}
\question{How about the quality of rendered images in novel views?}
\answer{The goal of our work is to reconstruct the target object against HSR accurately with multi-view images as supervision. We conduct an evaluation of novel view synthesis on `Bronze'. We select the first $\mathrm{3}$ images and the last $\mathrm{7}$ images of the sequential images for testing. The average scores of the test sets on PSNR and SSIM are present in \tabref{novel}.}

\question{Why use the same appearance function $F_c$ in two paths?}
\answer{Firstly, we use the same $F_c$ to save model parameters. Secondly, because the two paths of our framework are trained in one stage and the supervision is only the captured image, we adopt the same $F_c$ to separate the two appearances in the same domain. Lastly, we consider $F_c$ as an implicit function that maps 3D locations, normals, view directions, and neural features to color values. We use different 3D locations and normals from two paths as the input to acquire different color values by $F_c$.}

\question{Why is the auxiliary plane built in the camera coordinate system?}
\answer{Our model is trained by a view at each iteration, we build an auxiliary plane of each view in the camera coordinate system for simplification as \figref{supp_rt} shows. We transform the 3D locations to the camera coordinate system first, then we can apply the reflection transformation by the auxiliary plane directly.}

\question{How about the performance of NeuS-HSR in non-HSR scenes?}
\answer{NeuS-HSR is built on two physical assumptions in HSR scenes. In non-HSR scenes, we can set the ratio of the linear summation to $\mathrm{1.0}$, then NeuS-HSR degrades to NeuS and can achieve the same performance as NeuS. The qualitative results are shown in \figref{non_hsr}}

\end{itemize}

\section{Additional results on Synthetic Dataset}\label{Sec:mrsd}
\subsection{Signed Distance Fields}
\begin{figure*}[t]
	\centering
	\begin{tabular}{*{5}{c@{\hspace{1px}}}}
		\includegraphics[width=0.24\textwidth]{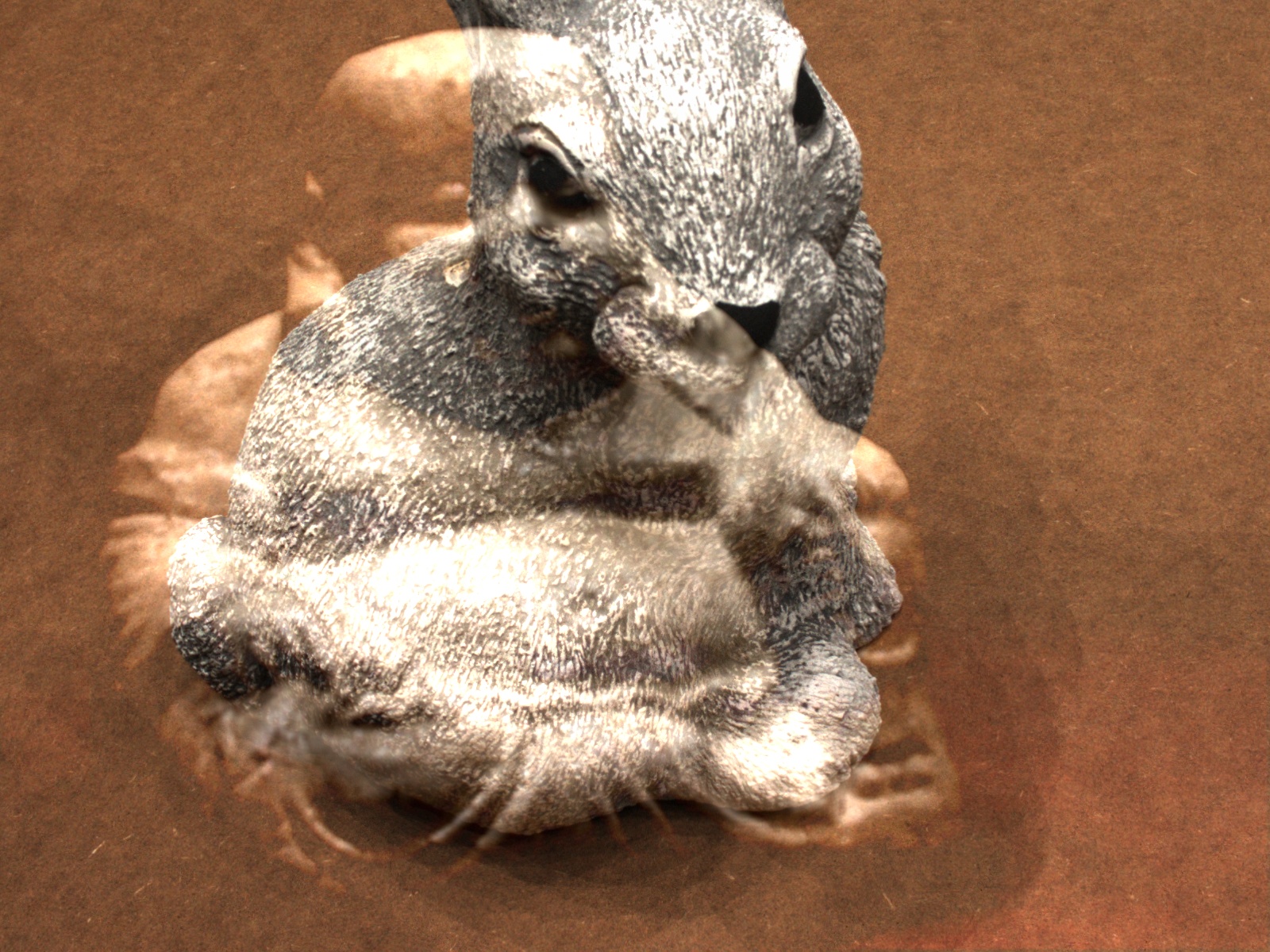}&
		\includegraphics[width=0.24\textwidth]{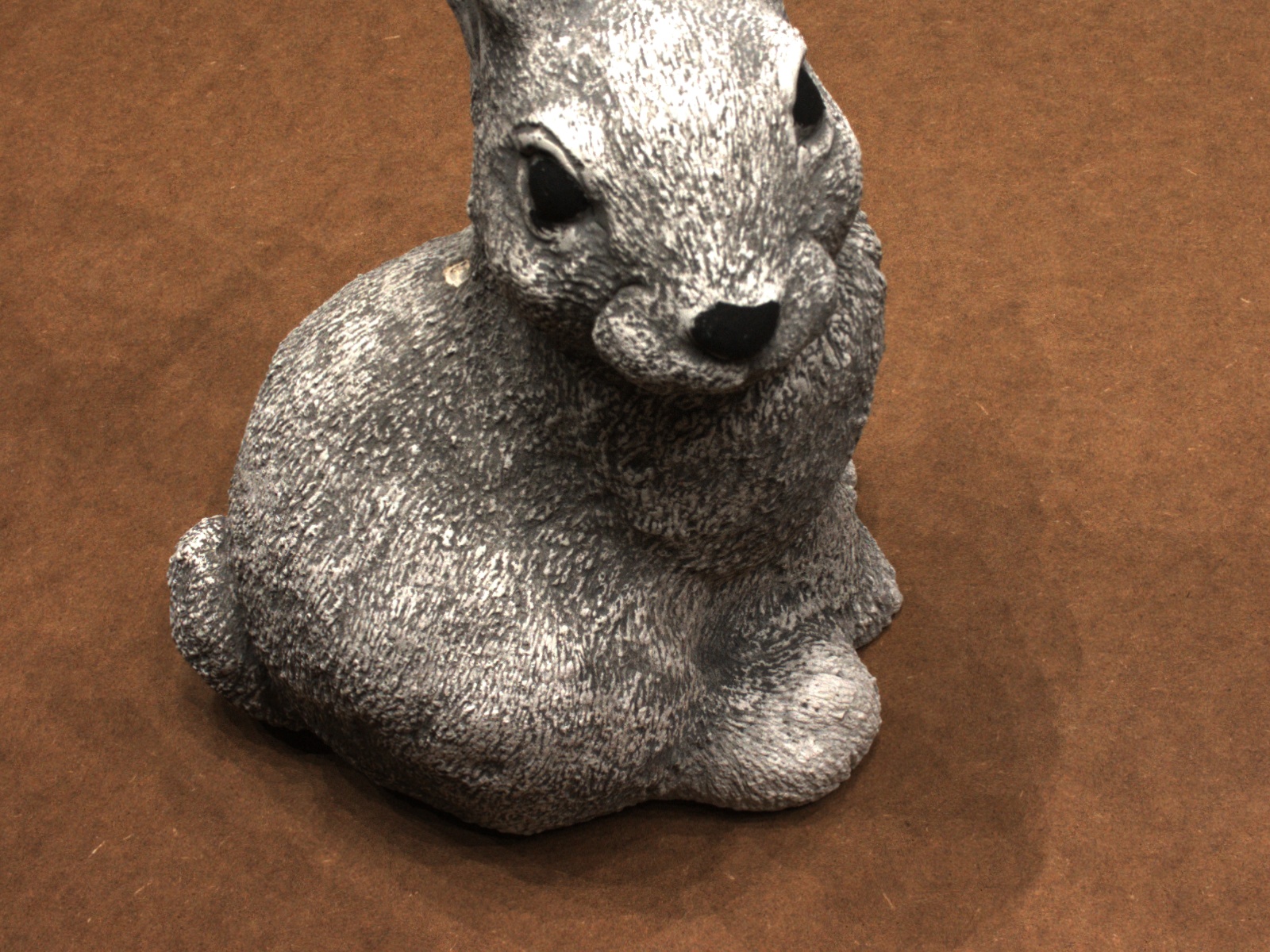}&
		\includegraphics[width=0.24\textwidth]{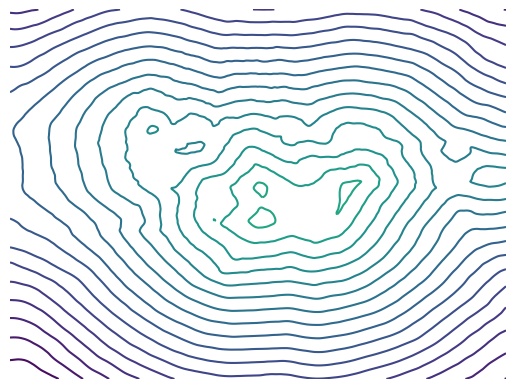}&
		\includegraphics[width=0.24\textwidth]{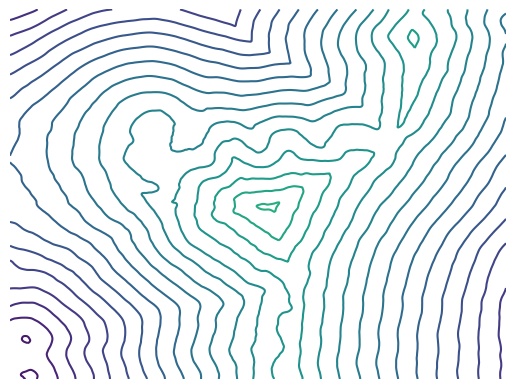}
		\\
		\raisebox{0.4\height}{Supervision}&
		\raisebox{0.4\height}{Reference}&
		\raisebox{0.4\height}{Ours}&
		\raisebox{0.4\height}{NeuS}&
	\end{tabular}
	\caption{Visualization of signed distance fields.}
	\label{supp_sdf}
\end{figure*}

We visualize the signed distance fields in \figref{supp_sdf}. Our model extracts a more accurate SDF of the scene than NeuS according to the distribution of signed distance fields. Specifically, our signed distance fields present the geometric characteristics of Bunny's tangent plane.

\subsection{Components}
\begin{figure*}[!htbp]
	\centering
	\begin{tabular}{*{7}{c@{\hspace{2px}}}}
	    \raisebox{0.5\height}{\rotatebox{90}{Scan24}}
		\includegraphics[width=0.15\textwidth]{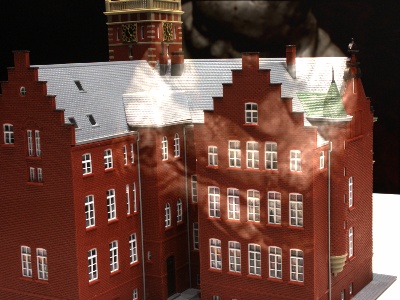}&
		\includegraphics[width=0.15\textwidth]{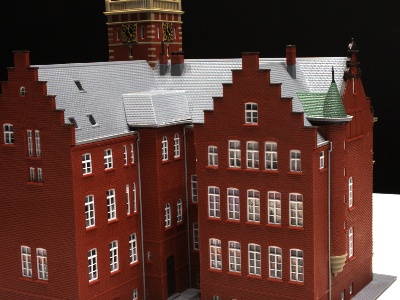}&
		\includegraphics[width=0.15\textwidth]{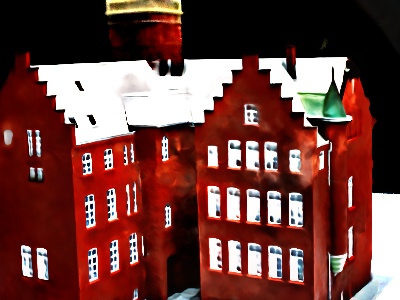}&
		\includegraphics[width=0.15\textwidth]{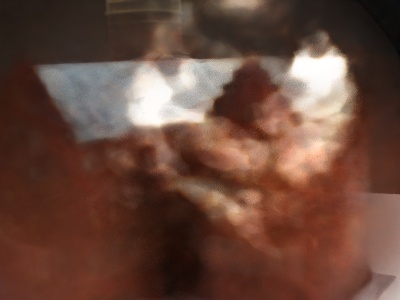}&
		\includegraphics[width=0.15\textwidth]{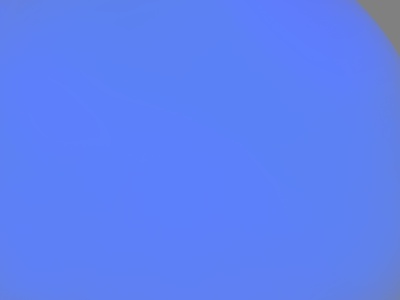}&
		\includegraphics[width=0.15\textwidth]{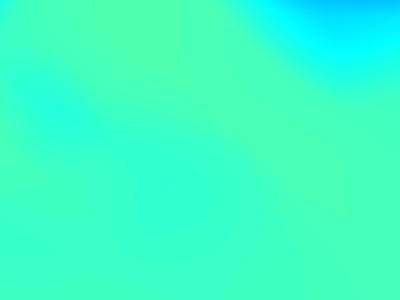}
        \\
        \raisebox{0.5\height}{\rotatebox{90}{Scan37}}
		\includegraphics[width=0.15\textwidth]{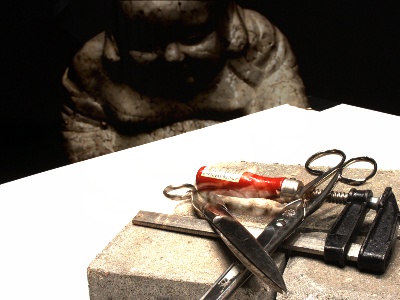}&
		\includegraphics[width=0.15\textwidth]{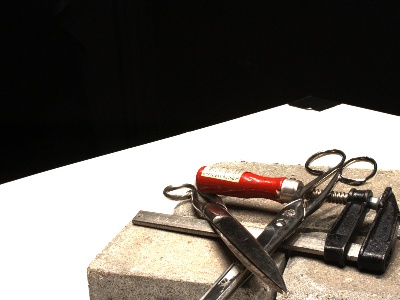}&
		\includegraphics[width=0.15\textwidth]{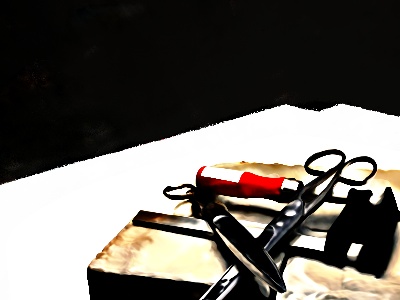}&
		\includegraphics[width=0.15\textwidth]{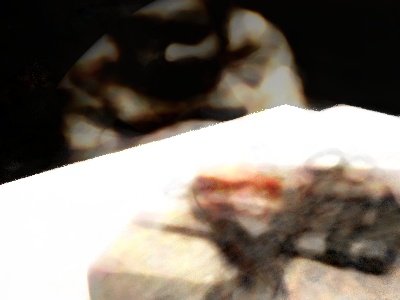}&
		\includegraphics[width=0.15\textwidth]{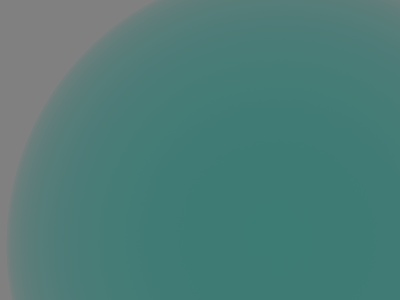}&
		\includegraphics[width=0.15\textwidth]{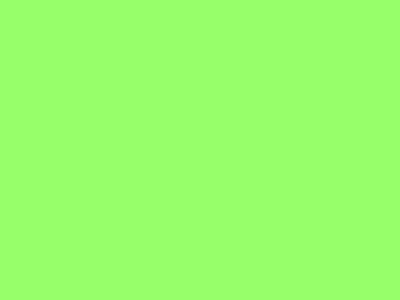}
        \\
        \raisebox{0.5\height}{\rotatebox{90}{Scan40}}
		\includegraphics[width=0.15\textwidth]{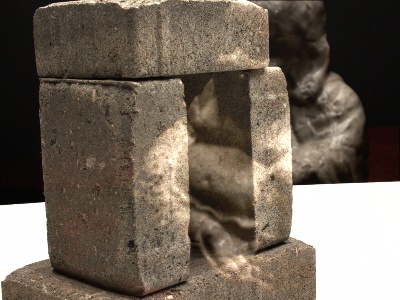}&
		\includegraphics[width=0.15\textwidth]{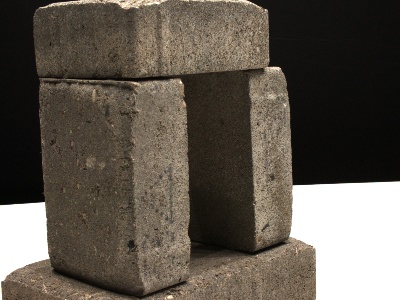}&
		\includegraphics[width=0.15\textwidth]{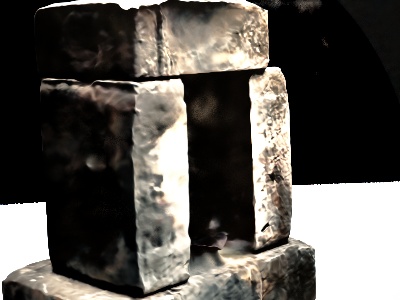}&
		\includegraphics[width=0.15\textwidth]{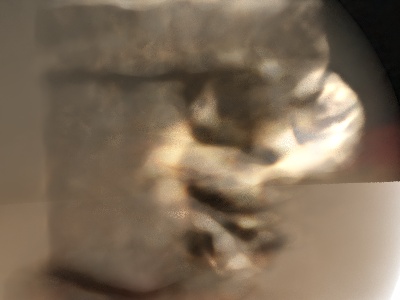}&
		\includegraphics[width=0.15\textwidth]{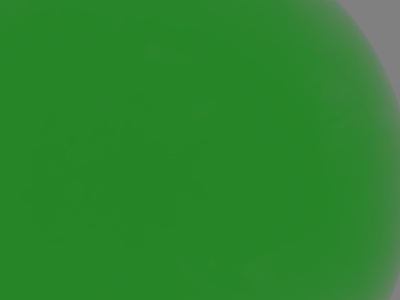}&
		\includegraphics[width=0.15\textwidth]{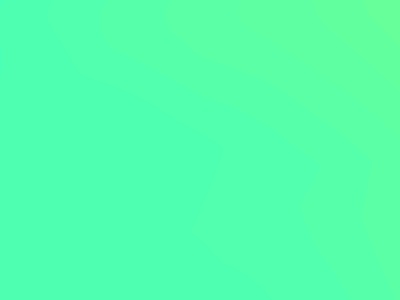}
        \\
        \raisebox{0.5\height}{\rotatebox{90}{Scan55}}
		\includegraphics[width=0.15\textwidth]{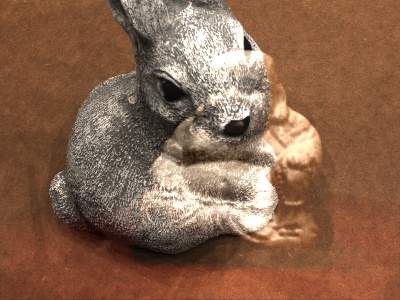}&
		\includegraphics[width=0.15\textwidth]{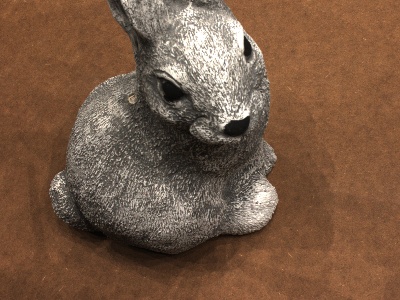}&
		\includegraphics[width=0.15\textwidth]{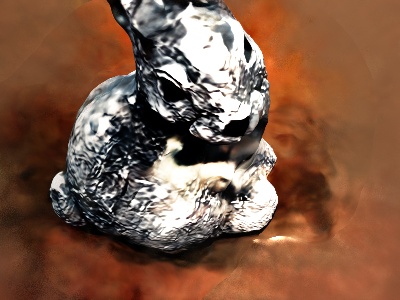}&
		\includegraphics[width=0.15\textwidth]{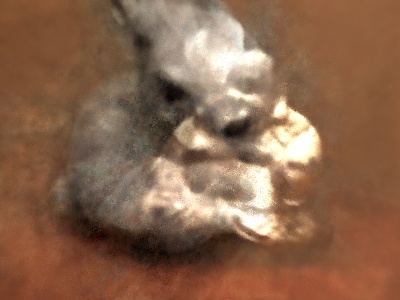}&
		\includegraphics[width=0.15\textwidth]{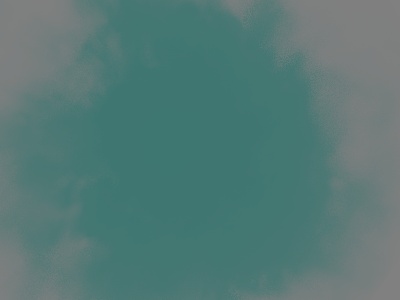}&
		\includegraphics[width=0.15\textwidth]{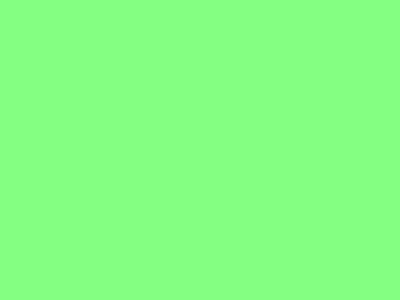}
        \\
        \raisebox{0.5\height}{\rotatebox{90}{Scan63}}
		\includegraphics[width=0.15\textwidth]{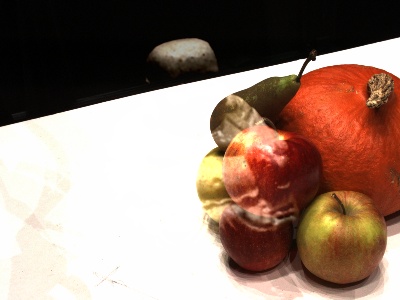}&
		\includegraphics[width=0.15\textwidth]{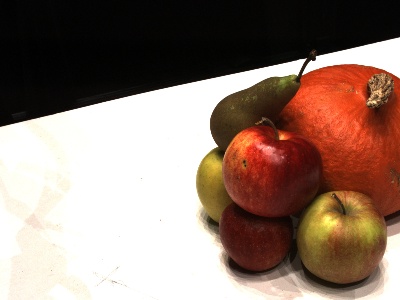}&
		\includegraphics[width=0.15\textwidth]{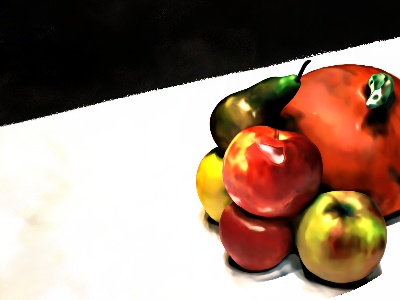}&
		\includegraphics[width=0.15\textwidth]{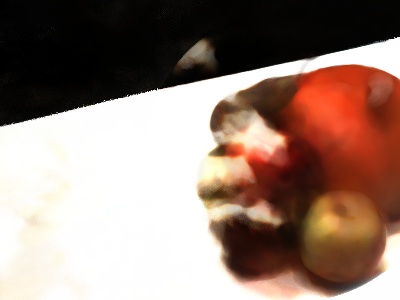}&
		\includegraphics[width=0.15\textwidth]{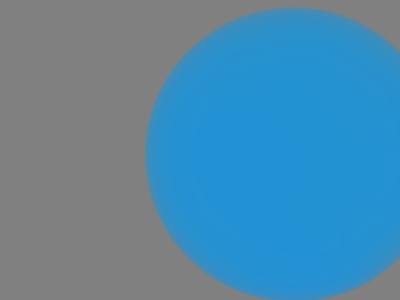}&
		\includegraphics[width=0.15\textwidth]{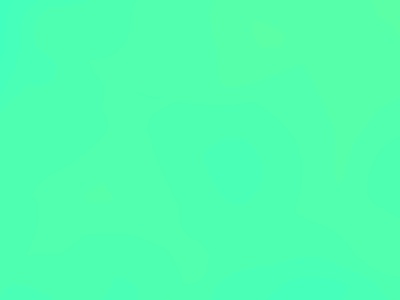}
        \\
        \raisebox{0.5\height}{\rotatebox{90}{Scan65}}
		\includegraphics[width=0.15\textwidth]{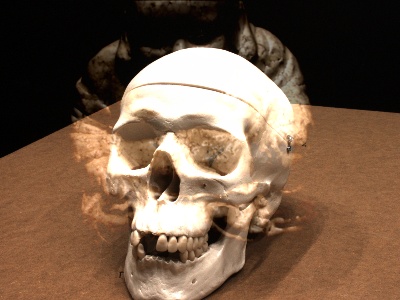}&
		\includegraphics[width=0.15\textwidth]{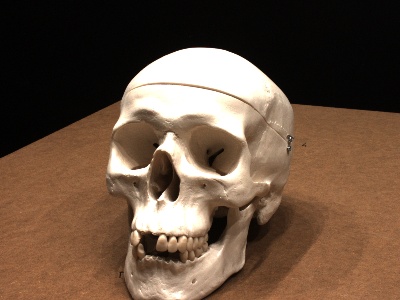}&
		\includegraphics[width=0.15\textwidth]{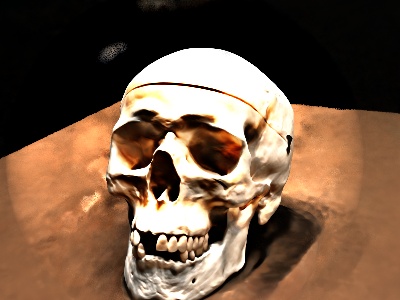}&
		\includegraphics[width=0.15\textwidth]{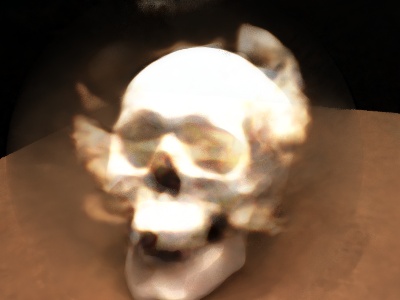}&
		\includegraphics[width=0.15\textwidth]{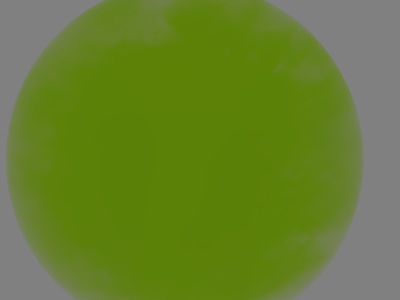}&
		\includegraphics[width=0.15\textwidth]{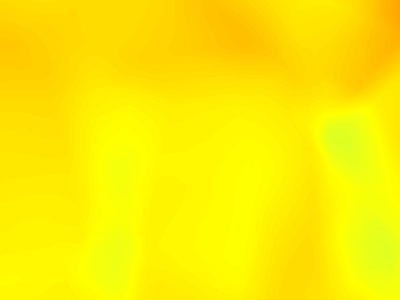}
		\\
		\raisebox{0.5\height}{\rotatebox{90}{Scan69}}
		\includegraphics[width=0.15\textwidth]{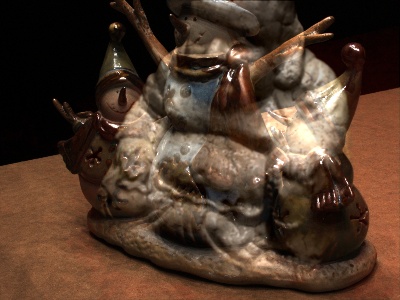}&
		\includegraphics[width=0.15\textwidth]{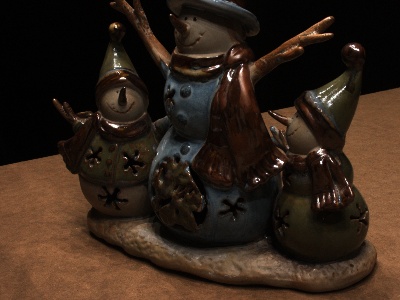}&
		\includegraphics[width=0.15\textwidth]{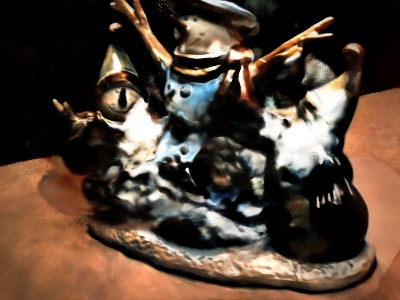}&
		\includegraphics[width=0.15\textwidth]{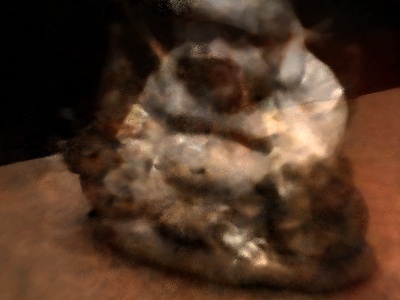}&
		\includegraphics[width=0.15\textwidth]{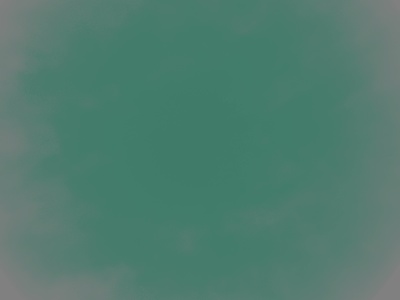}&
		\includegraphics[width=0.15\textwidth]{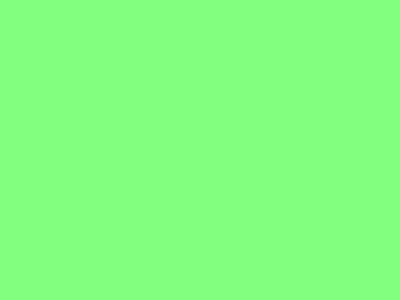}
		\\
		\raisebox{0.5\height}{\rotatebox{90}{Scan83}}
		\includegraphics[width=0.15\textwidth]{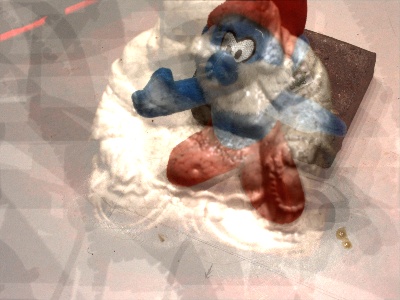}&
		\includegraphics[width=0.15\textwidth]{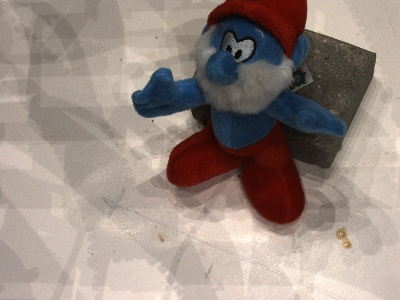}&
		\includegraphics[width=0.15\textwidth]{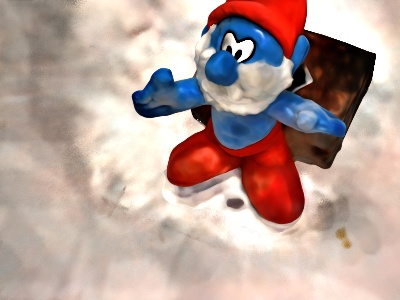}&
		\includegraphics[width=0.15\textwidth]{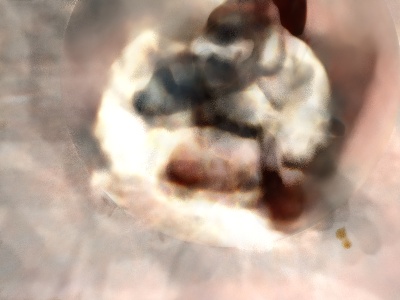}&
		\includegraphics[width=0.15\textwidth]{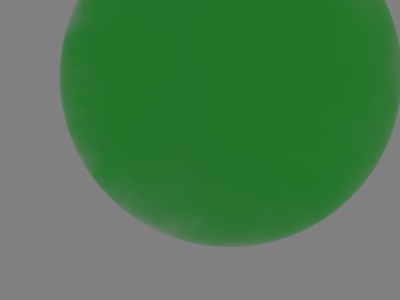}&
		\includegraphics[width=0.15\textwidth]{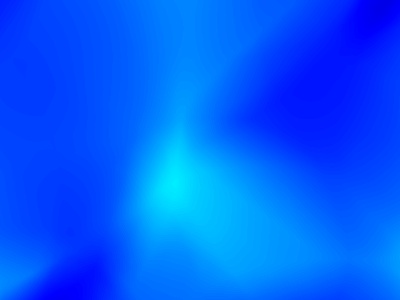}
		\\
		\raisebox{0.5\height}{\rotatebox{90}{Scan97}}
		\includegraphics[width=0.15\textwidth]{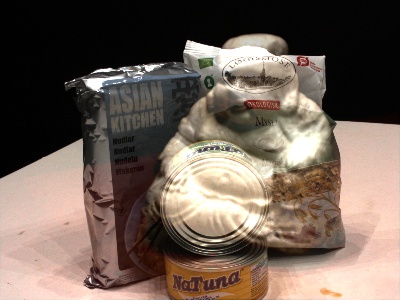}&
		\includegraphics[width=0.15\textwidth]{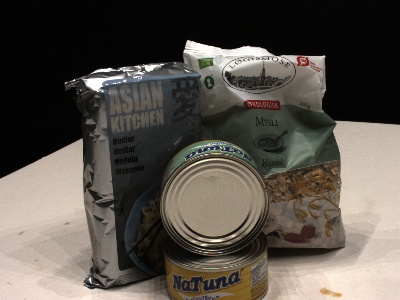}&
		\includegraphics[width=0.15\textwidth]{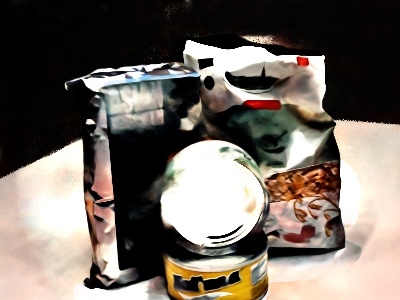}&
		\includegraphics[width=0.15\textwidth]{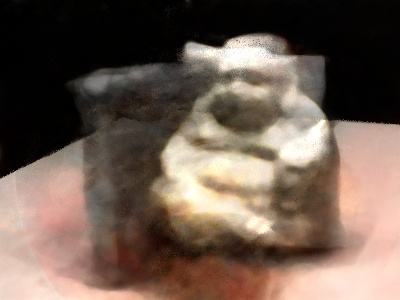}&
		\includegraphics[width=0.15\textwidth]{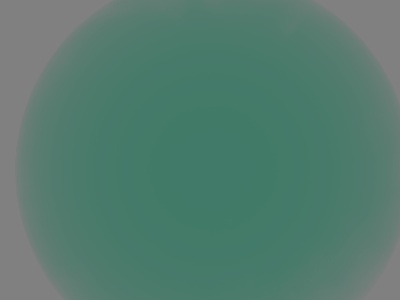}&
		\includegraphics[width=0.15\textwidth]{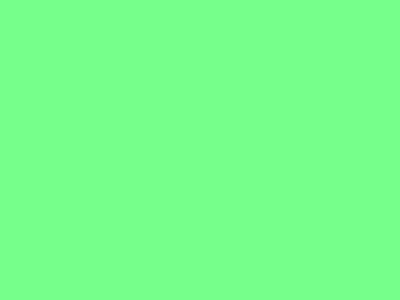}
		\\
		\raisebox{0.5\height}{\rotatebox{90}{Scan105}}
		\includegraphics[width=0.15\textwidth]{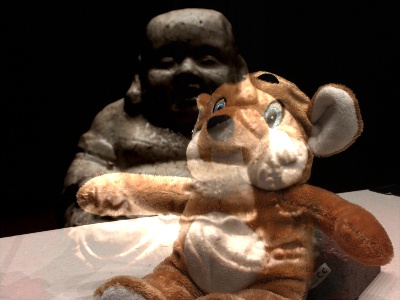}&
		\includegraphics[width=0.15\textwidth]{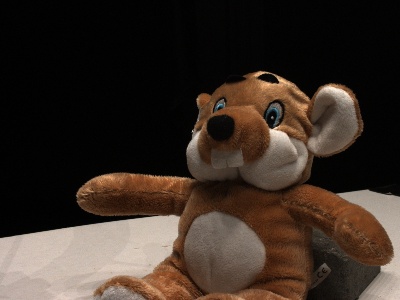}&
		\includegraphics[width=0.15\textwidth]{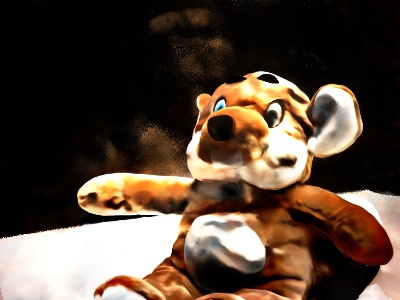}&
		\includegraphics[width=0.15\textwidth]{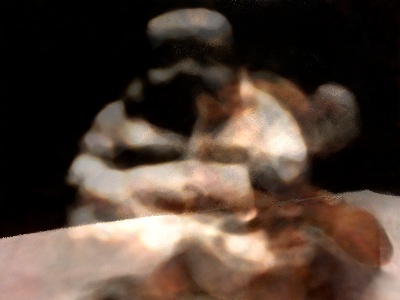}&
		\includegraphics[width=0.16\textwidth]{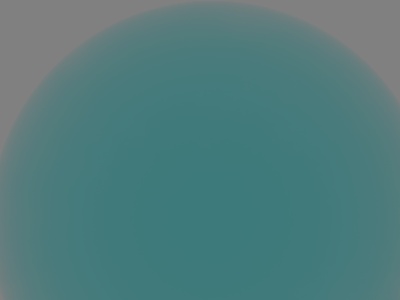}&
		\includegraphics[width=0.15\textwidth]{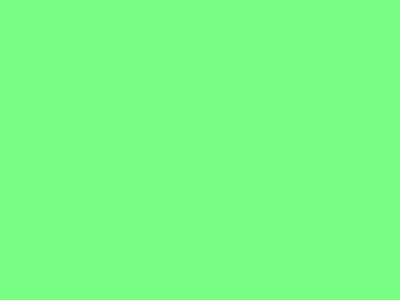}
		\\
		\raisebox{0.4\height}{Supervision}&
		\raisebox{0.4\height}{Reference}&
		\raisebox{0.4\height}{TOA}&
		\raisebox{0.4\height}{APA}&
		\raisebox{0.4\height}{Plane normal}&
		\raisebox{0.4\height}{Position}
	\end{tabular}
	\vspace*{-2mm}
	\caption{Components of NeuS-HSR on the synthetic dataset. `TOA': Target object appearance. `APA': Auxiliary plane appearance.}
	\label{supp_syn}
\end{figure*}

\figref{supp_syn} shows the components of NeuS-HSR on the synthetic dataset. Our method faithfully enhances the target object appearance and preserves HSR in the auxiliary plane appearance without any priors. Besides, the plane normal and position of each auxiliary plane on a camera ray in a view, indicate that the auxiliary planes tend to be a planar reflector. Hence, \figref{supp_syn} illustrates that our model achieves the physical decomposition of HSR scenes.


\subsection{Comparisons}
\begin{figure*}[!htbp]
	\centering
	\begin{tabular}{*{7}{c@{\hspace{1px}}}}
	    \raisebox{0.5\height}{\rotatebox{90}{Scan40}}
		\includegraphics[width=0.16\textwidth]{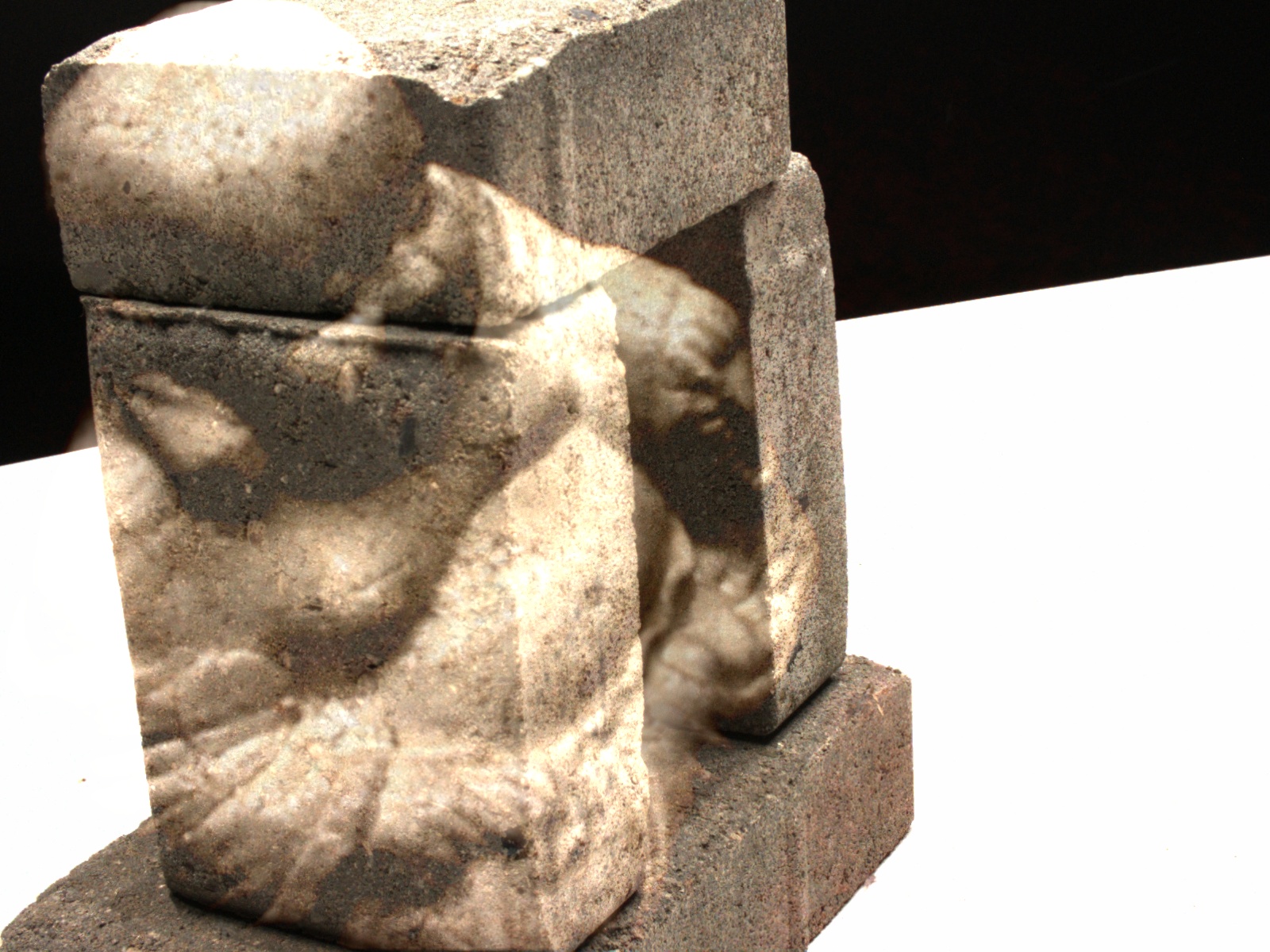}&
		\includegraphics[width=0.16\textwidth]{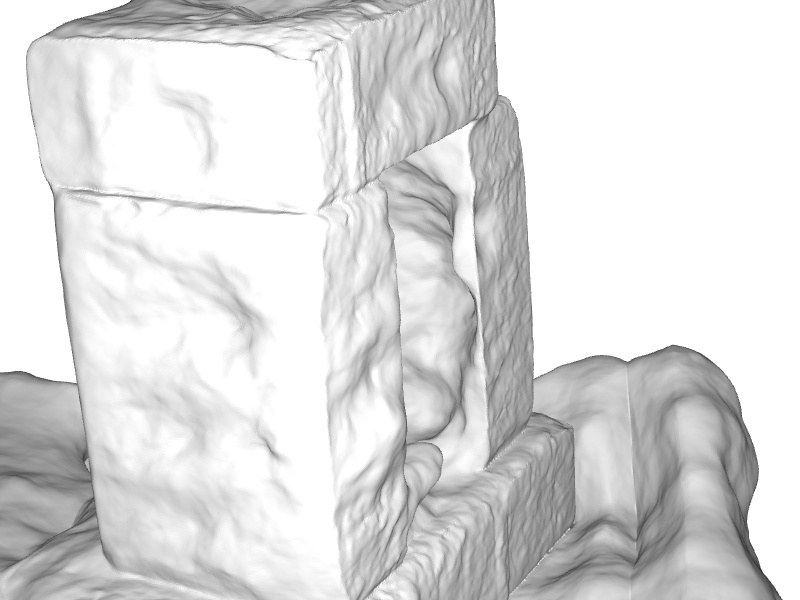}&
		\includegraphics[width=0.16\textwidth]{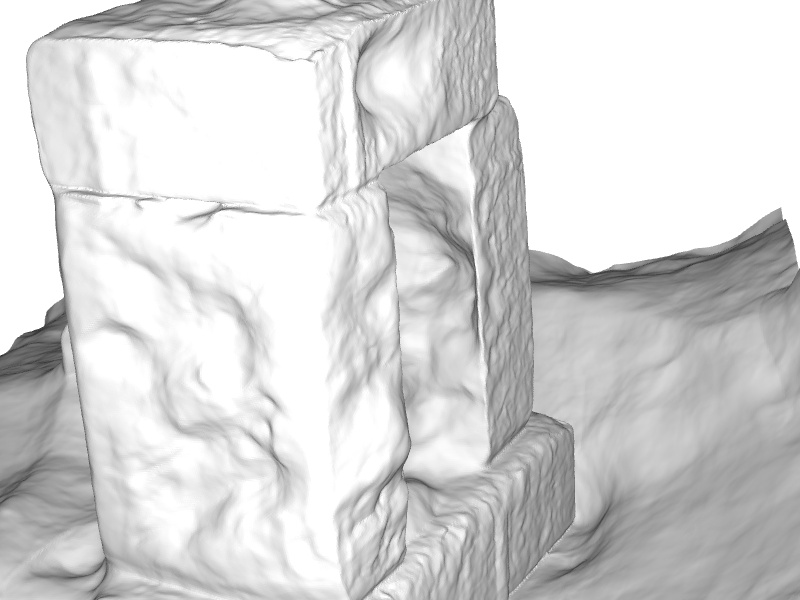}&
		\includegraphics[width=0.16\textwidth]{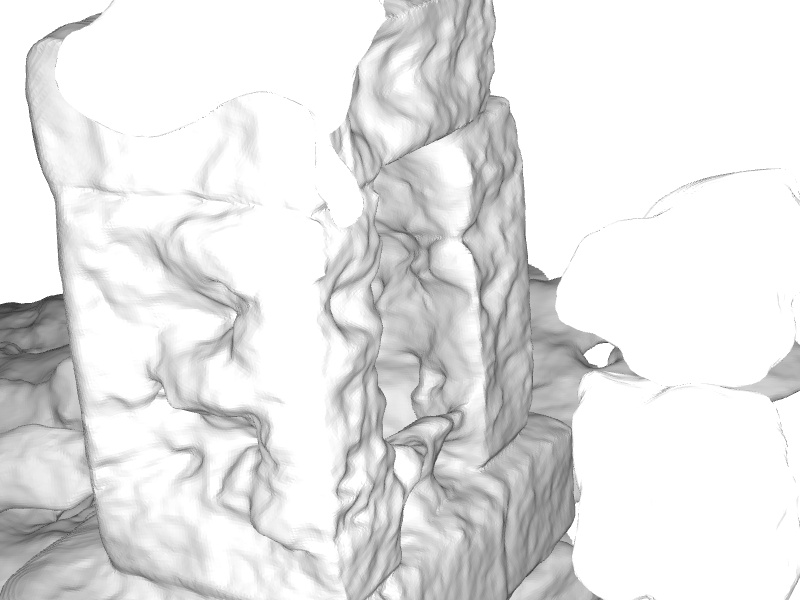}&
		\includegraphics[width=0.16\textwidth]{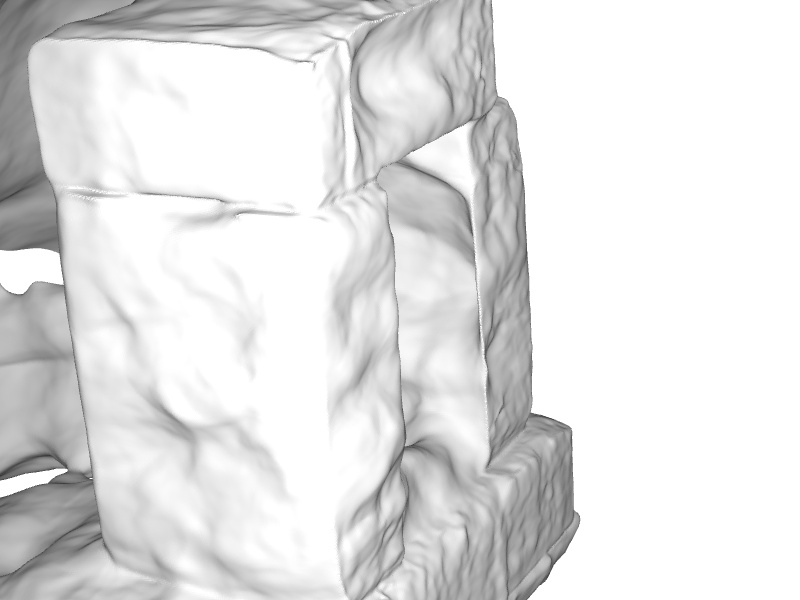}&
		\includegraphics[width=0.16\textwidth]{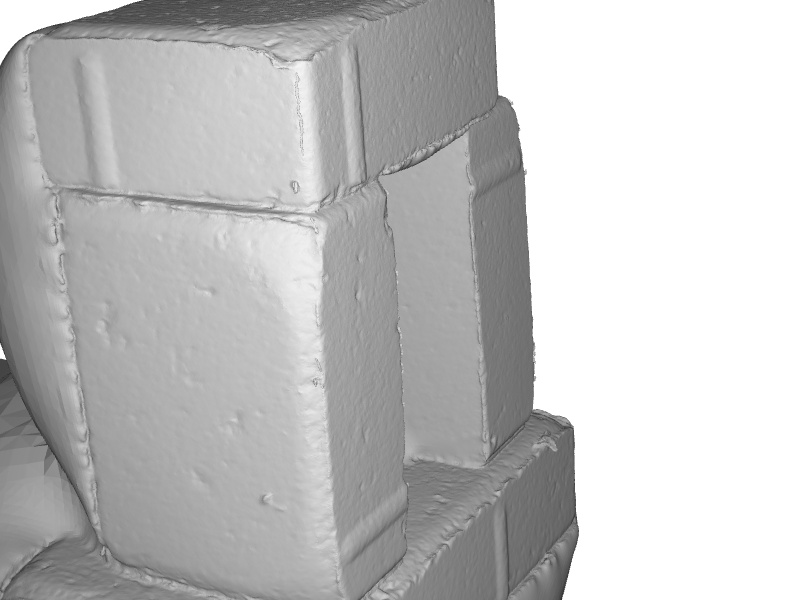}
        \\
        \raisebox{0.5\height}{\rotatebox{90}{Scan65}}
		\includegraphics[width=0.16\textwidth]{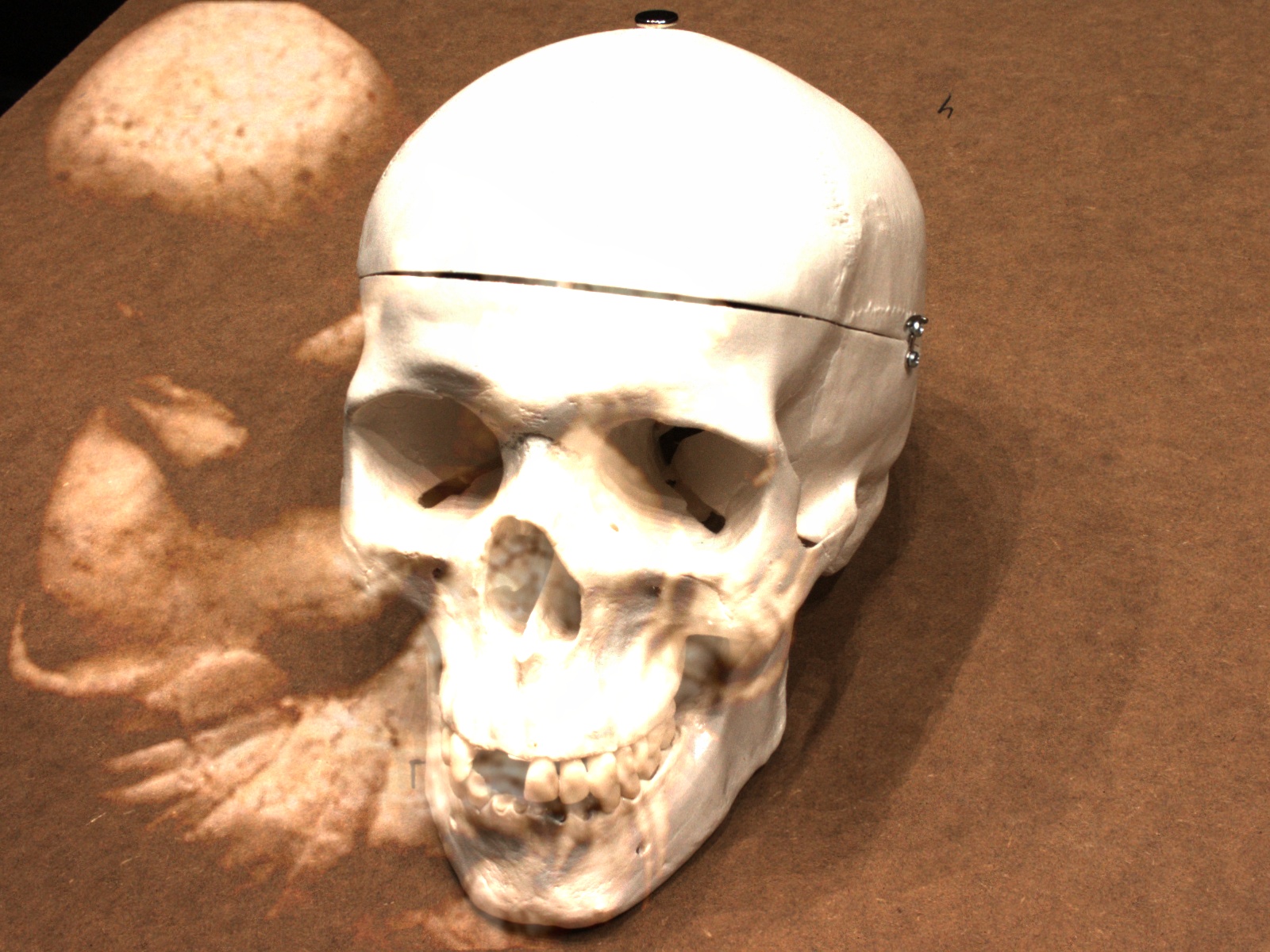}&
		\includegraphics[width=0.16\textwidth]{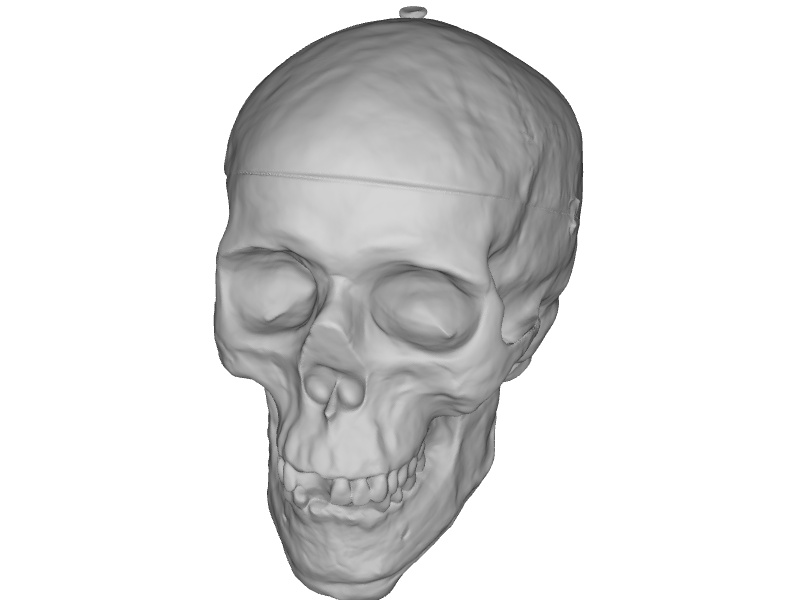}&
		\includegraphics[width=0.16\textwidth]{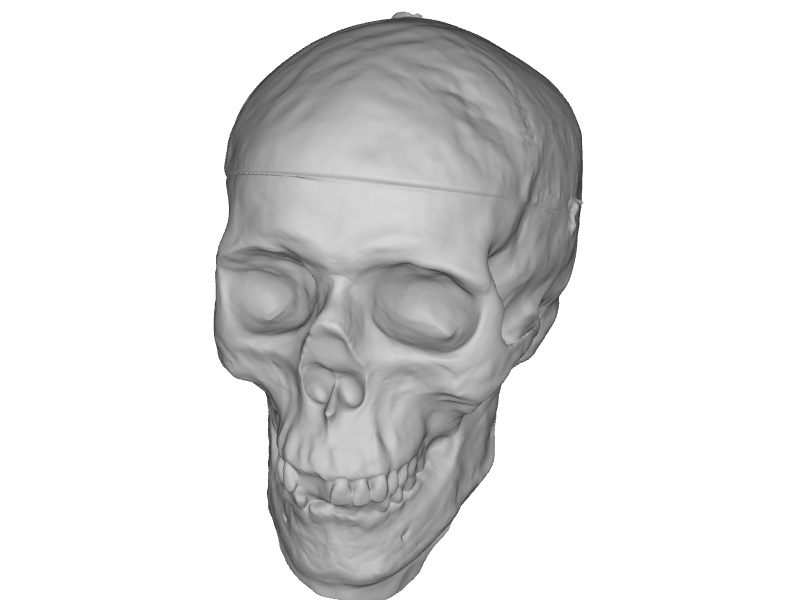}&
		\includegraphics[width=0.16\textwidth]{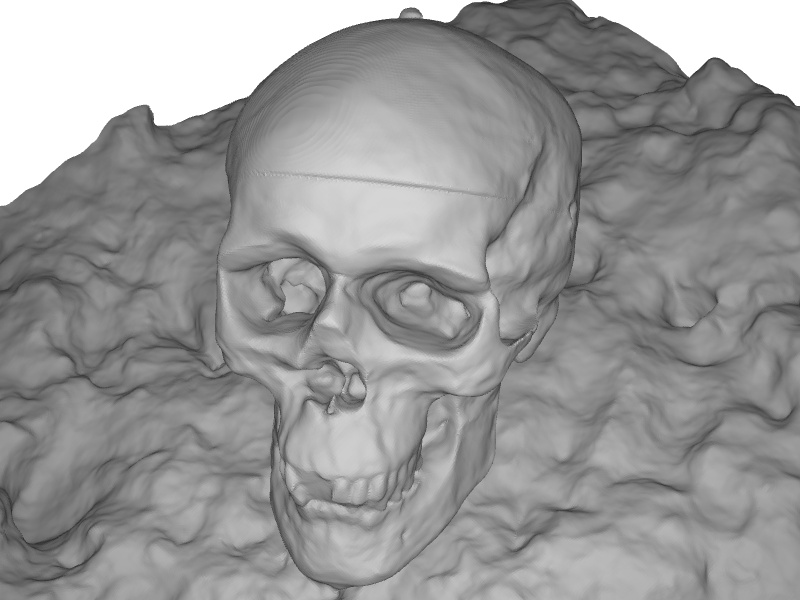}&
		&
		\includegraphics[width=0.16\textwidth]{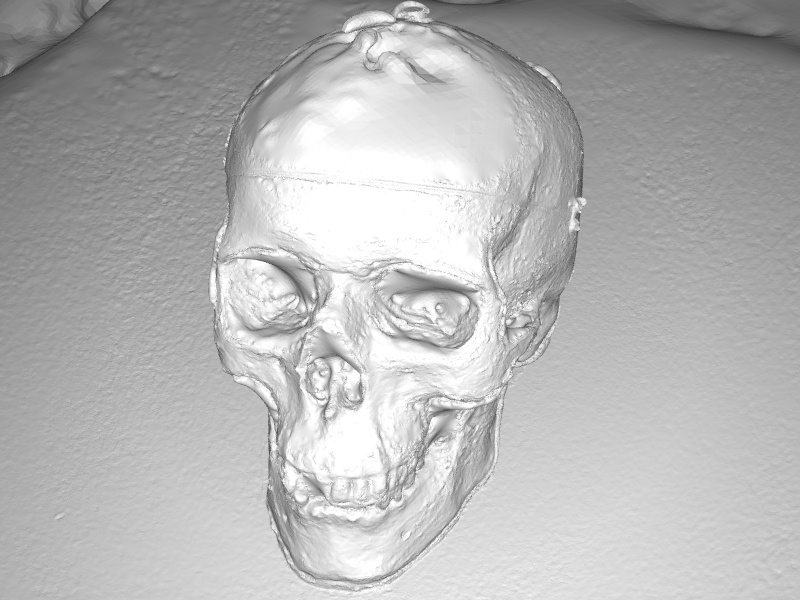}
        \\
        \raisebox{0.5\height}{\rotatebox{90}{Scan69}}
		\includegraphics[width=0.16\textwidth]{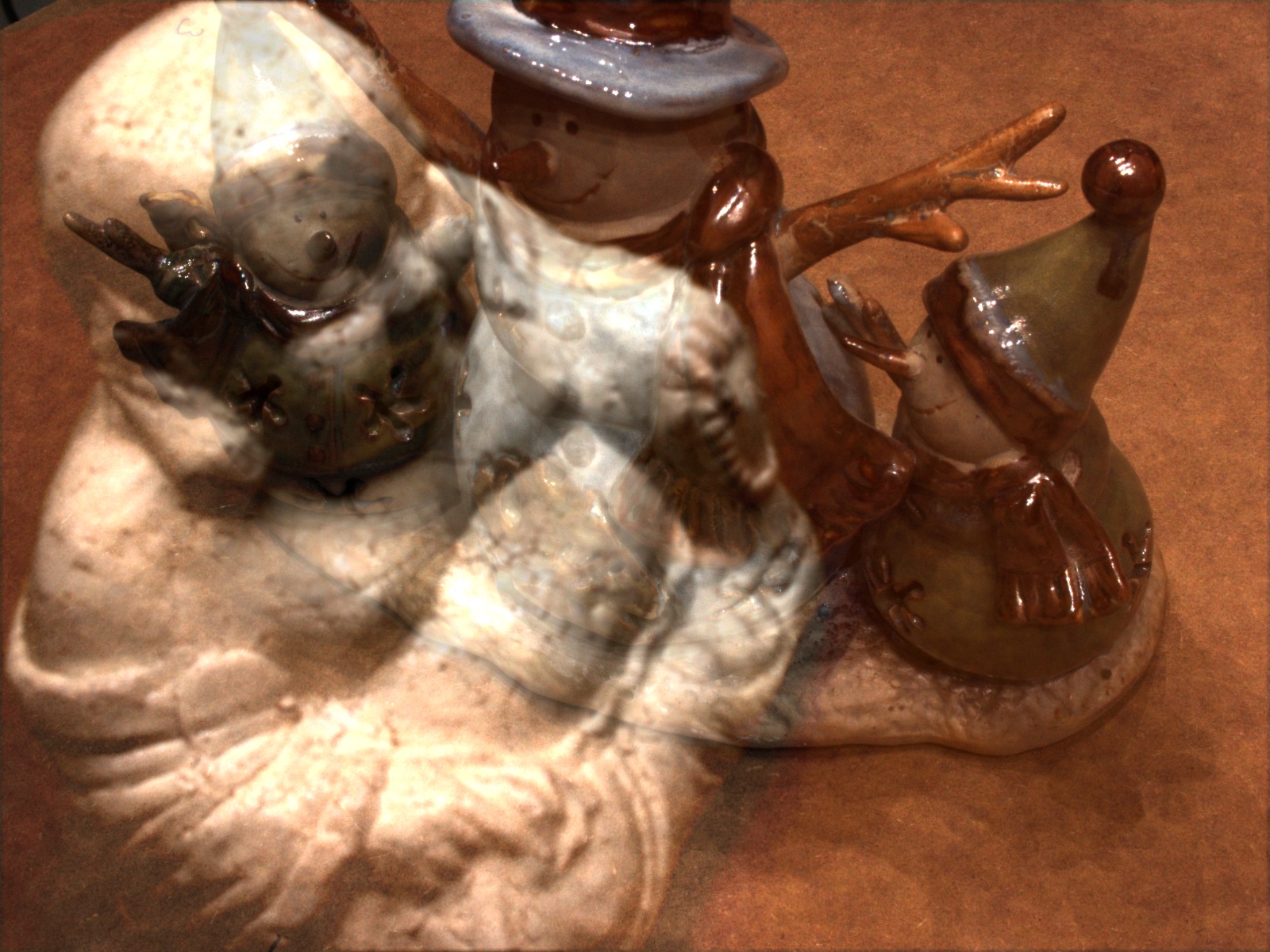}&
		\includegraphics[width=0.16\textwidth]{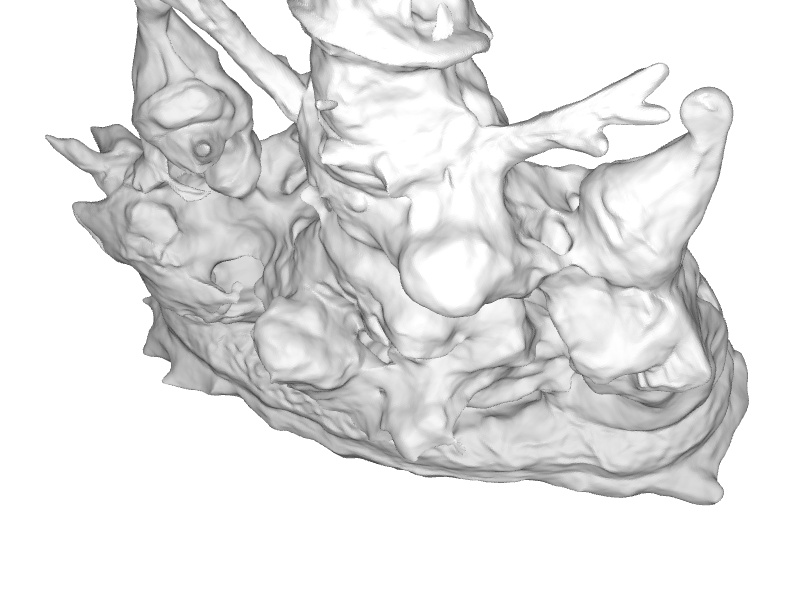}&
		\includegraphics[width=0.16\textwidth]{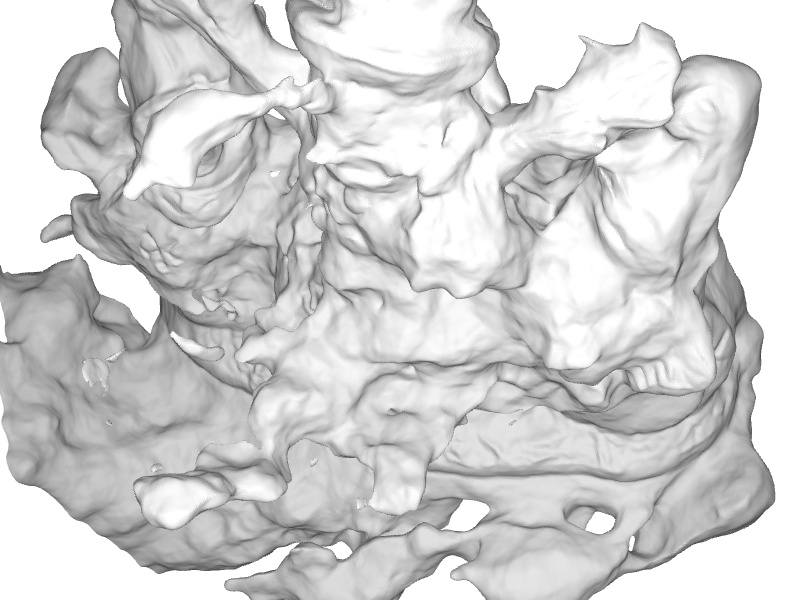}&
		\includegraphics[width=0.16\textwidth]{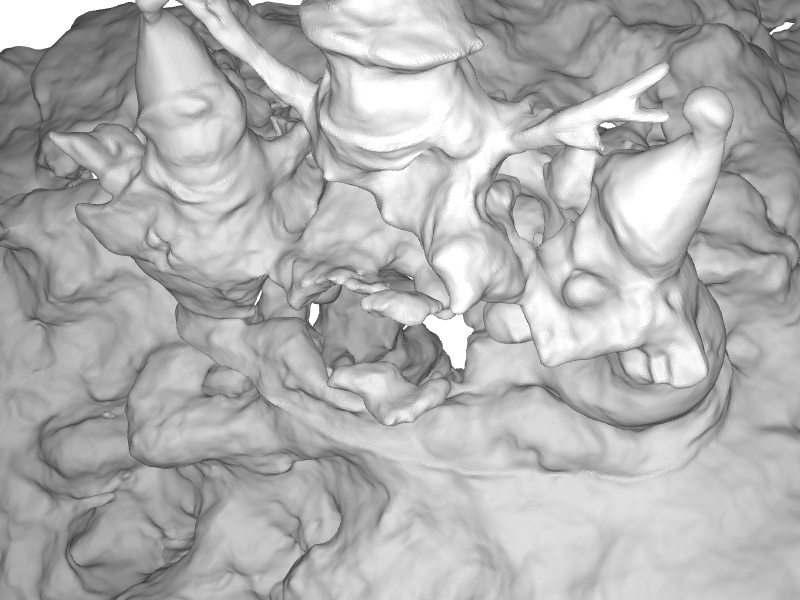}&
		&
		\includegraphics[width=0.16\textwidth]{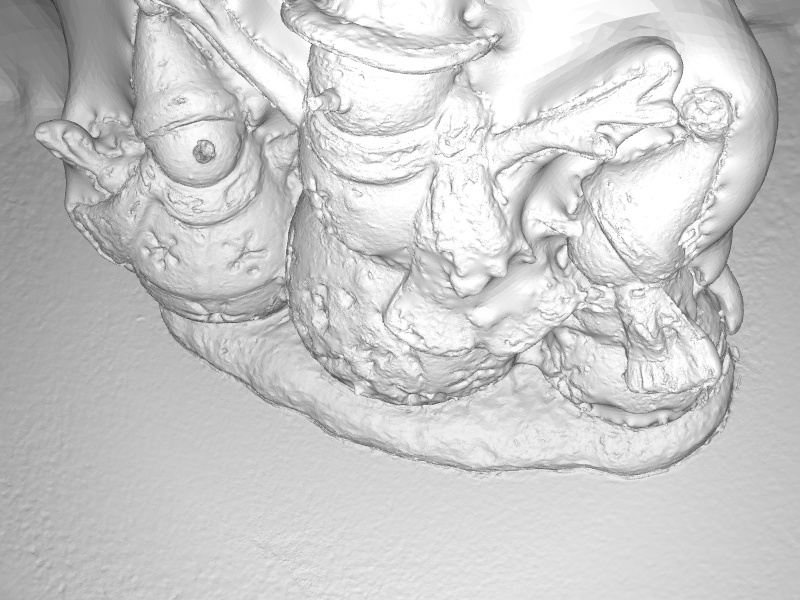}
        \\
        \raisebox{0.5\height}{\rotatebox{90}{Scan105}}
		\includegraphics[width=0.16\textwidth]{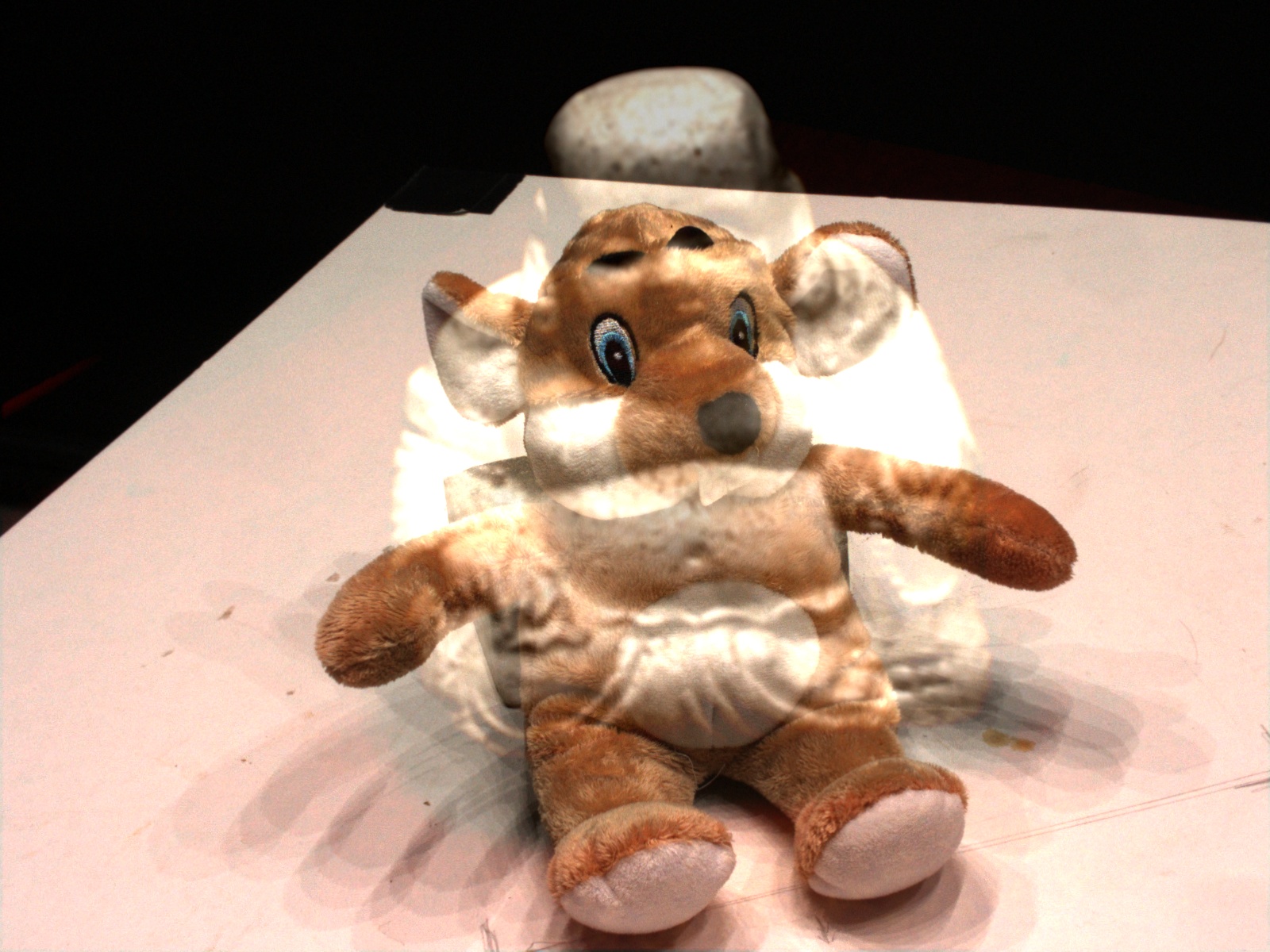}&
		\includegraphics[width=0.16\textwidth]{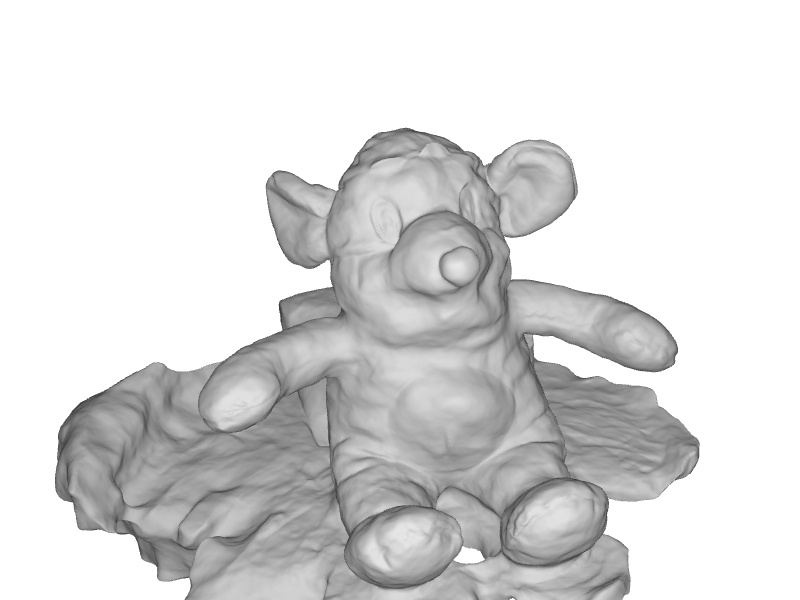}&
		\includegraphics[width=0.16\textwidth]{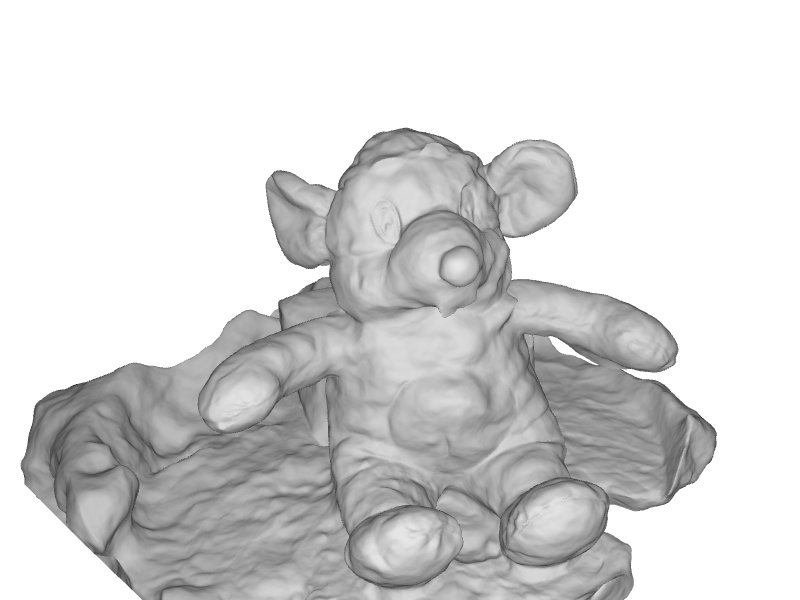}&
		\includegraphics[width=0.16\textwidth]{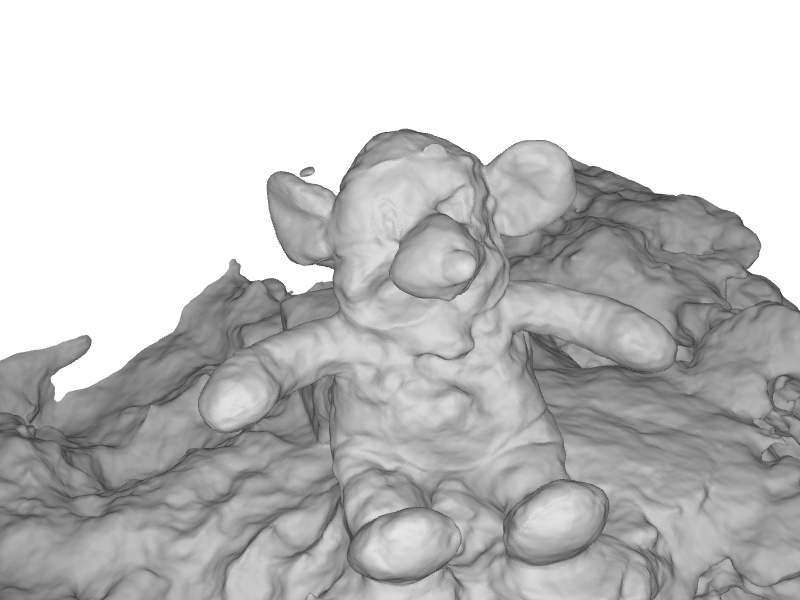}&
		&
		\includegraphics[width=0.16\textwidth]{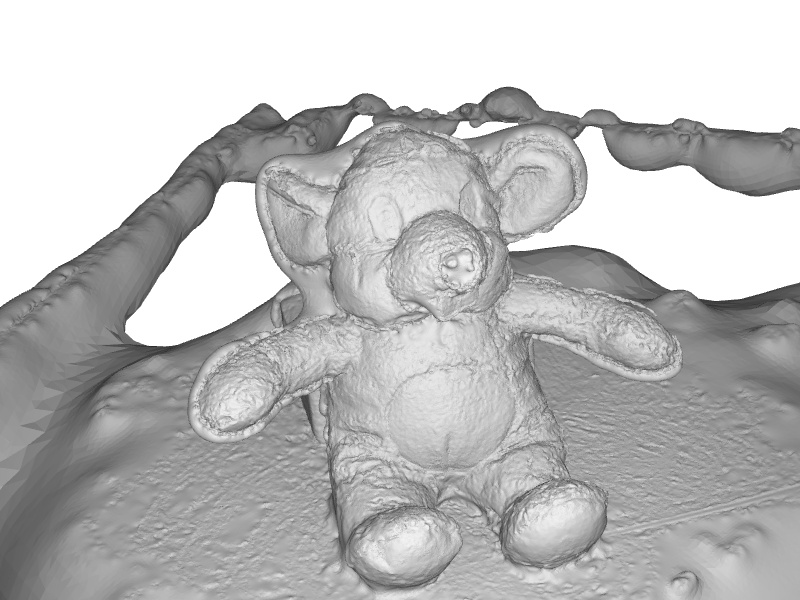}
        \\
		\raisebox{0.4\height}{Supervision}&
		\raisebox{0.4\height}{Ours}&
		\raisebox{0.4\height}{NeuS}&
		\raisebox{0.4\height}{UNISURF}&
		\raisebox{0.4\height}{VolSDF}&
		\raisebox{0.4\height}{COLMAP}
	\end{tabular}
	\caption{Qualitative comparisons between NeuS-HSR and baselines on the synthetic dataset.}
	\label{supp_syn0}
\end{figure*}

More qualitative comparisons between NeuS-HSR and other \sArt methods on the synthetic dataset are shown in \figref{supp_syn0}. All neural implicit approaches are trained without ground-truth masks.
COLMAP~\cite{schoenberger2016mvs} generates too much noise around the target object surface to calculate the metric (\ie, Chamfer distance) of its results in our manuscript.

\section{Trainable Standard Deviation on the Real-World Dataset}\label{Sec:mrrw}
In NeuS~\cite{wang2021neus}, the optimization process reduces the standard deviation automatically then the surface becomes sharper. We conduct a comparison between NeuS-HSR (blue curve) and NeuS (orange curve) on the trainable standard deviation. The standard deviation of our method converges to a smaller value compared to the standard deviation of NeuS, and our method achieves clearer and sharper results on the real-world dataset than NeuS as our manuscript shows. 

\section{Future Work}\label{Sec:fw}
In the future, we plan to extend our approach to handle glasses with different thicknesses. In our daily life, the thicker the glass, the more obvious the specular reflections. One possible scheme may be adding thickness to our auxiliary plane module. Besides, our method can also evolve to tackle highly reflective object surfaces (\eg, cars).  

\begin{figure*}[t]
	\centering
	\begin{tabular}{*{5}{c@{\hspace{1px}}}}
		\includegraphics[width=0.24\textwidth]{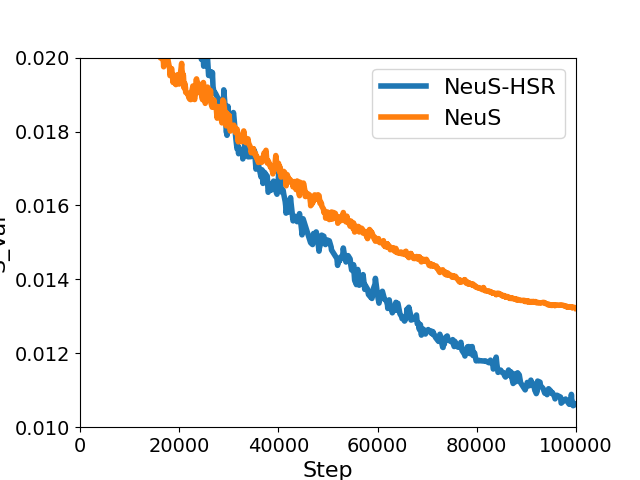}&
		\includegraphics[width=0.24\textwidth]{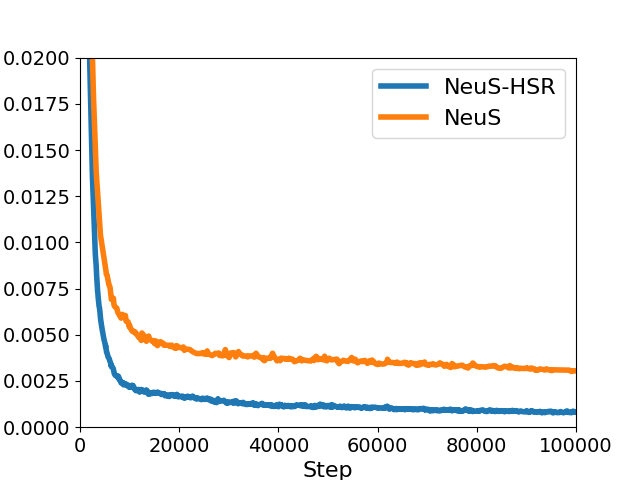}&
		\includegraphics[width=0.24\textwidth]{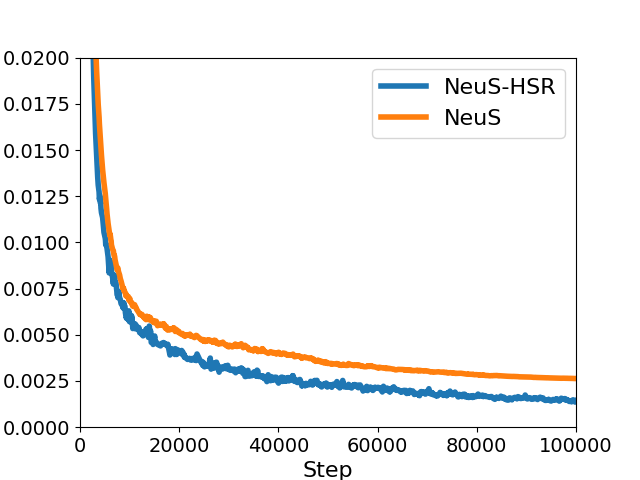}&
		\includegraphics[width=0.24\textwidth]{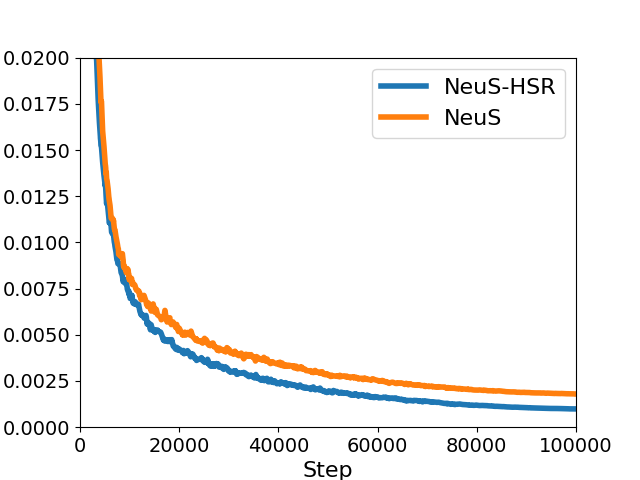}
		\\
		\raisebox{0.4\height}{Buddha}&
		\raisebox{0.4\height}{Plate}&
		\raisebox{0.4\height}{Porcelain}&
		\raisebox{0.4\height}{Bronze}
	\end{tabular}
	\caption{Comparison of trainable standard deviation.}
	\label{supp_real_s}
\end{figure*}

\end{document}